%% file: arxiv_main.tex
\pdfoutput=1

\documentclass{article}
\usepackage{arxiv}
\usepackage{graphicx}
\usepackage{float}

\usepackage[utf8]{inputenc} 
\usepackage[T1]{fontenc}    
\usepackage{hyperref}       
\usepackage{url}            
\usepackage{booktabs}       
\usepackage{amsfonts}       
\usepackage{amsmath}
\usepackage{amssymb}
\usepackage{dsfont}
\usepackage{mathtools}
\usepackage{amsthm}
\usepackage[lined,boxed,ruled,norelsize,algo2e]{algorithm2e}

\usepackage{bm}
\usepackage{nicefrac}       
\usepackage{microtype}      
\usepackage{xcolor}         
\usepackage{parskip}
\usepackage{overpic}
\usepackage{multirow}
\usepackage{makecell}
\usepackage{soul}
\usepackage{cleveref}
\usepackage{subcaption}

\theoremstyle{remark}

\theoremstyle{plain}
\newtheorem{theorem}{Theorem}
\newtheorem{proposition}{Proposition}
\newtheorem{corollary}{Corollary}
\newtheorem{lemma}{Lemma}

\theoremstyle{definition}

\newtheorem{assumption}{Assumption}

\theoremstyle{plain}
\makeatletter
\newtheorem*{rep@theorem}{\rep@title}
\newcommand{\newreptheorem}[2]{%
  \newenvironment{rep#1}[1]{%
     \def\rep@title{#2 \ref{##1}}%
     \begin{rep@theorem}
  }%
  {\end{rep@theorem}}%
}
\makeatother

\newreptheorem{theorem}{Theorem}
\newreptheorem{lemma}{Lemma}
\newreptheorem{proposition}{Proposition}
\newreptheorem{corollary}{Corollary}

\crefname{equation}{Eq.}{Eqs.}
\crefname{assumption}{Assumption}{Assumptions}
\crefname{definition}{Definition}{Definitions}
\crefname{theorem}{Theorem}{Theorems}
\crefname{lemma}{Lemma}{Lemmas}
\crefname{proposition}{Proposition}{Propositions}
\crefname{remark}{Remark}{Remarks}
\crefname{corollary}{Corollary}{Corollaries}
\crefname{algocf}{Algorithm}{Algorithms}
\crefname{figure}{Figure}{Figures}
\crefname{table}{Table}{Tables}

\usepackage[toc,page,header]{appendix}
\usepackage{titletoc}

\usepackage{comment}

\input{notation}

\title{Risk-Averse Inference-Time Diffusion Model Alignment with Finite-Sample Guarantees}

\author{%
  Yuchen Liang\thanks{Equal contribution.} \\
  {\small Indiana University Indianapolis}
  \And
  Xian Yu\footnotemark[1]\textsuperscript{,}%
  \thanks{Corresponding author: Xian Yu
    (\texttt{yu.3610@osu.edu}).} \\
  {\small The Ohio State University}
}

\begin{document}

\maketitle

\begin{abstract}
Inference-time alignment steers diffusion models toward downstream objectives without costly fine-tuning, but existing methods typically optimize expected cost (or reward) without explicitly controlling tail risk.
In this work, we propose a risk-averse framework for inference-time alignment of diffusion models that balances expected cost and Conditional Value-at-Risk (CVaR) under trajectory-level Kullback-Leibler regularization. We characterize the optimal path distribution and reverse transitions as functions of the CVaR threshold, reducing the alignment problem to a one-dimensional optimization problem over the threshold. We show that this one-dimensional optimization problem can be nonconvex in general, establish sufficient conditions for convexity, and introduce quadratic regularization to enable efficient derivative-free ternary search. Under suitable assumptions, we provide finite-sample guarantees for threshold optimization and value estimation, and combine them into a terminal total-variation bound relative to the regularized target under exact resampling. Experiments on synthetic data and image generation using Stable Diffusion demonstrate improved tail performance over risk-neutral baselines while maintaining competitive average cost.
\end{abstract}

\input{intro-and-works}

\input{background}

\input{risk-averse-method}

\input{theory}

\input{experiment}

\input{conclusion}


\bibliography{diffusion,refs,Xian_bib}

\newpage
\appendix
\begin{center}
    \huge \textbf{Appendix}
\end{center}


\allowdisplaybreaks
\startcontents[section]
{
\hypersetup{linkcolor=blue}
\printcontents[section]{l}{1}{\setcounter{tocdepth}{2}}
}

\crefalias{section}{appendix}
\crefalias{subsection}{appendix}
\crefalias{subsubsection}{appendix}

\newpage

\input{appendix}

\end{document}

%% file: notation.tex
\usepackage{graphicx,accents}

\providecommand{\E}{\mathbb{E}}
\providecommand{\Var}{\mathrm{Var}}

\newcommand{\norm}[1]{\left\lVert #1 \right\rVert}
\newcommand{\abs}[1]{\left| #1 \right|}
\makeatletter
\DeclareRobustCommand{\rev}[1]{%
  \mathpalette\do@rev{#1}%
}
\newcommand{\do@rev}[2]{%
  \fix@rev{#1}{+}%
  \reflectbox{$\m@th#1\vec{\reflectbox{$\fix@rev{#1}{-}\m@th#1#2\fix@rev{#1}{+}$}}$}%
  \fix@rev{#1}{-}%
}
\newcommand{\fix@rev}[2]{%
  \ifx#1\displaystyle
    \mkern#23mu
  \else
    \ifx#1\textstyle
      \mkern#23mu
    \else
      \ifx#1\scriptstyle
        \mkern#22mu
      \else
        \mkern#22mu
      \fi
    \fi
  \fi
}
\makeatother

\providecommand{\KL}[2]{\mathrm{KL}( #1 || #2 )}
\providecommand{\TV}[2]{\mathrm{TV}( #1 , #2 )}
\newcommand{\brc}[1]{\left( #1 \right)}
\newcommand{\sbrc}[1]{\left[ #1 \right]}
\newcommand{\cbrc}[1]{\left\{ #1 \right\}}
\providecommand{\eps}{\varepsilon}

\renewcommand{\d}{\mathrm{d}}

\providecommand{\mbP}{\mathbb{P}}
\providecommand{\mbR}{\mathbb{R}}

\providecommand{\calN}{\mathcal{N}}
\providecommand{\calX}{\mathcal{X}}

\newcommand{\cvar}{\mathrm{CVaR}}

%% file: intro-and-works.tex
\section{Introduction}

Pretrained diffusion models \cite{ho2020ddpm,song2020sde} have shown remarkable success in generating high-quality images and videos \cite{dalle2,rombach2022stable-diffusion,melnik2024video-diffusion-survey}, but their outputs may not meet the specific needs of a downstream application. One approach is to fine-tune the pretrained model using task-specific rewards \cite{uehara2024finetuning}, but repeated fine-tuning can be expensive as needs change and may be infeasible when model weights are unavailable. An alternative approach is inference-time alignment, which steers the sampling process toward desired outputs without modifying the pretrained model \cite{li2025svdd,singhal2025fksteering}.
A classical formulation of inference-time alignment is to optimize the expected cost (resp. reward) at the output, while ensuring that the steered sampling path is close to the pretrained one \cite{uehara2024finetuning,li2025svdd}.
However, in practice, good average performance does not preclude the possibility of undesirable tail outcomes.
For example, 
in molecular design where the expected chemical stability is optimized \cite{wang2026cvar}, there could still be a non-negligible probability of generating highly unstable candidates. As another example, 
diffusion-based robot motion planning requires explicit attention to collision avoidance \cite{xiao2025safediffuser,ma2025robotics}, since strong average task performance alone does not rule out occasional dangerous trajectories.

To this end, there is an urgent need for risk-averse inference-time alignment that explicitly accounts for risk by augmenting the expectation-based objective with a coherent risk measure. Among coherent risk measures, 
Conditional Value-at-Risk (CVaR) provides a principled way to capture tail behavior, as it measures the average cost beyond a prescribed threshold (i.e., Value-at-Risk) of the output distribution. According to \cite{rockafellar2000optimization,rockafellar2002conditional}, CVaR can be expressed as an optimization problem over this one-dimensional threshold. Recent works have incorporated CVaR into generative model fine-tuning through sequential updates \cite{desanti2025fdc} or threshold-based reward transformation \cite{wang2026cvar}. However, these works focus on risk-averse fine-tuning, which can be computationally costly and impractical for large pretrained models. 

This paper aims to develop a risk-averse inference-time alignment framework that combines expected cost and CVaR under trajectory-level Kullback-Leibler (KL) regularization while keeping the pretrained model frozen. Establishing the reliability of such a framework requires a non-asymptotic analysis of its sampling accuracy. While existing risk-averse fine-tuning work studies sensitivity to threshold errors \cite{wang2026cvar}, inference-time alignment must additionally account for estimation errors in the value functions used to guide sampling. A central theoretical challenge is therefore to characterize how errors in CVaR threshold optimization and value function estimation jointly affect the final output distribution, and how many samples are needed to achieve a prescribed accuracy. This leads to the following question:

\begin{center}
\emph{How can we align pretrained diffusion models at \underline{inference time} to control \underline{tail risk}, and what \underline{finite-sample guarantees} can we establish?}
\end{center}

\subsection{Our Contributions}

Our main contributions can be summarized as follows.

\noindent\textbf{CVaR-based inference-time alignment framework.}
We characterize the optimal risk-averse sampling policy as a function of the CVaR threshold, reducing the risk-averse alignment problem to a one-dimensional threshold optimization problem. This characterization enables existing risk-neutral inference-time samplers, including SVDD and SMC, to incorporate tail-risk control through risk-adjusted value functions while keeping the pretrained diffusion backbone fixed.

\noindent\textbf{Efficient ternary search algorithm for CVaR threshold optimization.}
We show that the threshold objective can be nonconvex in general and provide sufficient conditions for convexity. For the potentially nonconvex case, we introduce quadratic regularization to obtain a strongly convex objective and employ a ternary search algorithm with an inexact objective evaluation oracle, establishing finite-iteration complexity.

\noindent\textbf{End-to-end sampling guarantees.}
We further establish finite-sample guarantees for threshold optimization and value estimation under 
suitable assumptions. Under exact tilted-transition sampling, we combine these results into an end-to-end total-variation bound, 
quantifying the joint effects of threshold and value-estimation errors on sampling accuracy.

\noindent\textbf{Synthetic and image-generation experiments.}
We evaluate our risk-averse framework through extensive experiments on heavy-tailed synthetic data and Stable Diffusion image generation \cite{rombach2022stable-diffusion} with compressibility and aesthetic rewards. Notably, CVaR-SVDD improves empirical tail performance, consistent with our risk-sensitive objective. We further conduct ablation studies across different prompts, sampling procedures, and threshold-optimization algorithms. 

\subsection{Related Work}
\label{sec:related-works}

We review the most relevant works below and discuss additional related works in \Cref{app:additional-related-works}.

\paragraph{CVaR-based generative alignment.}
Recent works have incorporated CVaR into generative model fine-tuning.
Flow Density Control \cite{desanti2025fdc} optimizes distributional objectives, including CVaR, through a sequence of fine-tuning problems.
Tail-aware Flow Fine-Tuning \cite{wang2026cvar} characterizes an optimal terminal distribution through scalar threshold optimization and uses the resulting transformed reward for fine-tuning.
Differently, our framework incorporates CVaR at inference time while keeping the pretrained model fixed.
Related inference-time work, RS-Diffuser \cite{gong2026rsdiffuser}, guides denoising using gradients of CVaR computed from a distributional critic conditioned on noisy candidate plans. 
In contrast, our formulation optimizes mean-CVaR over the terminal distribution induced by the sampling policy, under trajectory-level KL regularization.

\paragraph{Theoretical guarantees for alignment.}
Several recent works have established theoretical guarantees for inference-time alignment.
Learnable Chernoff Baselines \cite{madhow2026lcb} provide total-variation guarantees for approximate rejection sampling from exponentially tilted transitions.
DOIT \cite{zhu2026doit} analyzes sampling accuracy under approximation of the dynamic correction induced by Doob's $h$-transform.
\cite{zhu2026smc} studies how approximate value functions, quantified through Bellman errors, affect inference-time sampling, including finite-particle SMC complexity.
For CVaR-based fine-tuning, \cite{wang2026cvar} also studies sensitivity of the target distribution to threshold errors.
Our analysis complements these results by establishing sample-complexity bounds for CVaR threshold optimization and value estimation, and quantifying how these two sources of error jointly affect the final output distribution in total variation.

%% file: background.tex
\section{Background}

In this section, we introduce the necessary background.

\subsection{(Continuous-Space) Diffusion Models}

Diffusion models \cite{ho2020ddpm,song2020sde} generate samples through a reverse process that progressively removes noise added by a forward process. Given initial data $x_0 \in \mbR^d$, the forward process gradually adds Gaussian noise:
\[
    x_t = \sqrt{1-\beta_t}\,x_{t-1} + \sqrt{\beta_t}\,w_t,
    \qquad t=1,\dots,T,
\]
where $w_t \stackrel{i.i.d.}{\sim} \calN(0,I_d)$ and
$\beta_t \in (0,1)$ is the \textit{noise schedule}.
Define $\alpha_t := 1-\beta_t$ and
$\Bar{\alpha}_t := \prod_{i=1}^t \alpha_i$.
Accumulating the forward steps gives
\[
    q_{t|0}(x_t \mid x_0) =
    \calN\bigl(\sqrt{\Bar{\alpha}_t}\,x_0,
    (1-\Bar{\alpha}_t)I_d\bigr).
\]
Thus, when $\Bar{\alpha}_T$ is sufficiently small, the terminal
distribution is approximately standard Gaussian.

The reverse sampling process starts from
$x_T \sim q_T := \calN(0,I_d)$ and applies learned denoising
transitions for $t=T,\dots,1$. Tweedie's formula yields (cf. \cite{tweedie2011efron})
\[ q_{t-1|t}(x_{t-1}|x_t) = \calN\brc{m_t(x_t), \sigma_t^2 I_d }, \]
where $m_t(x_t) := \frac{1}{\sqrt{\alpha_t}} \brc{x_t+(1-\alpha_t)s_t(x_t)}$, $s_t$ estimates the \textit{score}, i.e., $\nabla \log q_t$ at time $t$, and $\sigma_t^2$
is a prescribed reverse variance.
To unify notation, we also define $x_{T+1}$ as a (dummy) deterministic state, with $q_{T|T+1}(x_T|x_{T+1}) = q_T(x_T)$.

Throughout this paper, $q$ denotes the joint distribution induced
by the pretrained sampling path:
\[
    q(x_{0:T})
    = q_T(x_T)\prod_{t=1}^T q_{t-1|t}(x_{t-1}|x_t),
\]
with $q_t$ denoting the marginal distribution at time $t$.
In particular, $q_0$ is the pretrained model's output distribution, which serves as the reference distribution for inference-time sampling.

\subsection{(Risk-Neutral) Inference-Time Alignment}

In practice, one often seeks to align a pretrained diffusion model with a reward function to generate samples with higher rewards. 
To this end, inference-time alignment aims to generate samples with higher rewards
\textit{without changing the weights of the pretrained model}.
Writing the terminal cost as $c(x_0):=-r(x_0)$, a typical
KL-regularized objective is
\cite{li2025svdd,wang2025finetuning}:
\begin{equation} \label{eq:obj-path-neutral}
    \min_{\pi}
    \E_{\pi}[c(x_0)]
    + \alpha \KL{\pi_{1:T+1}}{\mu_{1:T+1}},
\end{equation}
where $\alpha>0$ controls the regularization strength.
Here $\pi := \pi_{1:T+1}$ denotes the joint path distribution induced by
the initial distribution $\pi_{T+1}$ and reverse transitions
$\pi_t$, $t=1,\dots,T$.
The reference distribution $\mu_{1:T+1}=q$ is induced by the
pretrained transitions $\mu_t:=q_{t-1|t}$ with initialization
$\mu_{T+1}=q_T=\calN(0,I_d)$.

For $t=1,\dots,T+1$, let $V_t^*(x_t)$ denote the optimal regularized cost-to-go (or soft value function)
from state $x_t$, defined as follows:
\begin{equation}
\label{eq:opt-val-def}
    V_t^*(x_t) := \min_{\pi_{1:t}} \E_\pi\sbrc{ c(x_0) + \alpha \KL{\pi_{1:t}}{\mu_{1:t}}\mid x_t}.
\end{equation}
Here $\pi_{1:t}(\cdot\mid x_t)$ and
$\mu_{1:t}(\cdot\mid x_t)$ denote the laws of the
remaining trajectory $x_{0:t-1}$ starting from $x_t$.
In other words, $V_t^*(x_t)$ accounts for both the expected terminal
cost and the accumulated KL penalties over the remaining
reverse steps. 
Note that the terminal condition is
$V_0^*(x_0)=c(x_0)$.
From \cite{li2025svdd}, 
the optimal soft value function at time $t$ has the following recursion
\begin{equation}
\label{eq:vt-induction-def}
    V_t^*(x_t) = - \alpha \log \E_{\mu_t} \sbrc{e^{-V^*_{t-1}(x_{t-1}) / \alpha} \bigg| x_t},
\end{equation}
and the corresponding optimal reverse transitions satisfy
\begin{align*}
    \pi_t^*(x_{t-1}|x_t)
    &= \mu_t(x_{t-1}|x_t)
    e^{-(V_{t-1}^*(x_{t-1})-V_t^*(x_t))/\alpha} \\
    &\propto \mu_t(x_{t-1}|x_t)
    e^{-V_{t-1}^*(x_{t-1})/\alpha}.
\end{align*}
Concatenation over $t$ further yields 
\begin{align*}
    V_t^*(x_t) &= -\alpha \log \E_{q_{0|t}} \sbrc{e^{-V^*_0(x_{0}) / \alpha} \bigg| x_t} \\
    &= -\alpha \log \E_{q_{0|t}} \sbrc{e^{-c(x_0) / \alpha} \bigg| x_t}.
\end{align*}
Denote the initial value as $V_{T+1}^*$, which is independent of the sampling path.
Let $p_t^*$ denote the marginal distribution at time $t$ under the optimal policy. The optimal output distribution is the
exponentially tilted distribution
\cite{li2025svdd,wang2025finetuning}: $p_0^*(x_0)
    \propto q_0(x_0)e^{-c(x_0)/\alpha}$.
For the rest of this paper, we omit the conditional dependence of $V_t^*$ on $x_t$ when it is obvious from the context.

In practice, the optimal value function $V_t^*$ is typically
intractable to obtain and is replaced by an estimate $\hat{V}_t$. 
Although sampling from the pretrained reverse kernels is
straightforward, incorporating the exponential tilt requires
additional sampling procedures. Common approaches use
``sampling-resampling'' techniques, including SVDD \cite{li2025svdd}
and SMC \cite{singhal2025fksteering}. The corresponding pseudocode is
provided in \Cref{sec:inf-align-alg-reference}.

Given the value estimates $\hat{V}_t$, define the approximate
policy with exact tilting as
\[
    \hat{\pi}_t(x_{t-1}|x_t)
    \propto \mu_t(x_{t-1}|x_t)
    e^{-\hat{V}_{t-1}(x_{t-1})/\alpha},
\]
with the normalizing constant $\hat{Z}_t(x_t)
    := \E_{\mu_t} \sbrc{
        e^{-\hat{V}_{t-1}(x_{t-1})/\alpha}
        \bigg| x_t
    }$.
Unlike the optimal value functions, the estimates need not
satisfy the soft Bellman recursion \eqref{eq:vt-induction-def}. Consequently,
$\hat{Z}_t(x_t)$ generally does not equal
$e^{-\hat{V}_t(x_t)/\alpha}$.

\subsection{Conditional Value-at-Risk (CVaR)}

The risk-neutral objective in \eqref{eq:obj-path-neutral}
optimizes the expected cost without explicitly controlling
high-cost outcomes. 
To account for such tail risk, one popular method is to use CVaR \cite{rockafellar2002conditional}, which quantifies the expectation of the upper tail distribution. Given a pre-specified risk level $\beta\in(0,1)$ (close to 1), the CVaR of the cost distribution $c(x_0),\ x_0\sim \pi$ is defined as follows: 
\begin{equation} \label{eq:cvar-def}
    \cvar_{\beta,\pi}[c(x_0)]
    := \min_{\eta\in\mbR}
    \cbrc{
        \eta + \frac{1}{1-\beta}
        \E_{\pi}\sbrc{(c(x_0)-\eta)^+}
    },
\end{equation}
where $(a)^+:=\max\{0,a\}$. According to \cite{rockafellar2002conditional}, the optimal threshold for \eqref{eq:cvar-def} is given by the Value-at-Risk (VaR), i.e., the $\beta$-quantile of the distribution $c(x_0)$:
\[
    \eta_\pi^*
    = \mathrm{VaR}_{\beta,\pi}[c(x_0)]
    := \inf\cbrc{
        v:\mbP_\pi\cbrc{c(x_0)\leq v}\geq\beta
    },
\]
which depends on $\pi$. For notational simplicity, we suppress this dependence and write $\eta^*$ instead.

%% file: risk-averse-method.tex
\section{Risk-averse Inference-time Alignment}

To balance average performance and tail risk, we consider
the following convex combination
\begin{equation} \label{eq:cvar-weighted-def}
    \rho_{\beta,\pi}[c(x_0)]
    := (1-\lambda)\,\E_\pi[c(x_0)]
    + \lambda\,\cvar_{\beta,\pi}[c(x_0)],
\end{equation}
where $\lambda\in[0,1]$ controls the degree of risk aversion.
The choices $\lambda=0$ and $\lambda=1$ recover the expected
cost and CVaR, respectively. Replacing the expectation with $\rho_{\beta,\pi}$, a risk-averse counterpart of
\eqref{eq:obj-path-neutral} is
\[
    \min_{\pi_{1:T+1}}
    \rho_{\beta,\pi}[c(x_0)]
    + \alpha\KL{\pi_{1:T+1}}{\mu_{1:T+1}}.
\]
Using the threshold representation of CVaR in \eqref{eq:cvar-def}, it is
equivalent to optimize jointly over $\pi$ and $\eta$:
\begin{equation} \label{eq:obj-path-risk}
    \min_{\pi,\eta}
        \E_\pi\sbrc{c_{\lambda,\eta}(x_0)}
        + \alpha\KL{\pi_{1:T+1}}{\mu_{1:T+1}},
\end{equation}
where the risk-adjusted terminal cost is defined as
\begin{equation} \label{eq:weighted-cost-def}
    c_{\lambda,\eta}(x_0)
    := (1-\lambda)c(x_0)
    + \lambda\brc{
        \eta+\frac{(c(x_0)-\eta)^+}{1-\beta}
    }.
\end{equation}

For each fixed $\eta$, let $V_{t,\eta}^*$ denote the optimal regularized cost-to-go obtained by replacing $c$ with $c_{\lambda,\eta}$ in the risk-neutral problem \eqref{eq:opt-val-def}.
The following theorem characterizes the optimal path distribution $\pi^*$ as a function of $\eta$, reducing \eqref{eq:obj-path-risk} to a one-dimensional optimization problem over $\eta$.

\begin{theorem}[Reduction to $\eta$ optimization]
\label{thm:risk-averse-reduction}
For every $\eta\in\mbR$, assume that $\mathbb E_{q_0}\!\left[
(1+|c(x_0)|) e^{-c_{\lambda,\eta}(x_0)/\alpha} \right]<\infty$. Then, the optimal reverse transitions
in \eqref{eq:obj-path-risk} satisfy
\[ \pi_{t,\eta}^*(x_{t-1}\mid x_t) \propto \mu_t(x_{t-1}\mid x_t) e^{-V_{t-1,\eta}^*(x_{t-1})/\alpha} \]
for each $t=T+1,\ldots,1$, where
\begin{equation}
\label{eq:Vt-eta-star-def}
    V_{t,\eta}^*(x_t) = -\alpha \log \E_{q_{0|t}} \sbrc{ e^{- c_{\lambda,\eta}(x_0) / \alpha} \Big| x_t},
\end{equation}
and $V_{0,\eta}^*(x_0) := c_{\lambda,\eta}(x_0)$.
Thus, \eqref{eq:obj-path-risk} reduces to the
one-dimensional optimization problem over $\eta$:
{\small
\begin{align} \label{eq:obj-eta}
    &\min_{\eta \in \mbR} G_\lambda(\eta) := V_{T+1,\eta}^* = - \alpha\log \E_{q_0} \sbrc{e^{-c_{\lambda,\eta}(x_0)/\alpha}} \nonumber\\
    &= \lambda \eta - \alpha \log \E_{q_0} \sbrc{e^{-\brc{(1-\lambda) c(x_0) +  \frac{\lambda}{1-\beta} (c(x_0)-\eta)^+}/\alpha} }.
\end{align}
}
\end{theorem}

The proof of \Cref{thm:risk-averse-reduction} is provided in \Cref{app:thm:risk-averse-reduction-proof}. It uses the same line of proof as \cite[Theorem~1]{uehara2024finetuning}, with the terminal cost replaced by $c_{\lambda,\eta}$. For each fixed $\eta$, the risk-averse problem therefore retains the structure of risk-neutral alignment: the pretrained reverse kernels remain unchanged, while the risk preference is incorporated through the modified value functions $V_{t,\eta}^*$. As a result, existing inference-time samplers, such as SVDD and SMC, can still be adapted by using the corresponding risk-adjusted value estimates, as long as the optimal $\eta^*$ is given.
For completeness, \Cref{alg:cvar-svdd} presents the CVaR extension of the SVDD sampler. The SMC sampler can be adapted analogously.

\begin{algorithm2e}
\caption{CVaR-SVDD}
\label{alg:cvar-svdd}
\SetAlgoLined
\DontPrintSemicolon

\textbf{Input:} Pretrained reverse kernels
$\{\mu_t\}_{t=1}^{T+1}$,
threshold $\eta$,
risk-adjusted cost $c_{\lambda,\eta}$,
value estimates $\{\widehat V_{t,\eta}\}_{t=1}^{T}$,
number of samples $N$,
number of candidates $M$,
temperature $\alpha>0$\;

Set $\widehat V_{0,\eta}=c_{\lambda,\eta}$\;

Run $\mathtt{SVDD}$ (\Cref{alg:svdd}) with the given
$\{\mu_t\}_{t=1}^{T+1}$, $N$, $M$, and $\alpha$, using
$\widehat V_t=\widehat V_{t,\eta}$ for
$t=0,\ldots,T$\;

\textbf{Output:} Samples $\{x_0^{(i)}\}_{i=1}^{N}$
returned by $\mathtt{SVDD}$\;

\end{algorithm2e}

\section{Threshold Optimization}

To employ \Cref{alg:cvar-svdd}, it remains to determine the optimal $\eta^*$. This section analyzes the convexity of the threshold optimization problem \eqref{eq:obj-eta} and proposes a ternary search algorithm to solve it efficiently.

\subsection{Convexity of the Threshold Objective}

The following proposition shows that the objective function $G_\lambda(\eta)$ in \eqref{eq:obj-eta} is nonconvex in general and provides a sufficient and necessary condition for establishing its convexity.

\begin{proposition}[Non-convexity of $G_\lambda$]
\label{prop:eta-obj-nonconvex}
Suppose that the distribution of $c(x_0)$ under $q_0$
admits a continuous density $h$. 
Define
\begin{align*}
    A_\lambda(\eta)
    &:= \E_{q_0}\sbrc{
        e^{-c_{\lambda,\eta}(x_0)/\alpha}
        \mathds{1}\cbrc{c(x_0)\leq\eta}
    }, \\
    B_\lambda(\eta)
    &:= \E_{q_0}\sbrc{
        e^{-c_{\lambda,\eta}(x_0)/\alpha}
        \mathds{1}\cbrc{c(x_0)>\eta}
    },
\end{align*}
and assume that
$Z_\lambda(\eta):=A_\lambda(\eta)+B_\lambda(\eta)$
is finite for every $\eta\in\mbR$.
Then $G_\lambda$ in \eqref{eq:obj-eta} is convex
if and only if
\begin{equation}
\label{eq:eta-convexity-condition-general-lambda}
    e^{-\eta/\alpha}h(\eta)Z_\lambda(\eta)
    \geq
    \frac{\lambda}{\alpha(1-\beta)}
    A_\lambda(\eta)B_\lambda(\eta)
\end{equation}
for every $\eta\in\mbR$. Moreover, we have $G_\lambda'(\eta) =\lambda-\frac{\lambda}{1-\beta} \cdot    \frac{B_\lambda(\eta)}{Z_\lambda(\eta)}$ and $G_\lambda''(\eta)  \geq -
\frac{\lambda^2}{4\alpha(1-\beta)^2}$ for all $\eta\in\mbR$.
\end{proposition}

The proof of \Cref{prop:eta-obj-nonconvex} is provided in \Cref{app:prop:eta-obj-nonconvex-proof}. Note that for each fixed policy $\pi$, the objective in \eqref{eq:obj-path-risk} is convex in $\eta$.
However, \textit{minimizing over $\pi$ does not generally preserve this convexity}. We provide a simple sufficient condition where the resulting $G_{\lambda}(\eta)$ is non-convex in $\eta$ towards the end of \Cref{app:prop:eta-obj-nonconvex-proof}. 

While \eqref{eq:eta-convexity-condition-general-lambda} characterizes the sufficient and necessary condition for convexity, verifying it requires
knowledge of the cost density $h$ and the corresponding
tilted expectations at every threshold $\eta$. 
One way to establish the convexity of $G_\lambda$ is to impose additional structural conditions, avoiding the need to directly verify \eqref{eq:eta-convexity-condition-general-lambda}.
The following proposition provides such an example.

\begin{proposition}[Convexity of $G_\lambda$]
\label{prop:eta-convexity}
    Suppose that the pretrained terminal density $q_0$ is log-concave and that $c(x_0)$ is convex and bounded on the support of $q_0$. Then, the objective function $G_\lambda(\eta)$ in \eqref{eq:obj-eta} is convex in $\eta$. 
\end{proposition}

The proof is in \Cref{app:prop:eta-convexity-proof}.
Note that the stated assumption is one sufficient condition for convexity, and other simple sufficient conditions may also be possible. 

\subsection{Quadratic Regularization}

Although $G_\lambda$ is nonconvex in general, its negative curvature is uniformly bounded by $\frac{\lambda^2}{4\alpha(1-\beta)^2}$, as shown in \Cref{prop:eta-obj-nonconvex}. In other words, $G_\lambda$ is $\frac{\lambda^2}{4\alpha(1-\beta)^2}$-weakly convex. This motivates a quadratically regularized threshold objective, defined as follows:
\begin{equation}\label{eq:regularized-obj-eta}
    \min_{\eta \in \mbR}
    G_{\lambda,\kappa}(\eta)
    :=G_\lambda(\eta)
    +\frac{\kappa}{2}(\eta-\eta_c)^2,
\end{equation}
where $\eta_c$ is an arbitrary center and $\kappa\geq0$ controls the regularization strength.
For the most general case, \Cref{prop:eta-obj-nonconvex} implies that $G_{\lambda,\kappa}$ is strongly convex with parameter $\kappa-\frac{\lambda^2}{4\alpha(1-\beta)^2}$
whenever
$\kappa>\frac{\lambda^2}{4\alpha(1-\beta)^2}$.
If $G_\lambda$ is already convex (as shown in \Cref{prop:eta-convexity}), any $\kappa>0$ instead yields $\kappa$-strong convexity. Furthermore, as we show in
\Cref{cor:risk-averse-reduction-regularized} in \Cref{appendix:regularized-reduction}, for any $\kappa\ge 0$ and $\eta_c\in\mathbb{R}$, the quadratically regularized version of \eqref{eq:obj-path-risk} can also be 
recast as the minimization of $G_{\lambda,\kappa}(\eta)$.

Let $\eta^*$ denote any minimizer of $G_\lambda$ (assuming it exists), and let $\eta_\kappa^*$ denote the unique minimizer of \eqref{eq:regularized-obj-eta} under the corresponding strong-convexity condition.
Optimality of $\eta_\kappa^*$ yields
\begin{align*}
    0 &\leq G_\lambda(\eta_\kappa^*)-G_\lambda(\eta^*)\\
    &\leq \frac{\kappa}{2} \left[(\eta^*-\eta_c)^2-(\eta_\kappa^*-\eta_c)^2\right] \leq \frac{\kappa}{2}(\eta^*-\eta_c)^2.
\end{align*}
Hence, the regularization bias in objective value vanishes
as $\kappa\to0$. 
Thus, in the remainder of this paper, we focus on solving the regularized objective in \eqref{eq:regularized-obj-eta}.

\subsection{Ternary Search with Inexact Objective Evaluation Oracle}

We next consider how to compute $\eta_\kappa^*$. According to \Cref{prop:eta-obj-nonconvex}, the first-order optimality condition of \eqref{eq:regularized-obj-eta} gives $G_{\lambda,\kappa}'(\eta)=\lambda-\frac{\lambda}{1-\beta} \cdot    \frac{B_\lambda(\eta)}{Z_\lambda(\eta)}+\kappa(\eta-\eta_c)=0$, which generally does not yield a closed-form solution.
We therefore turn to a derivative-free one-dimensional search that relies only on inexact estimates of the objective. To specify a finite search interval,
we impose the following regularity condition.

\begin{assumption}\label{assum:bounded-cost}
The cost function $c$ 
satisfies $c(x_0)\in[l,u]$ on the support of $q_0$, where $-\infty<l<u<\infty$. 
\end{assumption}

Such a bounded-cost assumption also appears in sampling-error
analyses, such as \cite{madhow2026lcb}. Under \Cref{assum:bounded-cost}, if $\eta_c\in [l,u]$, then $G_{\lambda,\kappa}'(\eta)$ is negative for $\eta<l$ and positive for $\eta>u$. So, $\eta_{\kappa}^*\in [l,u]$ and the search can be restricted to $[l,u]$ without loss of optimality.

\begin{algorithm2e}
\caption{CVaR-Optimizer 
}

\label{alg:onedopt}

\SetAlgoLined

\textbf{Input: }Interval $[l,u]\subset\mathbb{R}$, target error $\epsilon$, Lipschitz bound $L>0$

\textbf{Input}: Inexact objective evaluation oracle $\widehat{G}_{\lambda,\kappa}(\cdot):\mathbb{R}\to \mathbb{R}$

Let $\eta^{(0)}=l,$ $y_{l}^{(0)}=l,$ $y_{u}^{(0)}=u$ 

\For{$i=1,...,\left\lceil \log_{3/2}(\frac{L(u-l)}{\epsilon})\right\rceil $
}{

Let $z_{l}^{(i)}=\frac{2y_{l}^{(i-1)}+y_{u}^{(i-1)}}{3}$, $z_{u}^{(i)}=\frac{y_{l}^{(i-1)}+2y_{u}^{(i-1)}}{3}$

\uIf{ $\widehat{G}_{\lambda,\kappa}(z_{l}^{(i)})\leq \widehat{G}_{\lambda,\kappa}(z_{u}^{(i)})$ }{

Let $(y_{l}^{(i)},y_{u}^{(i)})=(y_{l}^{(i-1)},z_{u}^{(i)})$.

If $\widehat{G}_{\lambda,\kappa}(z_{l}^{(i)})\leq \widehat{G}_{\lambda,\kappa}(\eta^{(i-1)})$ update $\eta^{(i)}=z_{l}^{(i)}$; else update $\eta^{(i)}=\eta^{(i-1)}$

}\ElseIf{$\widehat{G}_{\lambda,\kappa}(z_{l}^{(i)})>\widehat{G}_{\lambda,\kappa}(z_{u}^{(i)})$}{

Let $(y_{l}^{(i)},y_{u}^{(i)})=(z_{l}^{(i)},y_{u}^{(i-1)})$.

If $\widehat{G}_{\lambda,\kappa}(z_{u}^{(i)})\leq \widehat{G}_{\lambda,\kappa}(\eta^{(i-1)})$ update $\eta^{(i)}=z_{u}^{(i)}$; else update $\eta^{(i)}=\eta^{(i-1)}$}}\textbf{Output}:
$\hat{\eta}:=\eta^{(\text{last})}$

\end{algorithm2e}

Under \Cref{assum:bounded-cost} and the strong-convexity
condition on $G_{\lambda,\kappa}$, we employ ternary search
\cite{cohen2016geometric} with an inexact objective oracle
$\widehat{G}_{\lambda,\kappa}$.
The procedure is summarized in \Cref{alg:onedopt}.
Specifically, the search starts from the full range $[l,u]$. At each iteration, the algorithm partitions the current interval into thirds by placing two interior points, $z_l$ at one-third of the way and $z_u$ at two-thirds, and queries the inexact oracle to evaluate the objective at these two points. 
Now, if $G_{\lambda,\kappa}(z_l)\le G_{\lambda,\kappa}(z_u)$, then for a convex function the minimizer cannot lie to the right of $z_u$. As a result, the algorithm discards the right third and keeps $[y_l,z_u]$.
With the inexact oracle $\widehat G_{\lambda,\kappa}$, the resulting objective error can still be controlled with a uniform bound on the oracle error (see \Cref{lem:onedimmin}).
The algorithm then compares $z_l$ with the incumbent and makes $z_l$ as the new incumbent if it is as least as good. Otherwise, the algorithm discards the left third and keeps $[z_l,y_u]$. Each iteration shrinks the interval by a factor of 2/3 and after $\left\lceil \log_{3/2}(\frac{L(u-l)}{\epsilon})\right\rceil$ iterations, its length is at most $\epsilon/L$.
We will establish the iteration and sample complexity guarantees of this derivative-free ternary search algorithm in \Cref{sec:optimization-guarantees}. Based on the numerical results in \Cref{sec:experiment}, \Cref{alg:onedopt} is much more efficient than grid search and the gradient descent algorithm.

%% file: theory.tex
\section{Theoretical Guarantees}

In this section, we establish theoretical guarantees
for the sampler in \Cref{alg:cvar-svdd}.
We first analyze the error of the threshold optimization in \Cref{alg:onedopt} (\Cref{sec:optimization-guarantees}), then the value estimation error (\Cref{sec:value-estimation}), and finally combine them into an end-to-end total variation bound on the output distribution (\Cref{sec:sampling-guarantee}).

\subsection{Threshold Optimization Guarantees}
\label{sec:optimization-guarantees}

We first quantify the accuracy of the threshold returned by \Cref{alg:onedopt}. Since the regularized objective \eqref{eq:regularized-obj-eta} involves an expectation over the pretrained distribution $q_0$, we evaluate it using a Monte Carlo estimator based on i.i.d. samples from $q_0$.
Specifically, given a single batch of i.i.d. samples
$\{x_0^i\}_{i=1}^{N}\sim q_0$, define an inexact objective evaluation oracle $\widehat G_{\lambda,\kappa,N}(\cdot):\mathbb{R} \to\mathbb{R}$ as
\begin{align*}
    \widehat G_{\lambda,\kappa,N}(\eta)
    :=-\alpha\log\Big(
    \frac{1}{N}\sum_{i=1}^{N}
    e^{-c_{\lambda,\eta}(x_0^i)/\alpha}
    \Big)+\frac{\kappa}{2}(\eta-\eta_c)^2.
\end{align*}

The following theorem states a finite-sample guarantee that \Cref{alg:onedopt} produces a near-optimal solution to \eqref{eq:regularized-obj-eta} when using the Monte Carlo oracle $\widehat G_{\lambda,\kappa,N}(\eta)$.
\begin{theorem}[Finite-sample guarantees for threshold optimization]
\label{thm:eta-optimization}
Suppose \Cref{assum:bounded-cost} holds.
Let $\eta_c \in [l,u]$ be arbitrary, and suppose that
$G_{\lambda,\kappa}$ is $m_\kappa$-strongly convex on
$[l,u]$ for some $m_\kappa>0$.
Define
\begin{equation}
\label{eq:obj-L-Lkappa-def}
    L := \lambda\max\left\{1,\frac{\beta}{1-\beta}\right\},\quad L_\kappa := L + \kappa (u-l).
\end{equation}
For any $\delta\in(0,1)$ and sufficiently small
$\eps_\eta>0$, define
\[
  K:=\Bigl\lceil\log_{3/2}\Bigl(\frac{8L_\kappa(u-l)}{m_\kappa\eps_\eta^2}\Bigr)\Bigr\rceil,
  \ 
  J:=\Bigl\lceil\frac{32L(u-l)}{m_\kappa\eps_\eta^2}\Bigr\rceil+1.
\]

Then, with probability at least $1-\delta$, the output $\hat\eta$ of \Cref{alg:onedopt} (with Lipschitz bound $L_\kappa$ and target error $m_\kappa \eps_\eta^2/8$) satisfies
$|\hat\eta-\eta_\kappa^*|\le\eps_\eta$
as long as
\[
  N\ \ge\ \frac{4\,C_G}{m_\kappa^2\,\eps_\eta^4}\log\frac{2J}{\delta}
\]
where $C_G>0$ is a constant depending on
$\alpha,\beta,\lambda,l,u$. Furthermore, the algorithm calls the Monte Carlo oracle $\widehat G_{\lambda,\kappa,N}$ at most $1+2K$ times. 
\end{theorem}

The proof is in \Cref{app:thm:eta-optimization-proof}. We first use Bernstein's inequality to obtain a pointwise bound on the error of the inexact objective oracle, and then discretize the $\eta$-space to turn this pointwise bound into a uniform one using a union bound, which yields the final sample complexity. The oracle round (iteration) complexity is $O(\log\frac{L_\kappa(u-l)}{m_\kappa\eps_\eta^2})$, and the
sample complexity is
$O\bigl(\frac{1}{m_\kappa^2\eps_\eta^4}\log\frac{L(u-l)}{m_\kappa\eps_\eta^2\delta}\bigr)$. Note that these samples are drawn once and reused for all queried points in \Cref{alg:onedopt}. 

\subsection{Value Estimation Guarantees}\label{sec:value-estimation}

Having established the complexity of the threshold optimization problem, we next proceed to the sample complexity of the value estimates used by the SVDD sampler. Different from \Cref{sec:optimization-guarantees} where $G_{\lambda,\kappa}(\eta)$ is estimated at a fixed terminal state by averaging i.i.d. samples from $q_0$, the value functions $V^*_{t,\eta}$ will be estimated by regressing on paired samples $\cbrc{(x_t^i,x_0^i)}_{i=1}^N$.

For each $t=1,\dots,T$, suppose that $\cbrc{(x_t^i,x_0^i)}_{i=1}^N \overset{\mathrm{i.i.d.}}{\sim} q_t(x_t) q_{0|t}(x_0\mid x_t)$. Recall $c_{\lambda,\eta}$ from \eqref{eq:weighted-cost-def} and $V_{t,\eta}^*$ from \eqref{eq:Vt-eta-star-def}. Define the (normalized) target as $Y_\eta(x_0) := e^{-\frac{c_{\lambda,\eta}(x_0)- a_\eta}{\alpha}}$ where $a_\eta := (1-\lambda) l + \lambda \eta$ is a lower bound on $c_{\lambda,\eta}(x_0)$. Let 
\begin{equation}
\label{eq:overline-U-def}
    \xi:= \frac{2(u-l)}{\alpha}\left(1-\lambda+\frac{\lambda}{1-\beta}\right).
\end{equation}
Then $Y_{\eta}$ takes values in $[e^{-\xi/2},1]$. Note that $V_{t,\eta}^* = a_\eta - \alpha \log \E_{q_{0|t}} Y_{\eta}(x_0)$.

Let $\mathcal F_t$ be the class of linear regressors with $F$
fixed features bounded by one, coefficients in $[-1,1]$, and
outputs clipped to $[e^{-\xi/2},1]$. For convenience of analysis, suppose that the ground-truth regression function $\E[Y_\eta(X_0)\mid X_t]\in\mathcal F_t$ for every $t,\eta$, and we have access to an approximate empirical risk minimizer (ERM) $\widehat f_{t,\eta}\in\mathcal F_t$, from which we construct a value function estimate $\widehat V_{t,\eta,N}$ for $V_{t,\eta}^*$.

The following theorem provides the sample complexity to establish an error bound between $\widehat V_{t,\eta,N}$ and $V_{t,\eta}^*$ that holds uniformly over $\eta \in [l,u]$ and $t=1,\dots,T$ (and thus any $\eta$ returned by \Cref{alg:onedopt}).

\begin{theorem}[Value estimation from paired samples]
\label{thm:paired-value-estimation}
Suppose that \Cref{assum:bounded-cost} holds.
Define the loss $\widehat{\mathcal L}_{t,\eta}(g)
    :=\frac1N\sum_{i=1}^N
    \brc{g(x_t^i)-Y_\eta(x_0^i)}^2$. For any $\delta\in(0,1)$ and sufficiently small
$\eps_v>0$, suppose that we have access to an approximate ERM $\widehat f_{t,\eta}\in\mathcal F_t$ such that, for all $t=1,\dots,T,~ \eta \in [l,u]$,
\[
    \widehat{\mathcal L}_{t,\eta}(\widehat f_{t,\eta})
    \leq \min_{g\in\mathcal F_t} \widehat{\mathcal L}_{t,\eta}(g) + \frac{\varepsilon_v^2}{8(1+\alpha^2)}e^{-\xi}
\]
Now construct value function estimates with the approximate ERM $\widehat f_{t,\eta}$ as follows:
\begin{align}
    \widehat V_{t,\eta,N}
    := a_\eta-\alpha\log\widehat f_{t,\eta},\ \forall t=1,\dots,T.\label{eq:value-function-estimate}
\end{align}
Then, with probability at least $1-\delta$, we have $\sup_{\eta\in[l,u]} \max_{t \in \{1,\dots,T\}} \norm{\widehat V_{t,\eta,N}-V_{t,\eta}^*}_{L_2(q_t)} \le\eps_v$, as long as
\[
    N\ge
    \frac{C}{\eps_v^2}
    \log\frac{T}{\delta\eps_v},
\]
where $C>0$ is a constant depending on $\alpha,\beta,\lambda,l,u,F$.
\end{theorem}

The proof is in \Cref{app:thm:paired-value-estimation-proof}. Similarly, we first use Bernstein's inequality to obtain a pointwise bound on the difference between the population excess loss and its empirical counterpart for a fixed predictor $g\in\mathcal{F}_t$ and a fixed threshold $\eta$. Then, we discretize both the function class $\mathcal{F}_t$ and the $\eta$-space and take a union bound to turn this pointwise bound into a uniform one. Together with the realizability assumption and the near-optimality of the ERM, this yields the final sample complexity. Note that \Cref{alg:cvar-svdd}, \Cref{alg:onedopt} and their theoretical guarantees can also be applied to risk-neutral inference-time alignment by setting $\lambda=0$. 

\subsection{End-to-End Sampling Guarantees}\label{sec:sampling-guarantee}

We now combine the threshold optimization and value estimation guarantees to provide an end-to-end guarantee for the sampling process. Specifically, for any $\eta$, let $p^*_\eta$ and $\hat{p}_\eta$ denote the terminal marginal distributions induced by the exact and estimated value functions, respectively. We provide a high-probability bound (over the training samples) for the total variation distance between the sampler's output distribution $\hat{p}_{\hat \eta}$ (with clipped value estimates) and the regularized target $p_{\eta_\kappa^*}^*$.

\begin{theorem} \label{thm:sampling}
Suppose that the assumptions of \Cref{thm:eta-optimization,thm:paired-value-estimation} hold.
Recall from \eqref{eq:overline-U-def} that
    $\xi:= \frac{2(u-l)}{\alpha}\left(1-\lambda+\frac{\lambda}{1-\beta}\right)$.
Then, under exact tilted-transition sampling, using \Cref{alg:onedopt} with the hyperparameters chosen as in \Cref{thm:eta-optimization}, and using the value estimator according to \eqref{eq:value-function-estimate}, we have, with probability at least $1-\delta$,
\[ \TV{p_{\eta_\kappa^*}^* (x_0)}{\hat{p}_{\hat \eta}(x_0)} \leq \frac{e^{\xi/2}}{2 \alpha} \sqrt{ T } \eps_v + \frac{\lambda \eps_\eta}{4 \alpha(1-\beta)}. \]
\end{theorem}

The proof is in \Cref{app:thm:sampling-proof}. The two terms in \Cref{thm:sampling} quantify distinct sources of error. The first arises from approximating the value functions used to tilt the reverse transitions (\Cref{thm:paired-value-estimation}). These errors accumulate across the denoising steps, yielding the $\sqrt{T}\eps_v$ dependence in TV distance. The second arises from using the estimated threshold $\hat\eta$ instead of $\eta_\kappa^*$ (\Cref{thm:eta-optimization}). It controls the change in the terminal tilted distribution and is independent of $T$. Thus, for fixed problem parameters and $T$, both contributions vanish as the training sample size increases.

%% file: experiment.tex
\section{Numerical Experiments}
\label{sec:experiment}

We numerically validate our risk-averse inference-time alignment framework, including the sampling (\Cref{alg:cvar-svdd}) and threshold optimization (\Cref{alg:onedopt}) procedures, on synthetic data in \Cref{sec:synthetic-main} and on real-world image data in \Cref{sec:image-main}, respectively.

\subsection{Synthetic Data}
\label{sec:synthetic-main}

We first evaluate whether CVaR risk-control reduces high-cost outcomes on a two-dimensional mixture of rotated product-Cauchy distributions (see \eqref{eq:mixed-cauchy-data}). This heavy-tailed example is evaluated using a bounded quadratic terminal cost (see \eqref{eq:mixed-cauchy-cost}).
We compare our risk-averse CVaR-SVDD sampler against two baselines: (i) the frozen base diffusion sampler, without any inference-time guidance, and (ii) vanilla risk-neutral SVDD, with $\lambda=0$. 
For CVaR-SVDD, we use the pure tail-risk objective with $\lambda=1$, with the same hyperparameters as in vanilla SVDD whenever applicable. 

We set the CVaR confidence level to $\beta=0.8$, the temperature to $\alpha=1$, and the regularization strength to $\kappa=0.02$. For CVaR-SVDD, we optimize the threshold $\eta$ using \Cref{alg:onedopt}. 
For value function estimation, we compare linear regressors with neural networks in \Cref{fig:value-estimator-mixed-cauchy}. Preliminary results show that neural-network value estimators produce fewer high-cost outcomes, so we adopt neural networks to estimate value functions in the remaining experiments. 
Additional experimental setups are provided in \Cref{app:synthetic-setup}.

\begin{figure}[ht!]
\centering
    \vspace{0pt}
    \centering
    \begin{subfigure}[t]{0.45\linewidth}
        \centering
        \includegraphics[width=0.9\linewidth]{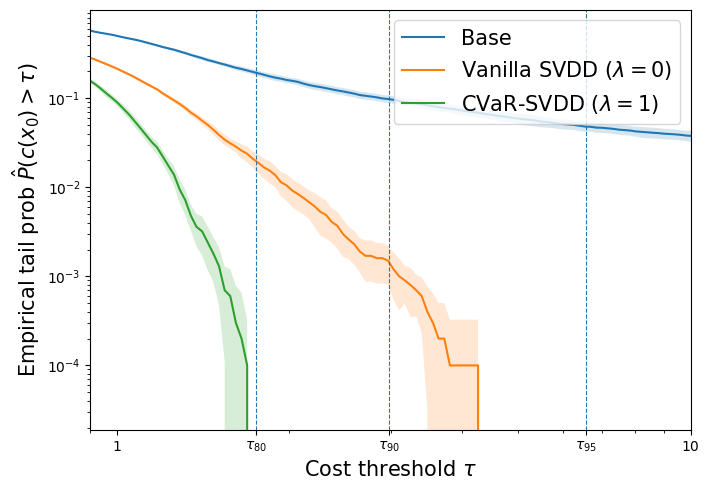}
        \caption{Empirical cost survival functions.}
        \label{fig:synthetic-survival}
    \end{subfigure}
    \hfill
    \begin{subfigure}[t]{0.45\linewidth}
        \centering
        \includegraphics[width=0.9\linewidth]{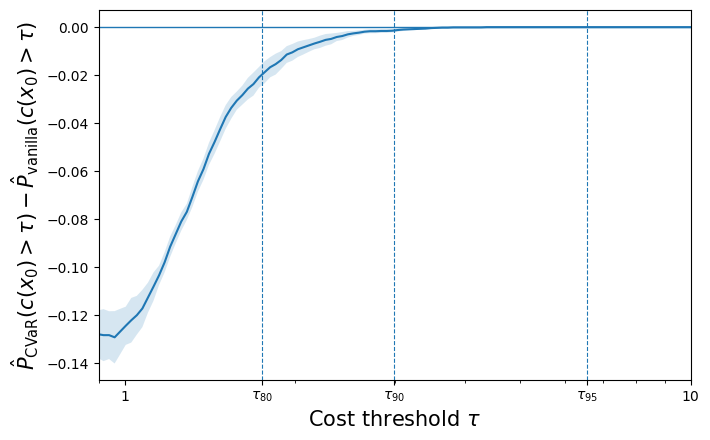}
        \caption{Paired tail-probability differences (CVaR-SVDD minus vanilla SVDD).}
        \label{fig:synthetic-diff}
    \end{subfigure}
    \caption{
    Synthetic mixed-Cauchy results. 
    Vertical lines mark the $80$-th, $90$-th, and $95$-th cost percentiles
    estimated from $50000$ base-model samples.
    Shading indicates pointwise $95\%$ confidence intervals.
    }
    \label{fig:synthetic-main}
\end{figure}

\paragraph{Results.}
\Cref{tab:synthetic-main} shows that CVaR-SVDD improves empirical tail performance. Compared with vanilla SVDD, it reduces empirical CVaR from $1.343$ to $1.017$ and lowers the fraction of samples with costs exceeding $\tau_{80}$, which is the empirical $80$-th percentile estimated from $50000$ samples from the base model. The survival curves in \Cref{fig:synthetic-main} further show lower tail probabilities across a broad range of cost thresholds. 

Interestingly, \Cref{tab:synthetic-main} shows that CVaR-SVDD also reduces the mean cost. This observation is consistent with CVaR penalization shifting probability mass away from high-cost regions, as illustrated in the top panel of \Cref{fig:mixed-cauchy}. However, improvement in the mean depends on the underlying distribution and cost function and is not guaranteed by the CVaR objective.
We also observe a similar improvement in the average reward in the real-image experiments in \Cref{sec:image-main}.

\begin{table}[t]
\centering
\caption{
Sampling performance on synthetic Cauchy data. Entries are mean $\pm$ standard error across ten seeds. All metrics are \textbf{costs}, so smaller values are better. ``0 observed'' indicates no sample exceeded $\tau_{80}$.
}
\label{tab:synthetic-main}
\footnotesize
\setlength{\tabcolsep}{3pt}
\renewcommand{\arraystretch}{1.15}
\begin{tabular}{@{}lccc@{}}
\toprule
Method
& Mean Cost $\downarrow$
& \makecell{Empirical\\CVaR $\downarrow$}
& \makecell{$\hat{\mbP}(C>\tau_{80})$\\(\%) $\downarrow$} \\
\midrule
Base
& $2.373 \pm 0.109$
& $8.379 \pm 0.526$
& $19.37 \pm 0.62$ \\
SVDD \cite{li2025svdd}
& $0.706 \pm 0.003$
& $1.343 \pm 0.011$
& $1.95 \pm 0.20$ \\
CVaR-SVDD
& $\mathbf{0.587 \pm 0.002}$
& $\mathbf{1.017 \pm 0.006}$
& \textbf{0 observed} \\
\bottomrule
\end{tabular}
\end{table}

\paragraph{Ablations.}
Additional ablation results are provided in
\Cref{app:synthetic-ablations}. \underline{\textit{(1) Inference-time sampler:}} The reduction in high-cost outcomes persists when SVDD is replaced by SMC (see \Cref{fig:smc-ablation-mixed-cauchy}). \underline{\textit{(2) Temperature $\alpha$:}} At every tested temperature $\alpha\in\{0.5,1,5,10\}$, CVaR-SVDD yields a lower empirical tail probability at $\tau_{80}$ than vanilla SVDD (see \Cref{fig:alpha-ablation-mixed-cauchy}). It also achieves a more favorable trade-off between tail performance and base log-likelihood. \underline{\textit{(3) Risk parameter $\lambda$:}} Increasing $\lambda\in\{0,0.25,0.5,0.75,1\}$ progressively reduces empirical CVaR and tail probabilities, illustrating the gradual transition from risk-neutral to pure-CVaR risk control (see \Cref{tab:lambda-ablation}). \underline{\textit{(4) Regularization strength $\kappa$:}} We vary $\kappa$ from $0$ to $2$ and observe nearly identical tail performance when $\kappa \in [0,0.2]$ (see \Cref{fig:kappa-ablation}). As a result, we fix $\kappa=0.02$ in our default setting.

\begin{figure}[ht!]
    \vspace{0pt}
    \centering
    \includegraphics[width=0.45\linewidth]
        {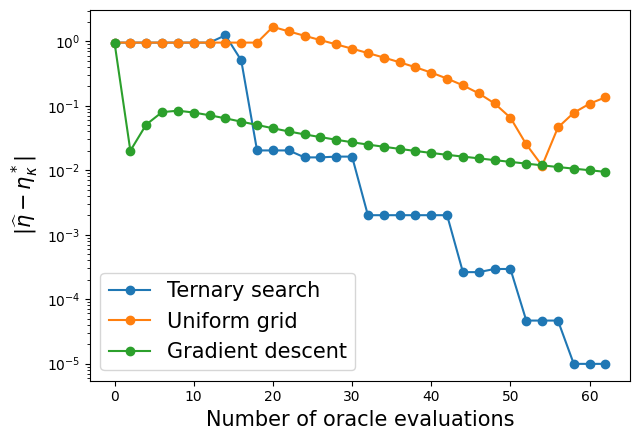}
    \caption{
    Absolute threshold error
    $|\widehat{\eta}-\eta_\kappa^\star|$
    versus the number of oracle evaluations.
    Gradient descent results are averaged over five
    randomly sampled learning rates.
    }
    \label{fig:synthetic-eta-main}
\end{figure}

\paragraph{Threshold optimization.} 
\Cref{fig:synthetic-eta-main} compares \Cref{alg:onedopt} with uniform-grid search and projected gradient descent. When the number of oracle evaluations increases, \Cref{alg:onedopt} achieves substantially smaller threshold errors than both baselines, demonstrating its effectiveness (see \Cref{fig:eta-optimizer-ablation} for more details).

\subsection{Image Data}
\label{sec:image-main}

We evaluate risk-averse inference-time sampling for text-to-image generation using a frozen Stable Diffusion v1.5 model \cite{rombach2022stable-diffusion}. We consider two terminal rewards: compressibility and aesthetic score \cite{wang2025finetuning}. In both cases, we set the terminal cost to $c(x_0)=-r(x_0)$. We again compare the base diffusion sampler, SVDD-MC \cite{li2025svdd}, and our CVaR-SVDD with a learned hinge-type value function. 
See \Cref{app:image-setup} for experimental details.

\begin{table}[t]
\centering
\caption{Image generation for the \texttt{cat} prompt. Entries are mean $\pm$ standard error across five seeds. Both metrics are \textbf{rewards}, so larger values are better.}
\label{tab:image-main}
\small
\setlength{\tabcolsep}{3pt}
\begin{tabular}{@{}lrr@{}}
\toprule
Method & Mean Reward $\uparrow$ & Avg Tail Reward $\uparrow$ \\
\midrule
\multicolumn{3}{l}{Compressibility} \\
Base & $-95.986\pm0.224$ & $-123.377\pm1.042$ \\
SVDD \cite{li2025svdd} & $-73.034\pm0.121$ & $-89.671\pm0.292$ \\
CVaR-SVDD & $\mathbf{-67.136\pm0.131}$ & $\mathbf{-78.405\pm0.135}$ \\
\midrule
\multicolumn{3}{l}{Aesthetic score} \\
Base & $5.8845\pm0.0076$ & $5.4600\pm0.0105$ \\
SVDD \cite{li2025svdd} & $5.9165\pm0.0036$ & $5.5154\pm0.0037$ \\
CVaR-SVDD & $\mathbf{5.9285\pm0.0041}$ & $\mathbf{5.5341\pm0.0043}$ \\
\bottomrule
\end{tabular}
\vspace{-2mm}
\end{table}

\paragraph{Results.}
\Cref{tab:image-main} shows that CVaR-SVDD improves both the
mean and lower-tail compressibility reward for \texttt{cat}.
The reward distributions and CDFs in \Cref{fig:image-cat-compressibility} further illustrate the reduction in low-reward outcomes. For aesthetics with \texttt{cat}, CVaR-SVDD increases the lower-tail reward while slightly improving the mean. 
The overall aesthetic distributions remain close, with the modest lower-tail gains most apparent in the enlarged CDF in \Cref{fig:image-cat-aesthetic}.

\paragraph{Ablations.} 
Additional ablation results are reported in \Cref{tab:image-full} and \Cref{app:image-results}. \underline{\textit{(1) Prompt:}} The compressibility gains persist for the \texttt{dog} and \texttt{rabbit} prompts. \underline{\textit{(2) Inference-time sampler:}} CVaR-SMC also yields a slightly higher observed lower-tail mean. 
\Cref{app:image-tail-examples} further shows randomly selected generated images from the worst-$20\%$ tail.

%% file: conclusion.tex
\section{Conclusions}

In this work, we developed a risk-averse inference-time diffusion alignment framework by combining CVaR and expected cost. By characterizing the optimal sampling policies, we reduced the alignment problem to scalar threshold optimization. Our finite-sample analysis quantifies how errors in threshold optimization and value estimation affect the final output distribution.
Synthetic and image-generation experiments
further demonstrate improved empirical tail control. Future work includes extending the guarantees to unbounded costs and explicitly accounting for the finite numbers of proposals or particles used by practical samplers.

%% file: appendix.tex
\section{Additional Related Work}
\label{app:additional-related-works}

We have provided the most relevant works in \Cref{sec:related-works}. As follows, we provide all the additional related work.

\paragraph{Reward-based fine-tuning.}
One way to align diffusion models for downstream task is reward-based fine-tuning, which adapts pretrained diffusion models by updating their model weights to improve downstream objectives (see \cite{uehara2024survey-finetuning} for a survey).
For example, Denoising Diffusion Policy Optimization (DDPO) \cite{black2024ddpo} formulates denoising as a sequential decision-making problem and applies policy-gradient methods to optimize downstream rewards, and
Direct Reward Fine-Tuning (DRaFT) \cite{clark2024draft} backpropagates differentiable rewards through the sampling procedure, with variants that truncate backpropagation to reduce computation.
These approaches incorporate alignment into the generative model's parameters. 

\paragraph{Inference-time alignment.}
Different from fine-tuning, inference-time alignment methods steer the diffusion generation process toward optimizing downstream objectives while keeping the pretrained generative model frozen (see \cite{uehara2025survey-alignment} for a survey).
For example, Soft Value-Based Decoding (SVDD) \cite{li2025svdd} uses soft value estimates to select intermediate candidates based on their expected future desirability, and Feynman-Kac steering \cite{singhal2025fksteering} provides a particle-based framework that guides generation through intermediate weighting and resampling.
Their standard formulations target reward-tilted distributions without explicitly optimizing the CVaR of the output distribution.
In contrast, our framework incorporates the mean-CVaR interpolated objective into this sampling perspective through a threshold-dependent terminal cost, enabling CVaR extensions of existing inference-time samplers.

\paragraph{Inference-time conditional and constrained sampling.}
Diffusion alignment is closely related to conditional and constrained sampling, for which numerous inference-time methods have been developed.
These include methods for zero-shot conditional sampling \cite{chung2023dps,liang2025conditional,song2023pgdm,wu2023tds,zhang2025daps}.
A related line incorporates explicit feasibility constraints into the sampling process \cite{christopher2024constrained,narasimhan2025constrained}.
Indeed, conditional sampling can be formulated as reward-guided alignment by using the conditional log-likelihood as the reward \cite{uehara2025survey-alignment}.
This connection suggests potential applications of our mean-CVaR framework to controlling the upper tail of negative log-likelihoods or constraint-violation costs.

\section{Vanilla Inference-time Alignment Algorithms}
\label{sec:inf-align-alg-reference}

In this section, we review two risk-neutral inference-time alignment algorithms: the SVDD (\Cref{alg:svdd}) and the SMC (\Cref{alg:smc}) sampling algorithm. Both of them can be used within our CVaR framework.

\begin{algorithm2e}
\caption{$\mathtt{SVDD}$ Sampling \cite{li2025svdd}}
\label{alg:svdd}

\SetAlgoLined
\DontPrintSemicolon

\textbf{Input:} Pretrained reverse kernels
$\{\mu_t(x_{t-1}\mid x_t)\}_{t=1}^{T+1}$, 
value estimates $\{\widehat V_{t}\}_{t=0}^{T}$ (with $\widehat V_{0}(\cdot) := c(\cdot)$),
number of samples $N$,
number of candidates $M$,
temperature $\alpha>0$\;

Obtain $N$ initial dummy variables $x_{T+1}^{(1)},\ldots,x_{T+1}^{(N)}$\;

\For{$t=T+1,T,\ldots,1$}{
    \For{$i=1,\ldots,N$}{
        Sample $M$ candidates independently: $z_{t,m}^{(i)}
            \sim \mu_t(\cdot\mid x_t^{(i)}),
            \qquad m=1,\ldots,M$.

        Calculate the normalized weights:
        \[
            \bar w_{t,m}^{(i)}
            =
            \frac{
                \exp\bigl(
                -\widehat V_{t-1}(z_{t,m}^{(i)})/\alpha
                \bigr)
            }{
                \sum_{j=1}^{M}
                \exp\bigl(
                -\widehat V_{t-1}(z_{t,j}^{(i)})/\alpha
                \bigr)
            },
            \quad m=1,\ldots,M.
        \]

        Sample
        $a_t^{(i)}\sim
        \mathrm{Categorical}
        (\bar w_{t,1}^{(i)},\ldots,\bar w_{t,M}^{(i)})$\;

        Set $x_{t-1}^{(i)}=z_{t,a_t^{(i)}}^{(i)}$\;
    }
}

\textbf{Output:} $\{x_0^{(i)}\}_{i=1}^{N}$\;

\end{algorithm2e}

\allowdisplaybreaks
\begin{algorithm2e}
\caption{$\mathtt{SMC}$ Sampling \cite{singhal2025fksteering}}
\label{alg:smc}

\SetAlgoLined

\textbf{Input:} Pretrained reverse kernels 
$\{\mu_t(x_{t-1}\mid x_t)\}_{t=1}^{T+1}$, 
value estimates
$\{\widehat V_{t}\}_{t=0}^{T+1}$ (with $\widehat V_{0}(\cdot) := c(\cdot)$),
number of particles $N$, 
temperature $\alpha>0$, 
ESS threshold $\gamma\in(0,1]$

Obtain $N$ initial dummy variables $x_{T+1}^{(1)},\ldots,x_{T+1}^{(N)}$\;

Let $\ell^{(i)}=0$ and 
$\phi^{(i)} =
- \widehat V_{T+1}/\alpha$
for all $i=1,\ldots,N$\;

\For{$t=T+1,T,\ldots,1$}{

Sample $\widetilde x_{t-1}^{(i)} \sim \mu_t(\cdot\mid x_t^{(i)}),~\forall i = 1,\dots,N$.

\For{$i=1,...,N$}{
Define $\widetilde \phi^{(i)} := \frac{ - \widehat V_{t-1} (\widetilde x_{t-1}^{(i)}) }{ \alpha }$

Update the accumulated log-weight: $\ell^{(i)} = \ell^{(i)} + \widetilde \phi^{(i)} - \phi^{(i)}$
}

Calculate weight $\bar w^{(i)} = \frac{ \exp(\ell^{(i)}) }{ \sum_{j=1}^{N}\exp(\ell^{(j)}) }$ for each $i = 1,\dots,N$ and $\mathrm{ESS} =\frac{1}{ \sum_{i=1}^{N} (\bar w^{(i)})^2 }$

\uIf{$t>1$ and $\mathrm{ESS} <\gamma N$}{

Sample ancestor indices
$a_1,\ldots,a_N$
independently from
$\mathrm{Categorical}
(\bar w^{(1)},\ldots,\bar w^{(N)})$

Perform resampling: i.e., for each $i=1,\ldots,N$, update
$x_{t-1}^{(i)} = \widetilde x_{t-1}^{(a_i)}, \ \phi^{(i)} = \widetilde \phi^{(a_i)}$

Reset $\ell^{(i)}=0$ for all $i=1,\ldots,N$
}\Else{
Keep $x_{t-1}^{(i)} = \widetilde x_{t-1}^{(i)}$ and $\phi^{(i)} = \widetilde \phi^{(i)}$ for each $i = 1,\dots,N$.
}

}

Calculate $\bar w^{(i)}$ and perform resampling again.

\textbf{Output}: 
$\{x_0^{(i)}\}_{i=1}^{N}$

\end{algorithm2e}

\section{Proof of Theorem \texorpdfstring{\ref{thm:risk-averse-reduction}}{1}}
\label{app:thm:risk-averse-reduction-proof}

Recall the tail-controlled objective in \eqref{eq:obj-path-risk}:
\[ \min_{\pi,\eta} \cbrc{ \E_{\pi} \sbrc{c_{\lambda, \eta}(x_0) } + \alpha \KL{\pi_{1:T+1}}{\mu_{1:T+1}} }. \]
For a fixed $\eta$, the integrability condition guarantees all expectations below are finite. Fix $t \in \{1,\dots,T+1\}$. Notice that the inner problem has the same entropy-regularized objective form as \eqref{eq:obj-path-neutral}.
Recall the variational definition of the
risk-adjusted value function:
\begin{equation}
\label{eq:value-tail-fixed-eta}
    V_{t,\eta}^*(x_t) := \min_{\pi_{1:t}} \E_{\pi} \sbrc{ c_{\lambda, \eta}(x_0) + \alpha \sum_{k=1}^{t}\KL{\pi_k}{\mu_k} \bigg| x_t}.
\end{equation}
We can also define the corresponding Q-function:
\begin{equation*}
\label{eq:q-tail-fixed-eta}
Q_{t,\eta}^*(x_t,x_{t-1}) := \min_{\pi_{1:t-1}} \E_{\pi} \sbrc{ c_{\lambda, \eta}(x_0) + \alpha \sum_{k=1}^{t-1}\KL{\pi_k}{\mu_k} \bigg| x_t, x_{t-1} }. 
\end{equation*}
Once $x_{t-1}$ is fixed, the remaining optimization
depends only on $x_{t-1}$ by the Markov property. Hence,
\begin{equation*}
    Q_{t,\eta}^*(x_t,x_{t-1})
    =
    V_{t-1,\eta}^*(x_{t-1}).
\end{equation*}

With this setup, the optimization in \eqref{eq:value-tail-fixed-eta} becomes
\begin{equation} \label{eq:value-tail-fixed-eta-Q}
    \min_{\pi_t} \E_{\pi_t} \sbrc{ V_{t-1,\eta}^*(x_{t-1}) \bigg| x_t} + \alpha \KL{\pi_t}{\mu_t},
\end{equation}
and thus, taking the derivative and setting it to zero, we get the optimal transition at time $t$ as
\begin{equation*}
\label{eq:policy-tail-fixed-eta}
    \pi_{t,\eta}^*(x_{t-1}\mid x_t) \propto \mu_t(x_{t-1}\mid x_t) e^{-Q_{t,\eta}^*(x_t,x_{t-1})/\alpha} =  \mu_t(x_{t-1}\mid x_t) e^{-V_{t-1,\eta}^*(x_{t-1})/\alpha}.
\end{equation*}
Plugging $\pi_{t,\eta}^*$ back into \eqref{eq:value-tail-fixed-eta-Q}, we have the following recursion:
\begin{equation*}
\label{eq:bellman-tail-fixed-eta}
    V_{t,\eta}^*(x_t) = -\alpha \log \E_{\mu_t} \sbrc{ e^{-V_{t-1,\eta}^*(x_{t-1})/\alpha} \bigg| x_t} \iff e^{-V_{t,\eta}^*(x_{t})/\alpha} = \E_{\mu_t} \sbrc{ e^{-V_{t-1,\eta}^*(x_{t-1})/\alpha} \bigg| x_t},
\end{equation*}
with $V_{0,\eta}^*(x_0) = c_{\lambda,\eta}(x_0)$.
Consequently,
\begin{equation*}
\label{eq:partition-tail-fixed-eta}
    V_{t,\eta}^*(x_t) = -\alpha \log \E_{q_{0|t}} \sbrc{ e^{-c_{\lambda,\eta}(x_0)/\alpha} \bigg| x_t}.
\end{equation*}
The proof is now complete.

\section{Proof of Proposition \texorpdfstring{\ref{prop:eta-obj-nonconvex}}{1} and An Example for Non-Convexity}
\label{app:prop:eta-obj-nonconvex-proof}

\begin{proof}[Proof of \Cref{prop:eta-obj-nonconvex}]

When $\lambda=0$, $G_0$ is constant in $\eta$ and thus convex. Note that $A_0$ and $B_0$ are still well defined, and the criterion always holds when $\lambda = 0$. As follows, we focus on the case where $\lambda > 0$.

Recall from \eqref{eq:weighted-cost-def} that
\[ c_{\lambda,\eta}(x_0) = (1-\lambda) c(x_0) + \lambda \brc{ \eta + \frac{1}{1-\beta} (c(x_0)-\eta)^+ }. \]
Also recall the definitions:
\begin{equation*}
    A_\lambda(\eta)
    := \E_{q_0}\sbrc{
        e^{-c_{\lambda,\eta}(x_0)/\alpha}
        \mathds{1}\cbrc{c(x_0)\leq\eta}
    }, \quad
    B_\lambda(\eta)
    := \E_{q_0}\sbrc{
        e^{-c_{\lambda,\eta}(x_0)/\alpha}
        \mathds{1}\cbrc{c(x_0)>\eta}
    }.
\end{equation*}
Let
\[ Z_{\lambda}(\eta) = \E_{q_0}\sbrc{ e^{-c_{\lambda,\eta}(x_0) / \alpha} }, \qquad G_\lambda(\eta) = -\alpha \log Z_{\lambda}(\eta). \]
Writing $C:=c(x_0)$ and splitting the expectation at $C=\eta$, we have
\begin{align}
Z_{\lambda}(\eta) &= A_{\lambda}(\eta) + B_{\lambda}(\eta) \nonumber \\
&= \int_{-\infty}^{\eta} \exp\brc{ -\frac{(1-\lambda)C+\lambda\eta}{\alpha}} h(C)\,dC + \int_{\eta}^{\infty} \exp\brc{ -\frac{(1-\lambda)C +\lambda\eta +\frac{\lambda}{1-\beta}(C-\eta)}{\alpha}} h(C)\,dC.
\label{eq:Z-lambda-eta-decomposition}
\end{align}

Since $c(x_0)$ has continuous density $h$, differentiating the first term in \eqref{eq:Z-lambda-eta-decomposition} yields
\begin{align}
    &\frac{d}{d\eta} A_\lambda(\eta) \nonumber \\
    &=\frac{d}{d\eta} \E_{q_0} \sbrc{ \exp\brc{ -\frac{(1-\lambda)c(x_0)+\lambda\eta}{\alpha} } \mathds{1}\cbrc{c(x_0)\leq\eta}} \nonumber \\
    &= \E_{q_0} \sbrc{ \brc{ \exp\brc{ -\frac{(1-\lambda)c(x_0)+\lambda\eta}{\alpha}}}' \mathds{1}\cbrc{c(x_0)\leq\eta}} + \E_{q_0} \sbrc{ \exp\brc{ -\frac{(1-\lambda)c(x_0)+\lambda\eta}{\alpha} } \brc{ \mathds{1}\cbrc{c(x_0)\leq\eta} }'}
    \nonumber\\
    &=
    -\frac{\lambda}{\alpha}
    \E_{q_0}
    \sbrc{
        \exp\brc{
            -\frac{(1-\lambda)c(x_0)+\lambda\eta}{\alpha}
        }
        \mathds{1}\cbrc{c(x_0)\leq\eta}
    }
    +
    e^{-\eta/\alpha}h(\eta) \nonumber\\
    &= -\frac{\lambda}{\alpha} A_\lambda(\eta) + e^{-\eta/\alpha}h(\eta),
    \label{eq:lower-term-derivative}
\end{align}
where the derivative of the indicator function follows again by the sifting property of the delta function.

Similarly, differentiating the second term in \eqref{eq:Z-lambda-eta-decomposition} gives
\begin{align}
    &\frac{d}{d\eta}
    B_\lambda(\eta) \nonumber \\
    &=\frac{d}{d\eta}
    \E_{q_0}
    \sbrc{
        \exp\brc{
            -\frac{(1-\lambda)c(x_0)+\lambda\eta
            +\frac{\lambda}{1-\beta}(c(x_0)-\eta)}{\alpha}
        }
        \mathds{1}\cbrc{c(x_0)>\eta}
    }
    \nonumber\\
    &=
    \E_{q_0}
    \sbrc{
        \brc{
            \exp\brc{
                -\frac{(1-\lambda)c(x_0)+\lambda\eta
                +\frac{\lambda}{1-\beta}(c(x_0)-\eta)}{\alpha}
            }
        }'
        \mathds{1}\cbrc{c(x_0)>\eta}
    }
    \nonumber\\
    &\quad+
    \E_{q_0}
    \sbrc{
        \exp\brc{
            -\frac{(1-\lambda)c(x_0)+\lambda\eta
            +\frac{\lambda}{1-\beta}(c(x_0)-\eta)}{\alpha}
        }
        \brc{
            \mathds{1}\cbrc{c(x_0)>\eta}
        }'
    }
    \nonumber\\
    &=
    \frac{\lambda\beta}{\alpha(1-\beta)}
    \E_{q_0}
    \sbrc{
        \exp\brc{
            -\frac{(1-\lambda)c(x_0)+\lambda\eta
            +\frac{\lambda}{1-\beta}(c(x_0)-\eta)}{\alpha}
        }
        \mathds{1}\cbrc{c(x_0)>\eta}
    }
    -
    e^{-\eta/\alpha}h(\eta) \nonumber \\
    &= \frac{\lambda\beta}{\alpha(1-\beta)} B_\lambda(\eta) -
    e^{-\eta/\alpha}h(\eta).
    \label{eq:upper-term-derivative}
\end{align}
Adding \eqref{eq:lower-term-derivative} and \eqref{eq:upper-term-derivative}, the two boundary terms cancel, and hence
\begin{equation}
\label{eq:Z-lambda-eta-first-derivative}
    Z_{\lambda}'(\eta) = -\frac{\lambda}{\alpha} A_\lambda(\eta) + \frac{\lambda\beta}{\alpha(1-\beta)} B_\lambda(\eta).
\end{equation}
Differentiating \eqref{eq:Z-lambda-eta-first-derivative}
and using \eqref{eq:lower-term-derivative}
and \eqref{eq:upper-term-derivative} yields
\begin{align}
    &Z_{\lambda}''(\eta) \nonumber \\
    &= -\frac{\lambda}{\alpha} \brc{ -\frac{\lambda}{\alpha} A_\lambda(\eta) + e^{-\eta/\alpha} h(\eta) } + 
    \frac{\lambda\beta}{\alpha(1-\beta)} \brc{ \frac{\lambda\beta}{\alpha(1-\beta)} B_\lambda(\eta) - e^{-\eta/\alpha}h(\eta) } \nonumber \\
    &= \frac{\lambda^2}{\alpha^2} A_\lambda(\eta) + \frac{\lambda^2\beta^2}{\alpha^2(1-\beta)^2} B_\lambda(\eta) - \frac{\lambda}{\alpha(1-\beta)} e^{-\eta/\alpha}h(\eta). \label{eq:Z-lambda-eta-second-derivative}
\end{align}

Now, since $G_\lambda(\eta) = -\alpha \log Z_{\lambda}(\eta)$, we have 
\begin{align*}
G_\lambda'(\eta) =-\alpha\frac{Z'_{\lambda}(\eta)}{Z_{\lambda}(\eta)}=\lambda-\frac{\lambda}{1-\beta} \cdot    \frac{B_\lambda(\eta)}{Z_\lambda(\eta)}
\end{align*}
and
\begin{equation}
\label{eq:G-lambda-second-derivative-in-Z}
    G_\lambda''(\eta) = -\alpha \frac{ Z_{\lambda}''(\eta) Z_{\lambda}(\eta) -(Z_{\lambda}'(\eta))^2}{Z_{\lambda}(\eta)^2}.
\end{equation}
Substituting \eqref{eq:Z-lambda-eta-first-derivative} and \eqref{eq:Z-lambda-eta-second-derivative} into \eqref{eq:G-lambda-second-derivative-in-Z} and simplifying yields
\begin{equation}
\label{eq:G-lambda-second-derivative}
G_\lambda''(\eta) = \frac{\lambda}{(1-\beta)Z_{\lambda}(\eta)^2}
\brc{
e^{-\eta/\alpha}h(\eta)Z_{\lambda}(\eta) -
\frac{\lambda}{\alpha(1-\beta)}
A_\lambda(\eta)
B_\lambda(\eta)
}.
\end{equation}
Note that here the coefficient $\frac{\lambda}{(1-\beta)Z_{\lambda}(\eta)^2}$ is strictly positive. Hence, $G_\lambda(\eta)$ is convex if and only if
\[
e^{-\eta/\alpha}h(\eta)Z_{\lambda}(\eta)
\geq
\frac{\lambda}{\alpha(1-\beta)}
A_\lambda(\eta)
B_\lambda(\eta),\quad \forall \eta \in \mbR,
\]
which is exactly \eqref{eq:eta-convexity-condition-general-lambda}.
Before we end the proof, we also note that, since the first term in \eqref{eq:G-lambda-second-derivative} is nonnegative,
\begin{equation}
\label{eq:G-weak-convex}
G_\lambda''(\eta) \geq -
\frac{\lambda^2}{\alpha(1-\beta)^2}
\frac{A_\lambda(\eta)
B_\lambda(\eta)}{Z_\lambda(\eta)^2} \geq -
\frac{\lambda^2}{4\alpha(1-\beta)^2}, 
\end{equation}
since $0 \leq (A-B)^2 = (A+B)^2 - 4 A B$ for $A,B \geq 0$. 
The proof is now complete.
\end{proof}

Next, we provide a simple sufficient condition for non-convexity.
Suppose that, for some
$\bar{\eta}\in\mbR$,
\begin{equation}
\label{eq:eta-nonconvex-example}
h(\bar{\eta})=0,
\qquad
0<\mbP_{q_0}\cbrc{c(x_0)\leq\bar{\eta}}<1.
\end{equation}
Intuitively, \eqref{eq:eta-nonconvex-example} means that there exists an interior splitting point of the distribution. This holds, for example, when the density $h$ is bimodal with a zero-probability belt in the midst. Note that this is only a sufficient condition rather than necessary.  
Then, at $\eta = \bar{\eta}$, the l.h.s. of \eqref{eq:eta-convexity-condition-general-lambda} becomes 0, whereas the r.h.s. remains positive.
Thus, $G_\lambda$ is not convex.

\section{Proof of Proposition \texorpdfstring{\ref{prop:eta-convexity}}{2}}
\label{app:prop:eta-convexity-proof}

Define
\[ g(x_0,\eta):=q_0(x_0)e^{-c_{\lambda,\eta}(x_0)/\alpha},\quad Z_\lambda(\eta):=\mathbb{E}_{x_0 \sim q_0}\left[e^{-c_{\lambda,\eta}(x_0)/\alpha}\right]=\int_{\calX} g(x_0,\eta) \d x_0.\]
From the definition of $c_{\lambda,\eta}$ in \eqref{eq:weighted-cost-def}, the logarithm of $g(x_0,\eta)$ is 
\[ \log g(x_0,\eta)=\log q_0(x_0)-\frac{1}{\alpha} c_{\lambda,\eta}(x_0)=\log q_0(x_0)-\frac{1-\lambda}{\alpha} c(x_0) - \frac{\lambda}{\alpha} \eta - \frac{\lambda}{\alpha(1-\beta)} (c(x_0)-\eta)^+ . \] Because $c(x_0)$ and $(c(x_0)-\eta)^+=\max\{c(x_0)-\eta,0\}$ are jointly convex in $(x_0,\eta)$ and $\log q_0(x_0)$ is concave in $(x_0,\eta)$, $g(x_0,\eta)$ is jointly log-concave in $(x_0,\eta)$. By Pr\'ekopa's Theorem, marginalization preserves log-concavity (see, e.g., Theorem 6 in \cite{prekopa1973logarithmic}). Therefore, $Z_\lambda(\eta)=\int_{x_0} g(x_0,\eta) \d x_0$ is log-concave in $\eta$. As a result, the objective function $G_\lambda(\eta)=-\alpha\log Z_\lambda(\eta)$ is convex in $\eta$.
The proof is now complete.

\section{Corollary \texorpdfstring{\ref{cor:risk-averse-reduction-regularized}}{1} and its proof} \label{appendix:regularized-reduction}

\begin{corollary}[Regularized version's reduction to $\eta$ optimization]
\label{cor:risk-averse-reduction-regularized}
Under the conditions of \Cref{thm:risk-averse-reduction},
fix $\kappa\geq0$ and $\eta_c\in\mbR$.
The quadratically regularized version of \eqref{eq:obj-path-risk} is:
\[
\min_{\pi,\eta}
\left\{
\E_{\pi}\sbrc{c_{\lambda,\eta}(x_0)}
+\alpha\KL{\pi_{1:T+1}}{\mu_{1:T+1}}
+\frac{\kappa}{2}(\eta-\eta_c)^2
\right\},
\]
which can be reduced to
\[
\min_{\eta\in\mbR}G_{\lambda,\kappa}(\eta) = G_\lambda(\eta)+\frac{\kappa}{2}(\eta-\eta_c)^2,
\]
as defined in \eqref{eq:regularized-obj-eta}.
Meanwhile, for any minimizer $\eta_\kappa^*$, the corresponding optimal
reverse transitions satisfy
\[
\pi_{t,\eta_\kappa^*}^*(x_{t-1}\mid x_t)
\propto
\mu_t(x_{t-1}\mid x_t)
e^{-V_{t-1,\eta_\kappa^*}^*(x_{t-1})/\alpha},
\]
for $t=T+1,\ldots,1$, where $V_{t,\eta}^*$ is given
by \eqref{eq:Vt-eta-star-def}.
\end{corollary}

\begin{proof}[Proof of \Cref{cor:risk-averse-reduction-regularized}]
For fixed $\eta$, the regularization term
$\frac{\kappa}{2}(\eta-\eta_c)^2$ is independent of $\pi$.
It therefore leaves the minimizing policies in
\eqref{eq:value-tail-fixed-eta} unchanged.
By \Cref{thm:risk-averse-reduction} and \eqref{eq:obj-eta},
the optimal value of the regularized inner problem is
\[
V_{T+1,\eta}^*+\frac{\kappa}{2}(\eta-\eta_c)^2
=
G_\lambda(\eta)+\frac{\kappa}{2}(\eta-\eta_c)^2
=
G_{\lambda,\kappa}(\eta).
\]
Minimizing over $\eta$ proves the reduction.
Substituting any minimizer $\eta_\kappa^*$ into the optimal
transitions in \Cref{thm:risk-averse-reduction} gives an
optimal policy for the regularized problem.
\end{proof}

\section{Proof of Theorem \texorpdfstring{\ref{thm:eta-optimization}}{2}}
\label{app:thm:eta-optimization-proof}

To prove \Cref{thm:eta-optimization}, we first provide a helpful lemma for the ternary search algorithm below, adapted from Lemma 33 of \cite{cohen2016geometric}.

\begin{lemma}[Lemma 33, \cite{cohen2016geometric}]
\label{lem:onedimmin} 
Let $g:\mathbb{R}\rightarrow\mathbb{R}$ be an $L$-Lipschitz
convex function defined on the $[l,u]$ interval and let $\hat{g}\,:\,\mathbb{R}\rightarrow\mathbb{R}$
be an oracle such that $\left|\hat{g}(y)-g(y)\right|\leq\tilde{\epsilon}$ for all
$y$. With at most $1+2\lceil\log_{3/2}(\frac{L(u-l)}{\tilde{\epsilon}})\rceil$
calls to $\hat{g}$, the algorithm $\text{CVaR-Optimizer}$ with Lipschitz bound $L$ and target error $\tilde{\epsilon}$
outputs a point $x$ such that 
\[
g(x)-\min_{y\in[l,u]}g(y)\leq4\tilde{\epsilon}.
\]
\end{lemma}

To invoke \Cref{alg:onedopt} and apply \Cref{lem:onedimmin}, we provide the following lemma to characterize the Lipschitz constants for the regularized objectives. 

\begin{lemma}[Lipschitz continuity of the objectives]
\label{lem:eta-lipschitz}
Recall the definitions of $L$ and $L_\kappa$ in \eqref{eq:obj-L-Lkappa-def}:
\[ L := \lambda\max\left\{1,\frac{\beta}{1-\beta}\right\},\quad L_\kappa := L + \kappa (u-l). \]
Under the assumptions of
\Cref{thm:eta-optimization}, define the Monte Carlo estimate of the objective in \eqref{eq:obj-eta} as
\[
    \widehat G_{\lambda,N}(\eta)
    :=-\alpha\log\left(
    \frac1N\sum_{i=1}^N
    e^{-c_{\lambda,\eta}(x_0^i)/\alpha}
    \right).
\]
Then, we have
\begin{enumerate}
    \item Both $G_\lambda$ and $\widehat G_{\lambda,N}$ are $L$-Lipschitz;
    \item Both $G_{\lambda,\kappa}$ and $\widehat G_{\lambda,\kappa,N}$ are $L_\kappa$-Lipschitz on $[l,u]$;
    \item The difference $\widehat G_{\lambda,\kappa,N}-G_{\lambda,\kappa}$ is $2L$-Lipschitz on $[l,u]$.
\end{enumerate}
\end{lemma}
\begin{proof}
See \Cref{app:lem:eta-lipschitz-proof}.    
\end{proof}

While Lemma \ref{lem:onedimmin} establishes that the distance between the objective function value at the output solution and the optimal objective value can be upper bounded, for the regularized objective function \eqref{eq:regularized-obj-eta}, we can further bound the distance between the output $\hat{\eta}$ and the optimal $\eta_{\kappa}^*$ using the strongly convex property of $G_{\lambda,\kappa}(\eta)$.

\begin{lemma}\label{lem:onedopt-query-complexity}
Let $\delta\in(0,1)$ denote the confidence level, and define
$K:=\lceil\log_{3/2}(\frac{4L_\kappa(u-l)}{\epsilon})\rceil$. Suppose that $G_{\lambda,\kappa}(\eta)$ is $m_{\kappa}$-strongly convex on $[l,u]$, and suppose we have an inexact
objective evaluation oracle $\hat G_{\lambda,\kappa}(\eta)$ such that, with probability at least
$1-\delta$, $\sup_{\eta\in[l,u]}|G_{\lambda,\kappa}(\eta)-\hat G_{\lambda,\kappa}(\eta)|\le\epsilon/4$. Then, with probability at least $1-\delta$, the output
$\hat{\eta}$ of \Cref{alg:onedopt} (with Lipschitz bound $L_\kappa$ and target error $\tilde{\epsilon}=\epsilon/4$) is in the neighborhood of the optimal solution $\eta_\kappa^*$ of the regularized
problem~\eqref{eq:regularized-obj-eta} (i.e., $|\hat{\eta}-\eta_\kappa^*|\le\sqrt{2\epsilon/m_\kappa}$) by calling the
inexact oracle at most $1+2K$ times.
\end{lemma}
\begin{proof}
    See \Cref{app:lem:onedopt-query-complexity-proof}.
\end{proof}

For the regularized objective $G_{\lambda,\kappa}(\eta)$, suppose that we have a Monte Carlo
estimator $\hat G_{\lambda,\kappa,N}(\eta)$:
$\hat G_{\lambda,\kappa,N}(\eta)=-\alpha\log(\frac1N\sum_{i=1}^N e^{-c_{\lambda,\eta}(x_0^i)/\alpha})+\frac\kappa2(\eta-\eta_c)^2$,
where $\{x_0^i\}_{i=1}^N$ is a single batch of i.i.d.\ samples from $q_0$, used for every $\eta$. We next derive the sample complexity of this Monte Carlo
estimator, uniformly over $\eta$.

\begin{lemma}\label{lem:saa-sample-complexity}
Suppose \Cref{assum:bounded-cost} holds, and let
\[
  \underline{y}:=\exp\Bigl(-\tfrac{1-\lambda}{\alpha}\bigl(\mathbb{E}[c(X)]-l\bigr)
  -\tfrac{\lambda}{\alpha(1-\beta)}\mathbb{E}\bigl[(c(X)-l)^+\bigr]\Bigr),
\]
\[
  L:=\lambda\max\Bigl\{1,\frac{\beta}{1-\beta}\Bigr\},
  \qquad
  J:=\Bigl\lceil\frac{16L(u-l)}{\epsilon}\Bigr\rceil+1.
\]
Suppose $\epsilon\le 8\alpha$. Then
$\sup_{\eta\in[l,u]}|\hat G_{\lambda,\kappa,N}(\eta)-G_{\lambda,\kappa}(\eta)|\le\epsilon/4$ with
probability at least $1-\delta$ as long as
\[
  N\ \ge\ \frac{C_G}{\epsilon^2}\log\frac{2J}{\delta},
  \qquad\text{where}\quad C_G:=\frac{2048\,\alpha^2}{3\,\underline{y}}.
\]
\end{lemma}
\begin{proof}
    See \Cref{app:lem:saa-sample-complexity-proof}.
\end{proof}

\begin{reptheorem}{thm:eta-optimization}
Suppose \Cref{assum:bounded-cost} holds.
Let $\eta_c \in [l,u]$ be arbitrary, and suppose that
$G_{\lambda,\kappa}$ is $m_\kappa$-strongly convex on
$[l,u]$ for some $m_\kappa>0$.
Define
\begin{equation*}
    L := \lambda\max\left\{1,\frac{\beta}{1-\beta}\right\},\quad L_\kappa := L + \kappa (u-l).
\end{equation*}
For any $\delta\in(0,1)$ and sufficiently small
$\eps_\eta>0$, define
\[
  K:=\Bigl\lceil\log_{3/2}\Bigl(\frac{8L_\kappa(u-l)}{m_\kappa\eps_\eta^2}\Bigr)\Bigr\rceil,
  \quad
  J:=\Bigl\lceil\frac{32L(u-l)}{m_\kappa\eps_\eta^2}\Bigr\rceil+1.
\]

Then, with probability at least $1-\delta$, the output $\hat\eta$ of \Cref{alg:onedopt} (with Lipschitz bound $L_\kappa$ and target error $m_\kappa \eps_\eta^2/8$) satisfies
$|\hat\eta-\eta_\kappa^*|\le\eps_\eta$
as long as
\[
  N\ \ge\ \frac{4\,C_G}{m_\kappa^2\,\eps_\eta^4}\log\frac{2J}{\delta}
\]
where $C_G>0$ is a constant depending on
$\alpha,\beta,\lambda,l,u$. Furthermore, the algorithm calls the Monte Carlo oracle $\widehat G_{\lambda,\kappa,N}$ at most $1+2K$ times. 
\end{reptheorem}
\begin{proof}
Apply Lemmas~\ref{lem:onedopt-query-complexity} and~\ref{lem:saa-sample-complexity} with $\epsilon:=\frac{m_\kappa\eps_\eta^2}{2}$. Then
$\epsilon\le 8\alpha$ and $\epsilon/4=\frac{m_\kappa\eps_\eta^2}{8}$. Moreover, the quantities
$K$ of Lemma~\ref{lem:onedopt-query-complexity} and $J$ of Lemma~\ref{lem:saa-sample-complexity} coincide with the $K$ and $J$ defined
above, and $\frac{C_G}{\epsilon^2}=\frac{4C_G}{m_\kappa^2\eps_\eta^4}$.
 
By Lemma~\ref{lem:saa-sample-complexity}, with probability at least $1-\delta$,
$\sup_{\eta\in[l,u]}|\hat G_{\lambda,\kappa,N}(\eta)-G_{\lambda,\kappa}(\eta)|\le\epsilon/4$. This is exactly the assumption of Lemma~\ref{lem:onedopt-query-complexity}, with oracle
$\hat G_{\lambda,\kappa}=\hat G_{\lambda,\kappa,N}$. Therefore, with
probability at least $1-\delta$, the output satisfies
\[
  |\hat\eta-\eta_\kappa^*|\le\sqrt{\frac{2\epsilon}{m_\kappa}}=\eps_\eta,
\]
using at most $1+2K$ oracle calls.
\end{proof}

\section{Proof of Theorem \texorpdfstring{\ref{thm:paired-value-estimation}}{3}}
\label{app:thm:paired-value-estimation-proof}

First, for every $\eta\in[l,u]$, we have
\begin{equation} \label{eq:c_lambda_eta_uniform_bounds}
    0\le c_{\lambda,\eta}(x_0)-a_\eta
    =(1-\lambda)(c(x_0)-l)
      +\frac{\lambda}{1-\beta}(c(x_0)-\eta)^+
    \le (u-l)(1-\lambda+\frac{\lambda}{1-\beta})=\frac{\alpha\xi}{2},
\end{equation}
where $\xi$ is defined in \eqref{eq:overline-U-def}.
As a result, the targets $Y_{\eta}$, their conditional means $\mathbb{E}_{q_{0|t}}Y_{\eta}$, and all functions
in $\mathcal F_t$ take values in
$[e^{-\frac{\xi}{2}},1]$. By the mean value theorem, for any $a,b\in [e^{-\frac{\xi}{2}},1]$, we have $|\log a-\log b|=\frac{1}{\zeta}|a-b|$ for some $\zeta\in [a,b]$. Since $\zeta\ge e^{-\frac{\xi}{2}}$, this gives $|\log a-\log b|\le e^{\frac{\xi}{2}}|a-b|$. 
Thus, using the logarithmic
Lipschitz bound above, we obtain
\begin{align*}
   |\widehat V_{t,\eta,N}(x_t)-V_{t,\eta}^*(x_t)|^2
    \le \left|a_\eta-\alpha\log\widehat f_{t,\eta}(x_t)
      -V^\star_{t,\eta}(x_t)\right|^2 \le
    \alpha^2e^{\xi}
    |\widehat f_{t,\eta}(x_t)
          -\E_{q_{0|t}}Y_\eta(x_t)|^2,\quad \forall x_t.
\end{align*}
Integrating both sides w.r.t. $q_t(x_t)$ yields
\begin{equation}
    \norm{\widehat V_{t,\eta,N}-V_{t,\eta}^*}_{L_2(q_t)}^2
    \le
    \alpha^2e^{\xi}
    \norm{\widehat f_{t,\eta}
          -\E_{q_{0|t}}Y_\eta}_{L_2(q_t)}^2.
    \label{eq:value-error-from-regression}
\end{equation}
Starting from here, we omit the dependence of $\E_{q_{0|t}}Y_\eta$ on $x_t$.

We next control the regression error. For any fixed
$g\in\mathcal F_t$, 
we have 
\begin{align*}
(g(x_t)-Y_\eta(x_0))^2=&((g(x_t)-\E_{q_{0|t}}Y_\eta)+(\E_{q_{0|t}}Y_\eta-Y_\eta(x_0)))^2\\
=&(g(x_t)-\E_{q_{0|t}}Y_\eta)^2+\underbrace{2(g(x_t)-\E_{q_{0|t}}Y_\eta)(\E_{q_{0|t}}Y_\eta-Y_\eta(x_0))}_{\text{cross term}}+(\E_{q_{0|t}}Y_\eta-Y_\eta(x_0))^2.
\end{align*}
Now, take an expectation by conditioning on $x_t$ on both sides. By tower property, $\mathbb{E}[h]=\mathbb{E}_{x_t\sim q_t}[\mathbb{E}[h|x_t]]$. For the cross term, we have $\mathbb{E}[(g(x_t)-\E_{q_{0|t}}Y_\eta)(\E_{q_{0|t}}Y_\eta-Y_\eta(x_0))]=\mathbb{E}_{x_t\sim q_t}[(g(x_t)-\E_{q_{0|t}}Y_\eta)(\E_{q_{0|t}}Y_\eta-\mathbb{E}[Y_\eta(x_0)|x_t])]=0$ because $\E[Y_\eta(x_0)\mid x_t]=\E_{q_{0|t}}Y_\eta$.
As a result, we have
\[
    \E_{q_{0,t}}\sbrc{
        (g(x_t)-Y_\eta(x_0))^2
        -(\E_{q_{0|t}}Y_\eta-Y_\eta(x_0))^2
    }
    =
    \norm{g-\E_{q_{0|t}}Y_\eta}_{L_2(q_t)}^2.
\]
Moreover, the excess loss factors as
\[
    (g-Y_\eta)^2-(\E_{q_{0|t}}Y_\eta-Y_\eta)^2
    =
    (g-\E_{q_{0|t}}Y_\eta)
    (g+\E_{q_{0|t}}Y_\eta-2Y_\eta).
\]
Since all predictions and targets lie in $[0,1]$, its absolute
value is at most one, and $|g+\E_{q_{0|t}}Y_\eta-2Y_\eta|\le 2$. As a result, its second moment satisfies
\[
    \E_{q_{0,t}}\sbrc{
        (g-\E_{q_{0|t}}Y_\eta)^2
        (g+\E_{q_{0|t}}Y_\eta-2Y_\eta)^2
    }
    \le
    4\norm{g-\E_{q_{0|t}}Y_\eta}_{L_2(q_t)}^2.
\]
Thus its variance is controlled by its mean, allowing us to apply the
Bernstein's inequality.

Recall Bernstein's inequality: for i.i.d. $Z_i$ with $|Z_i|\le 1~~(\implies |Z_i - \E Z_i| \leq 2)$, 
\begin{align*}
    \mathbb{P}\left(\mathbb{E}[Z]-\frac{1}{N}\sum_{i=1}^N Z_i >\epsilon\right)\le \exp\left(-\frac{N\epsilon^2}{2\Var(Z)+\frac{4}{3}\epsilon}\right).
\end{align*}
Setting the right-hand side to $\delta'$, we get,
with probability at least $1-\delta'$, 
\begin{align*}
    \mathbb{E}[Z]-\frac{1}{N}\sum_{i=1}^NZ_i\le \sqrt{\frac{2\Var(Z)\log(1/\delta')}{N}}+\frac{
    4\log(1/\delta')}{3N}.
\end{align*}
Fix $g\in\mathcal{F}_t$ and $\eta \in [l,u]$, and let $\mu_\eta:=\E_{q_{0|t}}Y_\eta$. For each sample $i$, define $Z_i=(g(x_t^i)-Y_\eta(x_0^i))^2 -(\mu_\eta(x_t^i)-Y_\eta(x_0^i))^2$. Its population and empirical averages are
\begin{align*}
    D:=\mathbb{E}[Z]=\norm{g-\mu_\eta}_{L_2(q_t)}^2,\ \widehat{D}=\frac{1}{N}\sum_{i=1}^NZ_i=\widehat{\mathcal{L}}_{t,\eta}(g)-\widehat{\mathcal{L}}_{t,\eta}(\mu_\eta).
\end{align*}
Then we have, with probability at least $1-\delta'$, 
\begin{align}
    D\le \widehat{D}+\sqrt{\frac{8D\log(1/\delta')}{N}}+\frac{4\log(1/\delta')}{3N}. \label{eq:pointwise}
\end{align}

Next, we make this bound uniform using a grid with the union bound. Recall $\xi := \frac{2(u-l)}{\alpha}(1-\lambda+\frac{\lambda}{1-\beta})$. Let $m:=e^{-\xi/2}$,
and
$\tau := \frac{\varepsilon_v^2}{40(1+\alpha^2)}e^{-\xi}$.
 
\paragraph{Grid.}
Write $g_\theta:=\mathrm{clip}_{[m,1]}(\theta^\top\phi)$. Since clipping is $1$-Lipschitz and
$|\phi_j|\le 1$, $\|g_\theta-g_{\theta'}\|_\infty\le\|\theta-\theta'\|_1$. Now take a grid with spacing $2\tau/F$ in each coordinate. Every $\theta_j$ is within $\tau/F$ of a grid value $\theta'_j$ for $j=1,\ldots, F$. Summing over the $F$ coordinates gives $\|\theta-\theta'\|_1\le \tau$ and thus $\|g_\theta-g_{\theta'}\|_\infty\le \tau$. Therefore, a 
grid $\Theta_\tau\subset[-1,1]^F$ with spacing $2\tau/F$ in each coordinate yields
$\mathcal{G}_\tau:=\{g_\theta:\theta\in\Theta_\tau\}\subseteq\mathcal{F}_t$, which is a
$\tau$-cover of $\mathcal{F}_t$ in sup norm. Each coordinate has at most $2/(2\tau/F)+1=F/\tau+1\le 2F/\tau$ points. So, $|\mathcal{G}_\tau|=|\Theta_{\tau}|\le (\frac{2F}{\tau})^F=(\frac{80F(1+\alpha^2)e^{\xi}}{\varepsilon_v^2})^F$.
Since $c_{\lambda,\eta}-a_\eta=(1-\lambda)(c-l)+\frac{\lambda}{1-\beta}(c-\eta)^+$ and
$s\mapsto e^{-s/\alpha}$ is $\alpha^{-1}$-Lipschitz on $[0,\infty)$,
$\|Y_\eta-Y_{\eta'}\|_\infty\le\frac{\lambda}{\alpha(1-\beta)}|\eta-\eta'|$. By Jensen's
inequality, the same bound holds for $\|\mu_\eta-\mu_{\eta'}\|_\infty$. Hence there is a grid
$H_\tau\subset[l,u]$ with spacing $2\tau\alpha(1-\beta)/\lambda$ such that every $\eta\in [l,u]$ has some $\eta'\in H_\tau$ with
$\|Y_\eta-Y_{\eta'}\|_\infty\le\tau,\ \|\mu_\eta-\mu_{\eta'}\|_\infty\le\tau$. When $\lambda=0$, $c_{\lambda,\eta}-a_\eta=c-l$ does not depend on $\eta$, so $Y_\eta$
and $\mu_\eta$ are the same for all $\eta$, and we can take any single point, e.g., $H_\tau=\{l\}$. 

The number of points in this grid $H_\tau$ is then at most
\begin{align*}
    \frac{u-l}{2\tau\alpha(1-\beta)/\lambda}+1=\frac{10\lambda\xi(1+\alpha^2)e^{\xi}}{(1-\beta+\lambda\beta)\varepsilon_v^2}+1\le \frac{\frac{10\lambda\xi}{1-\beta+\lambda\beta}(1+\alpha^2)e^{\xi}+1}{\varepsilon_v^2}.
\end{align*}
Now take $C_0:=\max\{80F(1+\alpha^2)e^{\xi},\frac{10\lambda\xi}{1-\beta+\lambda\beta}(1+\alpha^2)e^{\xi}+1\}$. Then we have $|\mathcal{G}_{\tau}|\le (C_0/\varepsilon_v^2)^F$ and
$|H_\tau|\le C_0/\varepsilon_v^2$.
Both grids depend only on $(\alpha,\beta,\lambda,l,u,F,\varepsilon_v)$ and not on the data.
 
\paragraph{Uniform bound on the grid.}
Let $K:=T|\mathcal{G}_\tau||H_\tau|\le T(C_0/\varepsilon_v^2)^{F+1}$ and $L:=\log(K/\delta)$.
Apply \eqref{eq:pointwise} with $\delta'=\delta/K$ to every $(t,g,\eta)\in[T]\times\mathcal{G}_\tau\times H_\tau$,
and use the union bound $\mathbb{P}(\text{some grid bound fails})\le K\delta'=\delta$. Let $\mathcal{E}$ be the event that all $K$ bounds hold simultaneously. Then $\mathbb{P}(\mathcal{E})\ge 1-\delta$. In other words, with probability at least $1-\delta$, simultaneously for all grid triples, we have 
\begin{align*}
    D\le  \widehat{D}+\sqrt{\frac{8DL}{N}}+\frac{4L}{3N}\le \widehat{D}+\frac{D}{2}+\frac{4L}{N}+\frac{4L}{3N}
\end{align*}
where the second inequality uses the arithmetic mean-geometric mean (AM-GM) inequality $\sqrt{ab}\le (a+b)/2$. 
This is equivalent to 
\begin{align}
  \|g-\mu_\eta\|_{L_2(q_t)}^2&\le 2\bigl(\widehat{\mathcal{L}}_{t,\eta}(g)-\widehat{\mathcal{L}}_{t,\eta}(\mu_\eta)\bigr)+\frac{32}{3}\cdot\frac{L}{N}\nonumber\\
  &\le 2\bigl(\widehat{\mathcal{L}}_{t,\eta}(g)-\widehat{\mathcal{L}}_{t,\eta}(\mu_\eta)\bigr)+\frac{C}{N}\log\frac{T}{\delta\varepsilon_v} \label{eq:uniform-bound-grid}
\end{align}
where the second inequality is because for sufficiently small $\varepsilon_v$, we have
$L\le\log\frac{T}{\delta}+(F+1)\log\frac{C_0}{\varepsilon_v^2}\le[1+(F+1)(\log C_0+2)]\log\frac{T}{\delta\varepsilon_v}$ and all the constants are absorbed to $C:=\frac{32}{3}[1+(F+1)(\log C_0+2)]$.

 \paragraph{Perturbation.}
For $a,a',b,b'\in[0,1]$,
$|(a-b)^2-(a'-b')^2|\le 2(|a-a'|+|b-b'|)$, since
$(a-b)^2-(a'-b')^2=((a-b)-(a'-b'))((a-b)+(a'-b'))$ and the second factor is at most $2$ in
absolute value. Applying this pointwise, both the empirical losses and $D$ change by at most
$2(\|g-g'\|_\infty+\|h-h'\|_\infty)$ when the prediction $g$ and the target or conditional
mean $h$ are perturbed in sup norm. For example, for $D$, we have 
\begin{align}
|\|g-h\|_{L_2(q_t)}^2-\|g'-h'\|_{L_2(q_t)}^2|\le 2(\|g-g'\|_\infty+\|h-h'\|_\infty).\label{eq:perturbation-D}
\end{align}
For the empirical loss, we have
\begin{align}
\Bigl|\tfrac1N\textstyle\sum_{i=1}^N(\underbrace{g(x_t^i)}_{a}-\underbrace{Y(x_0^i)}_{b})^2
 -\tfrac1N\sum_{i=1}^N(\underbrace{g'(x_t^i)}_{a'}-\underbrace{Y'(x_0^i)}_{b'})^2\Bigr|\le 2\bigl(\|g-g'\|_\infty+\|Y-Y'\|_\infty\bigr).\label{eq:perturbation-empirical}
\end{align}

\paragraph{Transfer to $\widehat f_{t,\eta}$ and arbitrary $\eta$.} Let $\rho(\varepsilon_v):=\frac{\varepsilon_v^2}{8(1+\alpha^2)}e^{-\xi}$ denote the optimality gap of $\widehat f_{t,\eta}$.
On $\mathcal{E}$, fix $t$ and $\eta$. Pick $\eta'\in H_\tau$ and $g'\in\mathcal{G}_\tau$
such that $\|Y_\eta-Y_{\eta'}\|_\infty\le\tau,\ \|\mu_\eta-\mu_{\eta'}\|_\infty\le\tau,\ \|\widehat f_{t,\eta}-g'\|_\infty\le\tau$. By the perturbation bound, the near-optimality
of $\widehat f_{t,\eta}$, and realizability $\mu_\eta\in\mathcal{F}_t$,
\begin{align}
  \widehat{\mathcal{L}}_{t,\eta'}(g')
  \overset{(i)}{\le}\widehat{\mathcal{L}}_{t,\eta}(\widehat f_{t,\eta})+4\tau
  \overset{(ii)}{\le}\widehat{\mathcal{L}}_{t,\eta}(\mu_\eta)+\rho(\varepsilon_v)+4\tau
  \overset{(iii)}{\le}\widehat{\mathcal{L}}_{t,\eta'}(\mu_{\eta'})+\rho(\varepsilon_v)+8\tau, \label{eq:empirical-loss-bound}
\end{align}
where $(i)$ is by applying the perturbation bound \eqref{eq:perturbation-empirical} with $a=g'(x_t^i),\ a'=\hat{f}_{t,\eta}(x_t^i),\ b=Y_{\eta'}(x_0^i),\ b'=Y_{\eta}(x_0^i)$, $(ii)$ is because $\widehat{\mathcal{L}}_{t,\eta}(\widehat f_{t,\eta})\le \min_{g\in\mathcal F_t} \widehat{\mathcal L}_{t,\eta}(g)+\rho(\varepsilon_v) \le \widehat{\mathcal{L}}_{t,\eta}(\mu_\eta)+\rho(\varepsilon_v)$, and $(iii)$ is by applying the perturbation bound in \eqref{eq:perturbation-empirical} with $a=\mu_\eta(x_t^i),\ a'=\mu_{\eta'}(x_t^i),\ b=Y_{\eta}(x_0^i),\ b'=Y_{\eta'}(x_0^i)$.
Combining this with the perturbation bound for $D$ and with \eqref{eq:uniform-bound-grid} at $(t,g',\eta')$ gives
\[
  \|\widehat f_{t,\eta}-\mu_\eta\|_{L_2(q_t)}^2
  \overset{(iv)}{\le}\|g'-\mu_{\eta'}\|_{L_2(q_t)}^2+4\tau
  \overset{(v)}{\le} 20\tau+2\rho(\varepsilon_v)+\frac{C}{N}\log\frac{T}{\delta\varepsilon_v},
\]
where $(iv)$ is by applying the perturbation bound \eqref{eq:perturbation-D} and $(v)$ is by Eqs.~\eqref{eq:uniform-bound-grid} and \eqref{eq:empirical-loss-bound}.
Multiplying this by $\alpha^2e^\xi$ and using \eqref{eq:value-error-from-regression}, since $\alpha^2e^\xi\cdot 20\tau\le\varepsilon_v^2/2$
and $\alpha^2e^\xi\cdot 2\rho(\varepsilon_v)\le\varepsilon_v^2/4$, we have
\[
  \|\widehat V_{t,\eta,N}-V^*_{t,\eta}\|_{L_2(q_t)}^2
  \le\frac{3\varepsilon_v^2}{4}+\alpha^2e^\xi\,\frac{C}{N}\log\frac{T}{\delta\varepsilon_v}
  \le\varepsilon_v^2
\]
whenever $N\ge\frac{C}{\varepsilon_v^2}\log\frac{T}{\delta\varepsilon_v}$, with $4\alpha^2e^\xi$
absorbed into $C$. Since $t$ and $\eta$ were arbitrary and everything holds on $\mathcal{E}$,
the bound $\sup_{\eta\in[l,u]}\max_{t\in[T]}\|\widehat V_{t,\eta,N}-V^*_{t,\eta}\|_{L_2(q_t)}\le\varepsilon_v$
holds with probability at least $1-\delta$.

\section{Proof of Theorem \texorpdfstring{\ref{thm:sampling}}{4}}
\label{app:thm:sampling-proof}

Throughout the proof, we assume that tilted-transition sampling is exact (i.e., the number of candidates $M \to \infty$ in SVDD) and that the training events in \Cref{thm:eta-optimization,thm:paired-value-estimation} succeed with probability at least $1-\delta/2$, respectively. Specifically, define the events
\begin{align*}
    \mathcal E_\eta
    &:=
    \left\{
    |\widehat\eta-\eta_\kappa^*|
    \leq\eps_\eta
    \right\},\\
    \mathcal E_v
    &:=
    \left\{
    \sup_{\eta\in[l,u]}
    \max_{t\in[T]}
    \|\widehat V_{t,\eta}-V_{t,\eta}^*\|_{L^2(q_t)}
    \leq\eps_v
    \right\}.
\end{align*}
By \Cref{thm:eta-optimization} and
\Cref{thm:paired-value-estimation}, under the corresponding choice of hyperparameters (particularly with failure probability $\delta/2$), these events satisfy
\[
    \mbP(\mathcal E_\eta\cap\mathcal E_v)\geq1-\delta.
\]

We write $\eta$ for $\hat \eta$ for brevity. In the rest of the proof, we would fix a realization of
the estimation samples in
$\mathcal E_\eta\cap\mathcal E_v$. Thus, from \Cref{thm:eta-optimization},
\begin{equation} \label{assum:eta-optim-err}
    \abs{ \eta - \eta_\kappa^\star } \leq \eps_\eta.
\end{equation}
Also, from \Cref{thm:paired-value-estimation},
\begin{equation} \label{assum:value-est-err}
    \frac{1}{T} \sum_{t=1}^T \E_{x_t\sim q_t} \norm{\widehat V_{t,\eta}(x_t)-V_{t,\eta}^*(x_t)}^2 \le \eps_v^2.
\end{equation}

\paragraph{Total error decomposition.}
To begin, note that from \Cref{thm:risk-averse-reduction}, the marginal distribution induced by $\pi_{t,\eta}^*$ also satisfies (cf. \cite[Theorem~1]{wang2025finetuning})
\[ p_\eta^\star(x_0) \propto \exp\left(-\frac{c_{\lambda, \eta}(x_0)}{\alpha}\right) q_0(x_0). \]
Note that the target distribution of interest is $p_{\eta_\kappa^\star}^\star (x_0)$ (that solves the objective in \eqref{eq:regularized-obj-eta}). The actual sampling distribution is $\hat{p}_{\eta}(x_0)$, where $\eta$ is close to $\eta^\star_\kappa$ and $\hat{p}$ is different from $p^\star$ due to inaccurate value estimate. We can first decompose the sampling error as
\begin{equation} \label{eq:sampling-err-main}
    \TV{p_{\eta_\kappa^\star}^\star (x_0)}{\hat{p}_{\eta}(x_0)} \leq \underbrace{\TV{p_{\eta_\kappa^\star}^\star (x_0)}{p_{\eta}^\star (x_0)}}_{\text{error caused by inaccurately solved $\eta$}} + \underbrace{\TV{p_{\eta}^\star(x_0)}{\hat{p}_{\eta}(x_0)}}_{\text{error caused by inaccurate value estimates}}
\end{equation}

While as follows we assume perfect resampling (i.e., when $M \to \infty$), the corresponding resampling error can be incorporated in the TV triangle inequality above with additional error terms. 

\paragraph{On the error caused by inaccurate value estimates.}
We first tackle the second term in \eqref{eq:sampling-err-main}. The structure follows similarly from \cite[Theorem~3.1]{madhow2026lcb}.
Note that the step-wise KL error is, by data-processing inequality and the KL chain-rule,
\begin{equation} \label{eq:sampling-err-value-main}
    \KL{p_{\eta}^\star(x_0)}{\hat{p}_{\eta}(x_0)} \leq \sum_{t=0}^{T} \E_{p_{t+1,\eta}^\star} \sbrc{ \KL{p_{t|t+1,\eta}^\star}{\hat p_{t|t+1,\eta} \big| x_{t+1}} } = \sum_{t=0}^{T} \E_{p_{t+1,\eta}^\star} \sbrc{ \KL{\pi_{t+1,\eta}^\star}{\hat \pi_{t+1,\eta} \big| x_{t+1}}},
\end{equation}
where we note that $x_{T+1}$ is a dummy initialization state.
The following lemma, originally in \cite[Lemma~C.1]{madhow2026lcb}, is pasted below for convenience.

\begin{lemma}[Lemma C.1 in \cite{madhow2026lcb}]
\label{lem:madhow}
Let $p_a(x) \propto e^{a(x)}q(x)$ and $p_b(x) \propto e^{b(x)}q(x)$, with $|a(x)|\leq U$ and $|b(x)| \leq U$. Then, 
\[ \KL{p_a}{p_b} \leq \frac{e^{2 U}\|a(x) - b(x)\|_{L^2(q)}^2}{2}. \]
\end{lemma}

From \Cref{thm:risk-averse-reduction}, we have
\[ \pi_{t+1,\eta}^*(x_{t}\mid x_{t+1}) \propto \mu_{t+1}(x_{t}\mid x_{t+1}) e^{-V_{t,\eta}^*(x_{t})/\alpha}. \]

The following lemma characterizes the limiting transition kernel of SVDD as $M\to\infty$, with the value estimates held fixed. Related asymptotic consistency results for SMC estimators
are established in \cite{delmoral2006sequential}.

\begin{lemma} \label{lem:svdd-asymp}
Fix $\eta$ and bounded value estimates
$\{\hat V_{t,\eta}\}_{t=0}^T$.
Let $\hat\pi_{t+1,\eta}^{(M)}$ denote the SVDD
transition law with $M$ candidates, averaged over
candidate generation and resampling.
Then, for each $t$, $\eta$, $x_t$, and $x_{t+1}$, we have
\[ \hat\pi_{t+1,\eta}^{(M)}(x_t\mid x_{t+1}) \to \hat\pi_{t+1,\eta}(x_t\mid x_{t+1}),\quad \text{as}~~M \to \infty, \]
where
\[
    \hat\pi_{t+1,\eta}(x_t\mid x_{t+1})
    \propto
    \mu_{t+1}(x_t\mid x_{t+1})
    e^{-\hat V_{t,\eta}(x_t)/\alpha}.
\]
\end{lemma}
\begin{proof}
    See \Cref{app:lem:svdd-asymp-proof}.
\end{proof}

Note that, since both $\hat\pi_{t+1,\eta}^{(M)}$ and $\hat\pi_{t+1,\eta}$ are probability densities and the limit integrates to one, \Cref{lem:svdd-asymp} further implies a convergence in total variation by Scheffé’s lemma. Later, because of the tilting form, \Cref{lem:svdd-asymp} would enable us to invoke \Cref{lem:madhow} for $\KL{ \pi_{t+1,\eta}^\star(\cdot\mid x_{t+1}) }{ \hat\pi_{t+1,\eta}(\cdot\mid x_{t+1}) }$. Before that step, we first need to identify the correct upper bound $U$ to invoke the lemma.
From \eqref{eq:c_lambda_eta_uniform_bounds} in \Cref{thm:paired-value-estimation}, we have
\[ a_\eta \le c_{\lambda,\eta}(x_0) \le a_\eta + \frac{\alpha\xi}{2}, \]
where we recall that $a_\eta=(1-\lambda)l+\lambda\eta$ and $\xi=\frac{2(u-l)}{\alpha}
\left(1-\lambda+\frac{\lambda}{1-\beta}\right)$.
Note that the function $f(z) = -\alpha \log \E \sbrc{ e^{-z/\alpha} }$ preserves the bounds in $z$.
Thus, from the inductive relationship in \eqref{eq:vt-induction-def}, it follows that
\[ a_\eta \leq V_{t,\eta}^*(x_t) \leq a_\eta + \frac{\alpha\xi}{2},\qquad \forall t \geq 0,~\forall x_t \in \mbR^d.
\]

Now, for the estimated value function $\hat V_{t,\eta}(x_t)$, 
by the clipping range of $\widehat f_{t,\eta}$ and the definition
of the value estimator, we again have
\[
a_\eta
\le \hat V_{t,\eta}(x_t)
\le a_\eta+\frac{\alpha\xi}{2},
\qquad \forall t=0,\ldots,T,
\]
where at $t=0$ we use $\hat V_{0,\eta}=V^*_{0,\eta}=c_{\lambda,\eta}$.
Since adding a constant to a potential does not affect the
normalized tilted distribution, we center this common interval
at $a_\eta+\alpha\xi/4$. The normalized potentials in
\Cref{lem:madhow} are then bounded in absolute value by $\frac{\xi}{4}$.

Thus, by \Cref{lem:madhow}, given $x_{t+1}$,
\[ \KL{ \pi_{t+1,\eta}^\star(\cdot\mid x_{t+1}) }{ \hat\pi_{t+1,\eta}(\cdot\mid x_{t+1}) } \leq \frac{ \exp\left(\xi/2 \right) }{2 \alpha^2} \norm{ \hat V_{t,\eta}-V_{t,\eta}^* }_{L^2(\mu_{t+1}(\cdot\mid x_{t+1}))}^2. \]
From \eqref{eq:sampling-err-value-main} and using Pinsker's inequality, we get
\begin{align}
    \TV{p_{\eta}^\star(x_0)}{\hat{p}_{\eta}(x_0)} &\leq \sqrt{\frac{1}{2}  \sum_{t=0}^{T} \E_{p_{t+1,\eta}^\star} \sbrc{ \KL{\pi_{t+1,\eta}^\star}{\hat \pi_{t+1,\eta} \big| x_{t+1}}} } \nonumber \\
    &\stackrel{(i)}{\leq} \sqrt{\frac{1}{2} e^{\xi/2} \sum_{t=0}^{T} \E_{q_{t+1}} \sbrc{ \KL{\pi_{t+1,\eta}^\star}{\hat \pi_{t+1,\eta} \big| x_{t+1}}} } \nonumber \\
    &\leq \sqrt{\frac{1}{2} \frac{ e^{\xi}}{2 \alpha^2} \sum_{t=1}^{T} \E_{q_t} \norm{ \hat V_{t,\eta}-V_{t,\eta}^* }^2 } \nonumber \\
    &\stackrel{(ii)}{\leq} \frac{e^{\xi/2}}{2 \alpha} \sqrt{ T } \eps_v, \label{eq:sampling-err-value-result}
\end{align}
where $(ii)$ follows from \eqref{assum:value-est-err}.
To justify $(i)$, we note that under the optimal policy in \Cref{thm:risk-averse-reduction}, the corresponding marginal sampling distribution is also a tilted form of the pretrained marginal (cf. \cite[Theorem~1]{uehara2024finetuning}):
\[ p^\star_{t+1,\eta} \propto q_{t+1} e^{-V_{t+1,\eta}^*/\alpha} \implies p^\star_{t+1,\eta}(x_{t+1}) = \frac{q_{t+1}(x_{t+1}) \E_{q_{0|t+1}}\sbrc{e^{-c_{\lambda,\eta}(x_0)/\alpha}}}{\E_{q_{0}}\sbrc{e^{-c_{\lambda,\eta}(x_0)/\alpha}}}. \]
Since $0\le c_{\lambda,\eta}(x_0)-a_\eta\le\alpha\xi/2$, we have
\[
\frac{p^\star_{t+1,\eta}(x_{t+1})}{q_{t+1}(x_{t+1})}
=
\frac{
\E_{q_{0|t+1}}
\left[e^{-(c_{\lambda,\eta}(x_0)-a_\eta)/\alpha}\right]
}{
\E_{q_0}
\left[e^{-(c_{\lambda,\eta}(x_0)-a_\eta)/\alpha}\right]
}
\le e^{\xi/2}.
\]

\paragraph{On the error caused by inaccurately solved $\eta$.}
We now turn to the first term in \eqref{eq:sampling-err-main}. Without loss of generality assume $\eta > \eta_\kappa^*$.
Define
\[
    \Delta_{\lambda,\eta}(x_0):=
    \frac{\lambda}{\alpha(1-\beta)}
    \brc{(c(x_0)-\eta_\kappa^*)^+-(c(x_0)-\eta)^+} \in \sbrc{0, \frac{\lambda(\eta-\eta_\kappa^*)}{\alpha(1-\beta)}},
\]
since $c(x_0) \in [l,u]$.
Also, by the definition of $c_{\lambda, \eta}$ in \eqref{eq:weighted-cost-def}, we have
\[ p_{\eta}^* (x_0) \propto \exp\left(-\frac{c_{\lambda, \eta}(x_0)}{\alpha}\right) q_0(x_0) \propto \exp\left(\Delta_{\lambda,\eta}(x_0)\right) p_{\eta_\kappa^*}^*(x_0). \]

Write $\partial_s(\cdot)$ for $\frac{\partial}{\partial s}(\cdot)$. Thus, we have
\begin{align*}
\TV{p_{\eta_\kappa^*}^\star}{p_\eta^*} &= \frac{1}{2} \int_{\calX} \abs{p_{\eta}^* (x_0) - p_{\eta_\kappa^*}^* (x_0)} \d x_0 \\
&= \frac{1}{2} \int_{\calX} \abs{\frac{e^{\Delta_{\lambda,\eta}(x_0)} }{\E_{p_{\eta_\kappa^*}^*}\sbrc{e^{\Delta_{\lambda,\eta}(x_0)}}} - 1} p_{\eta_\kappa^*}^* (x_0) \d x_0 \\
&= \frac{1}{2} \int_{\calX} \abs{\int_0^1 \partial_s \frac{e^{s \Delta_{\lambda,\eta}(x_0)} }{\E_{p_{\eta_\kappa^*}^*}\sbrc{e^{s \Delta_{\lambda,\eta}(x_0)}}} \d s} p_{\eta_\kappa^*}^* (x_0) \d x_0 \\
&\leq \frac{1}{2} \int_{\calX} \int_0^1 \abs{\partial_s \frac{e^{s \Delta_{\lambda,\eta}(x_0)} }{\E_{p_{\eta_\kappa^*}^*}\sbrc{e^{s \Delta_{\lambda,\eta}(x_0)}}} } \d s \cdot p_{\eta_\kappa^*}^* (x_0) \d x_0 \\
&\stackrel{(i)}{=} \frac{1}{2} \int_{\calX} \int_0^1 \frac{e^{s \Delta_{\lambda,\eta}(x_0)} }{\E_{p_{\eta_\kappa^*}^*}\sbrc{e^{s \Delta_{\lambda,\eta}(x_0)}}} \abs{ \Delta_{\lambda,\eta}(x_0) - \E_{p_{s}^*} \Delta_{\lambda,\eta}(x_0) } \d s \cdot p_{\eta_\kappa^*}^* (x_0) \d x_0 \\
&= \frac{1}{2} \int_0^1 \E_{p_s^*} \abs{ \Delta_{\lambda,\eta}(x_0) - \E_{p_{s}^*} \Delta_{\lambda,\eta}(x_0) } \d s  \\
&\stackrel{(ii)}{\leq} \frac{1}{2} \int_0^1 \sqrt{\Var_{p_s^*}(\Delta_{\lambda,\eta}) } \d s,
\end{align*}
where $(i)$ follows because
\[ \partial_s \frac{e^{s \Delta_{\lambda,\eta}(x_0)} }{\E_{p_{\eta_\kappa^*}^*}\sbrc{e^{s \Delta_{\lambda,\eta}(x_0)}}} = \frac{e^{s \Delta_{\lambda,\eta}(x_0)} }{\E_{p_{\eta_\kappa^*}^*}\sbrc{e^{s \Delta_{\lambda,\eta}(x_0)}}} \brc{ \Delta_{\lambda,\eta}(x_0) - \E_{p_{s}^*} \Delta_{\lambda,\eta}(x_0) }, \]
where $p_s^*(x_0) \propto \exp\left(s \Delta_{\lambda,\eta}(x_0)\right) p_{\eta_\kappa^*}^*(x_0)$, and $(ii)$ follows from Jensen's inequality.
Further, since $\Delta_{\lambda,\eta}(x_0)$ is bounded, we have
\[ \sqrt{\Var_{p_s^*}(\Delta_{\lambda,\eta})} \leq \frac{1}{2} \cdot \frac{\lambda(\eta-\eta_\kappa^*)}{\alpha(1-\beta)}, \]
and thus, under \eqref{assum:eta-optim-err}, we have
\begin{equation} \label{eq:sampling-err-eta-result}
    \TV{p_{\eta_\kappa^*}^\star}{p_\eta^*} \leq \frac{1}{4} \frac{\lambda(\eta-\eta_\kappa^*)}{\alpha(1-\beta)} \leq \frac{1}{4} \cdot \frac{\lambda \eps_\eta}{\alpha(1-\beta)}.
\end{equation}

Plugging \eqref{eq:sampling-err-value-result} and \eqref{eq:sampling-err-eta-result} into \eqref{eq:sampling-err-main}, we finally get
\[ \TV{p_{\eta_\kappa^\star}^\star (x_0)}{\hat{p}_{\eta}(x_0)} \leq \frac{e^{\xi/2}}{2 \alpha} \sqrt{ T } \eps_v + \frac{\lambda \eps_\eta}{4 \alpha(1-\beta)},\]
where $\xi$ is defined in \eqref{eq:overline-U-def}. The proof is now complete.

\section{Proofs of Auxiliary Lemmas}\label{app:auxiliary}

We provide the proofs of all auxiliary lemmas in this appendix.

\subsection{Proof of Lemma \ref{lem:eta-lipschitz}}
\label{app:lem:eta-lipschitz-proof}

For the first statement, recall that from \eqref{eq:weighted-cost-def}, we have
\[ c_{\lambda,\eta}(x_0)
    = (1-\lambda)c(x_0)
    + \lambda\brc{
        \eta+\frac{(c(x_0)-\eta)^+}{1-\beta}
    }. \]
Thus, the function
$\eta \mapsto c_{\lambda,\eta}(x_0)$ is continuous and
piecewise affine, with slope
$\lambda- \lambda /(1-\beta)$ on the interval $c(x_0)>\eta$ and slope $\lambda$ on the interval $c(x_0)\le\eta$.
Thus,
\[
    |c_{\lambda,\eta}(x_0)
      -c_{\lambda,\eta'}(x_0)|
    \leq L|\eta-\eta'|,
\]
with $L = \lambda\max\left\{1,\frac{\beta}{1-\beta}\right\}$.
Organizing the terms, we get, for each $x_0$,
\begin{align*}
    &-L|\eta-\eta'| \leq c_{\lambda,\eta'}(x_0)
      -c_{\lambda,\eta}(x_0) \leq L|\eta-\eta'| \\
    \implies & e^{-L|\eta-\eta'|/\alpha} \leq e^{c_{\lambda,\eta'}(x_0)/\alpha
      -c_{\lambda,\eta}(x_0)/\alpha } \leq e^{L|\eta-\eta'|/\alpha} \\
    \implies & e^{-L|\eta-\eta'|/\alpha} e^{-c_{\lambda,\eta'}(x_0)/\alpha} \leq e^{-c_{\lambda,\eta}(x_0)/\alpha} \leq e^{L|\eta-\eta'|/\alpha} e^{-c_{\lambda,\eta'}(x_0)/\alpha}
\end{align*}
Since this holds for each $x_0$, it also holds for both the true and the empirical average at $x_0$. To see this, consider the case with empirical average:
\begin{align*}
    &e^{-L|\eta-\eta'|/\alpha} \frac1N\sum_{i=1}^N e^{-c_{\lambda,\eta'}(x_0^i)/\alpha} \leq \frac1N\sum_{i=1}^N e^{-c_{\lambda,\eta}(x_0^i)/\alpha} \leq e^{L|\eta-\eta'|/\alpha} \frac1N\sum_{i=1}^N e^{-c_{\lambda,\eta'}(x_0^i)/\alpha} \\
    \implies & e^{-L|\eta-\eta'|/\alpha}
    \leq
    \frac{
        \frac1N\sum_{i=1}^N
        e^{-c_{\lambda,\eta}(x_0^i)/\alpha}
    }{
        \frac1N\sum_{i=1}^N
        e^{-c_{\lambda,\eta'}(x_0^i)/\alpha}
    }
    \leq
    e^{L|\eta-\eta'|/\alpha}.
\end{align*}
Taking logarithms and using the definition of $\widehat G_{\lambda,N}$ yields
\[
    |\widehat G_{\lambda,N}(\eta)
      -\widehat G_{\lambda,N}(\eta')|
    \leq L|\eta-\eta'|.
\]
Thus, $\widehat G_{\lambda,N}$ is $L$-Lipschitz (for every realization of the samples). For the true average, we simply replace all $\frac{1}{N} \sum_{i=1}^N (\cdot)$ with $\E_{q_0}[\cdot]$. This proves the first statement.

For the second statement, note that for $\eta,\eta',\eta_c\in[l,u]$, we have
\[ \frac{\kappa}{2}
    |(\eta-\eta_c)^2-(\eta'-\eta_c)^2| =
    \frac{\kappa}{2}|\eta-\eta'|
    |\eta+\eta'-2\eta_c| \leq
    \kappa(u-l)|\eta-\eta'|. \]
Thus, adding this bound to the vanilla objective yields the stated regularized Lipschitz constant (for both the true and the empirical average). This proves the second statement.

Finally, for the last statement, note that the regularization terms cancel:
\[
    \widehat G_{\lambda,\kappa,N}
    -G_{\lambda,\kappa}
    =
    \widehat G_{\lambda,N}-G_\lambda.
\]
This explains why the Lipschitz constant $2 L$ (rather than $2 L_\kappa$) suffices. The proof is now complete.

\subsection{Proof of Lemma \ref{lem:onedopt-query-complexity}}
\label{app:lem:onedopt-query-complexity-proof}

Let $\eta_1,\dots,\eta_{1+2K}\in[l,u]$ be the queried points. These may depend on the oracle
values, and hence on the samples. Let
$\mathcal{A}:=\{\sup_{\eta\in[l,u]}|\hat G_{\lambda,\kappa}(\eta)-G_{\lambda,\kappa}(\eta)|\le\epsilon/4\}$.
By assumption, $\mathbb{P}(\mathcal{A})\ge 1-\delta$. On $\mathcal{A}$, we have
$\max_{1\le i\le 1+2K}|\hat G_{\lambda,\kappa}(\eta_i)-G_{\lambda,\kappa}(\eta_i)|\le\epsilon/4$,
regardless of how the queried points are chosen. Applying Lemma~\ref{lem:onedimmin} with
$g=G_{\lambda,\kappa}$, Lipschitz constant $L_\kappa$, and $\tilde\epsilon:=\epsilon/4$, we obtain
that on $\mathcal{A}$, Algorithm~\ref{alg:onedopt} outputs an $\hat{\eta}$ such that
$G_{\lambda,\kappa}(\hat{\eta})-G_{\lambda,\kappa}(\eta_\kappa^*)=G_{\lambda,\kappa}(\hat{\eta})-\min_{\eta\in[l,u]}G_{\lambda,\kappa}(\eta)\le\epsilon$
by calling the inexact oracle $\hat G_{\lambda,\kappa}$ at most
$1+2\lceil\log_{3/2}(L_\kappa(u-l)/\tilde\epsilon)\rceil=1+2K$ times.
Furthermore, because $G_{\lambda,\kappa}(\eta)$ is $m_\kappa$-strongly convex and $\eta_\kappa^*$
minimizes it over $[l,u]$, we have
$|\hat{\eta}-\eta_\kappa^*|\le\sqrt{\frac{2}{m_\kappa}(G_{\lambda,\kappa}(\hat{\eta})-G_{\lambda,\kappa}(\eta_\kappa^*))}\le\sqrt{\frac{2\epsilon}{m_\kappa}}$.

\subsection{Proof of Lemma \ref{lem:saa-sample-complexity}}
\label{app:lem:saa-sample-complexity-proof}

\emph{Step 1: Pointwise bound.} Fix $\eta\in[l,u]$. Let $a_\eta:=(1-\lambda)l+\lambda\eta$,
$Y_\eta(x):=e^{-(c_{\lambda,\eta}(x)-a_\eta)/\alpha}$, $\mu_\eta:=\mathbb{E}[Y_\eta(X)\mid X_{T+1}=x_{T+1}]$,
and $\bar Y_{\eta,N}:=\frac1N\sum_{i=1}^N Y_\eta(x_0^i)$. The regularizer
$\frac\kappa2(\eta-\eta_c)^2$ is deterministic and cancels, and so do the factors $e^{-a_\eta/\alpha}$.
Hence
$|\hat G_{\lambda,\kappa,N}(\eta)-G_{\lambda,\kappa}(\eta)|=\alpha|\log(\bar Y_{\eta,N}/\mu_\eta)|$. Since
$c_{\lambda,\eta}-a_\eta=(1-\lambda)(c-l)+\frac{\lambda}{1-\beta}(c-\eta)^+\ge 0$, we have
$Y_\eta\in[0,1]$. By Jensen's inequality and $(c-\eta)^+\le(c-l)^+$, $\mu_\eta \ge\ e^{-\mathbb{E}[c_{\lambda,\eta}-a_\eta]/\alpha} \ge\ \underline{y}$.
Let $\nu:=\epsilon/(8\alpha)\le 1$ and $r:=1-e^{-\nu}$. If
$|\bar Y_{\eta,N}-\mu_\eta|\le r\mu_\eta$, then
$\mu_\eta e^{-\nu}\le\bar Y_{\eta,N}\le\mu_\eta(2-e^{-\nu})\le\mu_\eta e^{\nu}$, so
$|\hat G_{\lambda,\kappa,N}(\eta)-G_{\lambda,\kappa}(\eta)|\le\epsilon/8$. Since $|Y_\eta(x_0^i)-\mu_\eta|\le 1$ and
$\operatorname{Var}(Y_\eta(x_0^i))\le\mathbb{E}[Y_\eta(x_0^i)^2]\le\mathbb{E}[Y_\eta(x_0^i)]=\mu_\eta$,
Bernstein's inequality gives
\[
  \mathbb{P}\bigl(|\bar Y_{\eta,N}-\mu_\eta|>r\mu_\eta\bigr)
  \le 2\exp\Bigl(-\frac{Nr^2\mu_{\eta}^2}{2\mu_{\eta}+\frac23 r\mu_{\eta}}\Bigr)
  \le 2\exp\Bigl(-\frac{Nr^2\underline{y}}{2(1+r/3)}\Bigr)
  \le 2\exp\Bigl(-\frac{3N\underline{y}\,\epsilon^2}{2048\,\alpha^2}\Bigr),
\]
where the last step uses $r\le 1$ and $r\ge\frac{\nu}{1+\nu}\ge\frac\nu2$ (since $e^\nu\ge 1+\nu$ and $e^{-\nu}\le 1/(1+\nu)$). 
 
\emph{Step 2: Lipschitz continuity in $\eta$.} By Lemma~\ref{lem:eta-lipschitz}, the error
$\Delta(\eta):=\hat G_{\lambda,\kappa,N}(\eta)-G_{\lambda,\kappa}(\eta)$ is $2L$-Lipschitz in $\eta$.
 
\emph{Step 3: Uniform bound.} Let $H\subset[l,u]$ be a grid with spacing at most
$h:=\epsilon/(16L)$ and $|H|\le J:=\Bigl\lceil\frac{16L(u-l)}{\epsilon}\Bigr\rceil+1$. (If $\lambda=0$, then $L=0$, $\Delta$ does not depend on
$\eta$, and one point suffices.) By Step 1 and a union bound over $H$,
with probability at least $1-2J\exp(-\frac{3N\underline{y}\epsilon^2}{2048\alpha^2})\ge 1-\delta$,
we have $|\Delta(\eta')|\le\epsilon/8$ for all $\eta'\in H$. On this
event, for any $\eta\in[l,u]$, choose $\eta'\in H$ with $|\eta-\eta'|\le h$. By Step 2,
\[
  |\Delta(\eta)|
  \le|\Delta(\eta')|+2L|\eta-\eta'|
  \le\frac\epsilon8+\frac\epsilon8=\frac\epsilon4.
\]
The probability condition holds whenever $N\ge\frac{C_G}{\epsilon^2}\log\frac{2J}{\delta}$.

\subsection{Proof of Lemma \ref{lem:svdd-asymp}}
\label{app:lem:svdd-asymp-proof}

Fix $\eta$, $t\in\{1,\ldots,T+1\}$, and $x_t$. Let us first recall the SVDD algorithm in \Cref{alg:svdd}. We first draw $M$ candidates from the pretrained model:
\[
    Z=(Z_1,\ldots,Z_M),\qquad
    Z_i\overset{\mathrm{i.i.d.}}{\sim}\mu_t(\cdot\mid x_t).
\]
Then, we define the unnormalized resampling weight
\[
    w_{t-1}(z):=e^{-\hat V_{t-1,\eta}(z)/\alpha} > 0,
\]
where $\eta$ is omitted for brevity. Since the value estimates are bounded, there exist constants $0<a\le b<\infty$ such that $a\le w_{t-1}(z)\le b$ for all $z$. (While our result certainly extends beyond bounded $c(x_0)$, it suffices to assume it here, as in the rest of the paper.)
Finally, conditioned on $Z$, SVDD selects an index $I\in\{1,\ldots,M\}$ according to
\[
    \mbP(I=i\mid Z,x_t)
    =\frac{w_{t-1}(Z_i)}{\sum_{k=1}^M w_{t-1}(Z_k)},
\]
and returns $X=Z_I$ as the next state. 
Also define $\delta(z_i-x)$ as the delta function at $z_i$.

With this setup, for each $x$, we have
\begin{align*}
\hat\pi_{t,\eta}^{(M)}(x \mid x_t)
&=
\int
\mathbb{P}(x \mid I,Z)
\d \mathbb{P}(I \mid Z)
\d \mathbb{P}(Z \mid x_t)
\\
&=
\int_{\mathcal{Z}}
\left(
    \prod_{k=1}^{M}
    \mu_t(z_k \mid x_t)
\right)
\sum_{i=1}^{M}
\left(
    \frac{w_{t-1}(z_i)}
         {\sum_k w_{t-1}(z_k)}
    \delta(z_i-x)
\right)
\d z
\\
&=
\sum_{i=1}^{M}
\int_{\mathcal{Z}}
\left(
    \prod_{k=1}^{M}
    \mu_t(z_k \mid x_t)
\right)
\left(
    \frac{w_{t-1}(z_i)}
         {\sum_k w_{t-1}(z_k)}
    \delta(z_i-x)
\right) \d z
\\
&=
\sum_{i=1}^{M}
\mu_t(x \mid x_t)\,w_{t-1}(x)
\int_{\mathcal{Z}_{-i}}
\left(
    \prod_{k\neq i}
    \mu_t(z_k \mid x_t)
\right)
\frac{1}
     {w_{t-1}(x)+\sum_{k\neq i}w_{t-1}(z_k)}
\d z_{-i}
\\
&=
M\,\mu_t(x \mid x_t)\,w_{t-1}(x)
\mathbb{E}_{\mu_t}
\left[
    \frac{1}
         {w_{t-1}(x)+
          \sum_{k=1}^{M-1}w_{t-1}(z_k)}
\right],
\end{align*}
where the last line follows from symmetry over $i$.
Thus, by the strong law of large numbers, as $M \to \infty$,
\begin{equation*}
\frac{1}{M}
\left(
    w_{t-1}(x)
    +
    \sum_{k=1}^{M-1}w_{t-1}(z_k)
\right)
\xrightarrow{\mathrm{a.s.}}
\mathbb{E}_{\mu_t}
\left[
    w_{t-1}(Z_1)
\right].
\end{equation*}
Moreover, for every $M$,
\[
0<
\frac{M}{
w_{t-1}(x)+\sum_{k=1}^{M-1}w_{t-1}(Z_k)
}
\le \frac1a.
\]
Thus, by the dominated convergence theorem, we may pass
the limit through the expectation in the preceding expression
for $\widehat\pi^{(M)}_{t,\eta}$, obtaining
\begin{equation*}
\widehat\pi^{(M)}_{t,\eta}(x \mid x_t)
\longrightarrow
\frac{
    \mu_t(x \mid x_t)\,w_{t-1}(x)
}{
    \displaystyle
    \int
    \mu_t(z \mid x_t)\,w_{t-1}(z) \d z
} =
\widehat\pi_{t,\eta}(x\mid x_t),\quad \text{as}~M \to \infty.
\end{equation*}
The proof is now complete.

\section{Experimental Result Details}\label{app:experimental}

In this section, we include all the experimental details omitted in the main text.

\subsection{Additional details for synthetic experiments}
\label{app:synthetic}
We now provide the implementation details and additional results for the synthetic experiments in \Cref{sec:synthetic-main}. 

\subsubsection{Experimental Setup and Complete Results}\label{app:synthetic-setup}
We first provide the details for the synthetic experiment, as follows.

\paragraph{Base distribution} We use a two-dimensional base distribution, which is a two-component mixture of rotated product-Cauchy distributions: 
\begin{equation} 
X_0 = \mu_K + R \operatorname{diag}(\gamma) Z, \qquad Z_j \stackrel{\mathrm{i.i.d.}}{\sim} \operatorname{Cauchy}(0,1), \label{eq:mixed-cauchy-data} 
\end{equation} 
where $K \in \{1,2\}$, $\mathbb{P}\{K=1\}=0.4=1-\mathbb{P}\{K=2\}$ and
\[ \mu_1 = \begin{pmatrix} 0.543\\ -0.958 \end{pmatrix}, \quad \mu_2 = \begin{pmatrix} 0.267\\0.498 \end{pmatrix},\quad R = \begin{pmatrix} -0.868 & 0.496\\ 0.496 & 0.868 \end{pmatrix}, \quad \gamma = \begin{pmatrix} 0.296\\ 0.683 \end{pmatrix}.  \]
Both mixture components use the same principal-axis rotation and principal Cauchy scales. 

\paragraph{Cost function} The terminal objective is the hard-capped quadratic cost 
\begin{equation} 
c(x) = \min \left\{ \frac{ \lVert x-x^{\star}\rVert_2^2 }{ 2s^2 }, B \right\}, \qquad x^{\star}=(3,3), \quad s=3, \quad B=50. \label{eq:mixed-cauchy-cost} \end{equation} 
Note that the cost is continuous, nonnegative, and bounded in $[0,B]$. For every threshold $\tau < B$, its level set is a circle centered at $x^{\star}$ with radius 
\[ r(\tau) = s\sqrt{2\tau}. \] 
We also define $\tau_q$ (for $q \in \{80, 90, 95\}$) as the threshold for the $q$-th empirical quantile, which is estimated with an independent set of $50000$ samples from the frozen base sampler.

\paragraph{Sampling Parameters} We use a frozen DDPM-style sampler with $T=60$ reverse steps. We set the temperature to $\alpha=1$ and the CVaR confidence level to $\beta=0.8$. The vanilla method uses $\lambda=0$, whereas the risk-averse method uses $\lambda=1$. Both guided methods use $M=10$ candidate reverse transitions at each of the $T=60$ reverse steps. 
For each method, we generate $1000$ terminal samples under each of $10$ inference seeds.

\paragraph{Value Estimations} For a terminal cost $c(x_0)$, recall the definition of weighted cost in \eqref{eq:weighted-cost-def}. Following SVDD-MC~\cite{li2025svdd}, we use conditional-mean regression as a practical approximation to the soft value. Specifically, we train two separate reverse-look-ahead models using rollouts from the frozen base sampler: 
\[ m_{\phi}(x_t,t) \approx \mathbb{E}_\mu \left[ c(x_0) \mid x_t \right], \quad h_{\psi}(x_t,t,\eta) \approx \mathbb{E}_\mu \left[ (c(x_0)-\eta)^{+} \mid x_t \right]. \]
Specifically, we generate $15000$ training trajectories. From each trajectory, we retain several states and pair them with the terminal cost of this trajectory to perform training.
At inference time, given $\eta$, these models are combined through the value surrogate:
\begin{equation} \widehat V_{\lambda,\beta,\eta}(x_t,t) = \left[ (1-\lambda)m_{\phi}(x_t,t) + \lambda \left( \eta + \frac{h_{\psi}(x_t,t,\eta)}{1-\beta} \right) \right].
\label{eq:synthetic-combined-value} 
\end{equation} 
Then we used $\widehat V_{\lambda,\beta,\eta}$ for resampling, with candidate probabilities proportional to $\exp(-\widehat V_{\lambda,\beta,\eta}/\alpha)$. 
We compare \eqref{eq:synthetic-combined-value} with alternative value estimators in \Cref{fig:value-estimator-mixed-cauchy}, including estimators based on the exponential targets used in \Cref{thm:paired-value-estimation}. The conditional-mean construction in \eqref{eq:synthetic-combined-value} is a practical approximation to soft-value guidance and generally differs from the exact construction even when the conditional expectations are evaluated exactly.

\paragraph{Choosing $\eta$} The threshold $\eta$ used by CVaR-SVDD is selected by minimizing the fixed-sample empirical regularized objective in \eqref{eq:regularized-obj-eta}, with $\mathbb{E}_\mu$ replaced by $20000$ base-sample average drawn from the frozen base sampler. We set $\kappa=0.02$ and $\eta_c=1.752599$ to be the empirical VaR calculated from the pretrained samples.
We use $20000$ base samples and report the minimizer from \Cref{alg:onedopt}. A separate empirical global minimizer is adopted, which returns a numerically nearly identical value of $\eta$, indicating that the reported result is insensitive to the numerical solver used for the one-dimensional outer problem. We have provided an ablation study on the optimizer below.

\begin{figure*}[t]
\vspace{-3mm}
\centering
\includegraphics[width=0.98\linewidth]{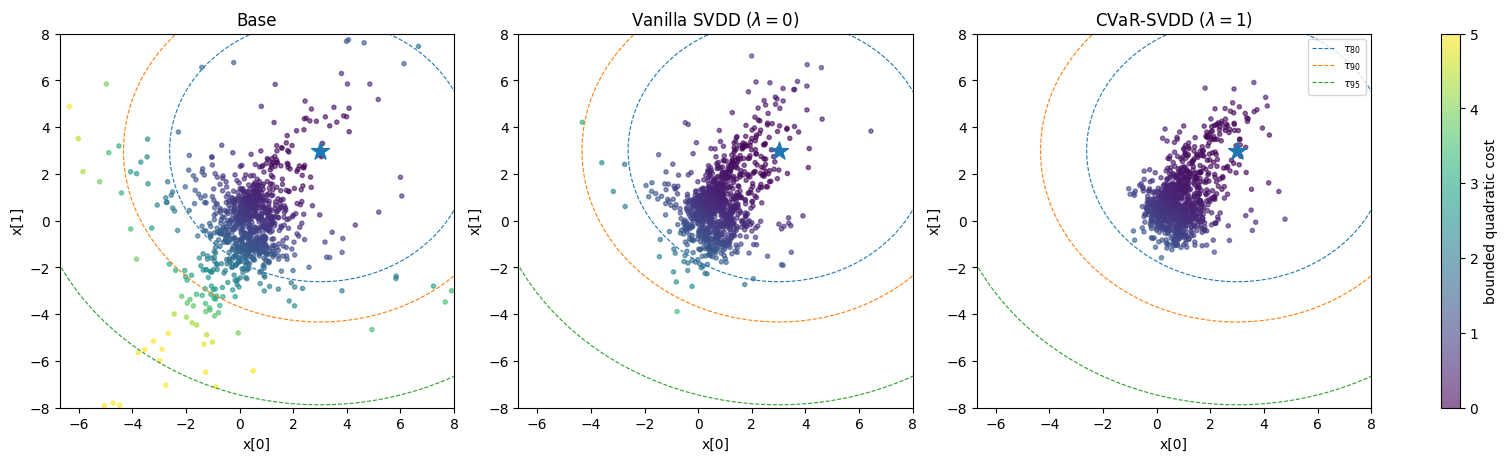}
\caption{ 
One realization of the terminal samples from the frozen base sampler, vanilla risk-neutral SVDD ($\lambda=0$), and CVaR-SVDD ($\lambda=1$) on the synthetic mixed-Cauchy example, using one representative random seed. The star denotes the target $x^{\star}=(3,3)$, the point colors denote the bounded quadratic cost, and the dashed circles denote the common cost thresholds $\tau_{80}$, $\tau_{90}$, and $\tau_{95}$.
}
\label{fig:mixed-cauchy}
\vspace{-3mm}
\end{figure*}

\paragraph{Complete Result}
\Cref{tab:synthetic-full} supplements the main-text comparison with median costs and exceedance probabilities at $\tau_{90}$ and $\tau_{95}$. In addition to the mean-cost and CVaR improvements reported in \Cref{sec:synthetic-main}, CVaR-SVDD lowers median cost from $0.6703$ to $0.6008$ relative to vanilla SVDD. At $\tau_{90}$, vanilla SVDD has an observed exceedance rate of $0.14\%$, whereas CVaR-SVDD has no observed exceedances. Neither guided method produces an exceedance at $\tau_{95}$. The terminal samples and survival curves in \Cref{fig:mixed-cauchy} provide complementary views of this shift toward lower costs.
Note that zero empirical exceedance rates indicate that no such events occurred within the available samples. Since the samples are random, they do not establish zero probabilities.
Overall, this experiment demonstrates that CVaR-SVDD can improve the full risk profile of the generated samples while using the same inference-time sampling budget as vanilla SVDD.

\begin{table}[t]
\centering
\caption{
Complete sampling performance of synthetic Cauchy data. Here $\tau_{90}$ and $\tau_{95}$ represent empirical VaR under the base model, calculated from $50000$ independent samples. Lower values are better throughout.
}
\label{tab:synthetic-full}
\footnotesize
\setlength{\tabcolsep}{4pt}
\renewcommand{\arraystretch}{1.15}
\begin{tabular}{@{}lcccccc@{}}
\toprule
Method
& Mean cost
& Median cost
& \makecell{Empirical\\CVaR}
& \makecell{$\hat{\mbP}(C>\tau_{80})$\\(\%)}
& \makecell{$\hat{\mbP}(C>\tau_{90})$\\(\%)}
& \makecell{$\hat{\mbP}(C>\tau_{95})$\\(\%)} \\
\midrule
Base
& \makecell{$2.373$\\$[2.126,\,2.621]$}
& \makecell{$1.004$\\$[0.9917,\,1.016]$}
& \makecell{$8.379$\\$[7.190,\,9.569]$}
& \makecell{$19.37$\\$[17.97,\,20.77]$}
& \makecell{$9.760$\\$[8.791,\,10.73]$}
& \makecell{$4.800$\\$[4.208,\,5.392]$} \\
\addlinespace
SVDD \cite{li2025svdd}
& \makecell{$0.7057$\\$[0.7000,\,0.7114]$}
& \makecell{$0.6703$\\$[0.6601,\,0.6806]$}
& \makecell{$1.343$\\$[1.319,\,1.367]$}
& \makecell{$1.950$\\$[1.507,\,2.393]$}
& \makecell{$0.1400$\\$[0.07968,\,0.2003]$}
& \textbf{0 observed} \\
\addlinespace
CVaR-SVDD
& \makecell{$\mathbf{0.5865}$\\$[0.5814,\,0.5917]$}
& \makecell{$\mathbf{0.6008}$\\$[0.5937,\,0.6078]$}
& \makecell{$\mathbf{1.017}$\\$[1.003,\,1.031]$}
& \textbf{0 observed}
& \textbf{0 observed}
& \textbf{0 observed} \\
\bottomrule
\end{tabular}
\end{table}

We also note that the improvement in the overall cost is indeed an added benefit, which depends heavily on the structures of both the underlying target distribution and the cost function. In particular, CVaR guidance does not distinguish among samples whose costs lie below the optimized threshold $\eta$. For example, when the distribution of $c(x_0)$ has substantial probability mass concentrated right below the corresponding VaR level, this cut-off effect can leave many generated samples in the moderate-cost region near the threshold. In this case, CVaR-SVDD can achieve better control of the upper tail, while possibly yielding a higher overall average cost than vanilla SVDD. 
Therefore, better average cost should be viewed as an additional distribution-dependent benefit rather than an intrinsic consequence of the CVaR objective.

\subsubsection{Synthetic Ablations}
\label{app:synthetic-ablations}

We now provide some ablation studies using the synthetic data.

\paragraph{Ablation on the sampler: SMC.}
We further test whether the benefit of our risk-averse objective depends on the particular inference-time sampler. In addition to SVDD, we consider a Sequential Monte Carlo (SMC) sampler, which differs from SVDD in its resampling mechanism. While SVDD draws multiple candidate transitions for each current sample and performs resampling locally within the corresponding candidate set, SMC propagates each particle once through the pretrained reverse transition and performs resampling globally over the entire particle population. We have provided a reference pseudo-code in \Cref{alg:smc}. 

For the experiment, we use the same frozen diffusion model, learned mean- and hinge-value models, optimized $\eta$, CVaR confidence level $\beta=0.8$, and temperature $\alpha=1$. We compare vanilla SMC with $\lambda=0$ against CVaR-SMC with $\lambda=1$. The number of SMC particles is set equal to the number of generated samples (i.e., $N = 10000$), and the ESS resampling threshold is fixed at $\gamma=0.5$. The result is averaged over 5 random seeds.

\begin{figure*}[t]
\vspace{-3mm}
\centering
\begin{minipage}{0.47\linewidth}
    \centering
    \includegraphics[
        width=\linewidth
    ]{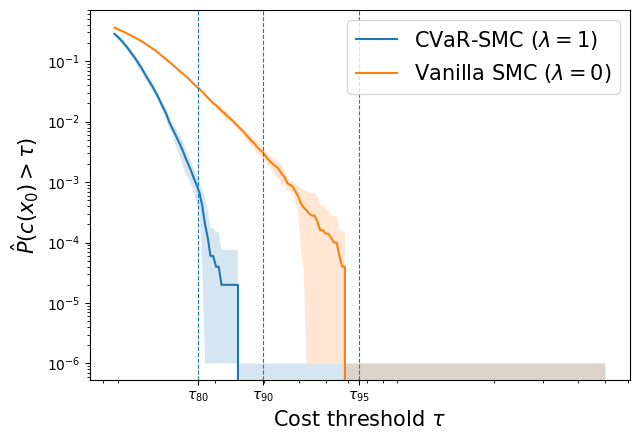}
\end{minipage}
\hfill
\begin{minipage}{0.47\linewidth}
    \centering
    \includegraphics[
        width=\linewidth
    ]{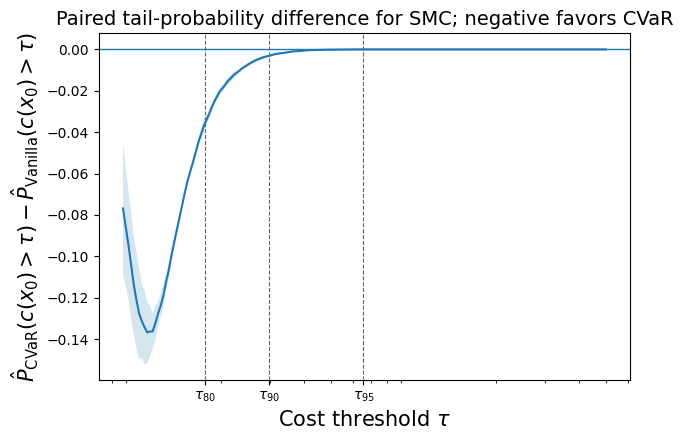}
\end{minipage}
\caption{
Cost survival curves for the synthetic mixed-Cauchy data using the SMC sampler in \Cref{alg:smc}.
\emph{Left:} empirical survival probabilities $\widehat{\mbP}\{c(x_0)>\tau\}$ for vanilla SMC ($\lambda=0$) and CVaR-SMC ($\lambda=1$).
\emph{Right:} paired difference $\widehat{\mbP}_{\mathrm{CVaR\text{-}SMC}}\{c(x_0)>\tau\} - \widehat{\mbP}_{\mathrm{Vanilla\text{-}SMC}}\{c(x_0)>\tau\}$.
Negative values indicate an improvement from CVaR guidance. The dashed vertical lines indicate the common thresholds $\tau_{80}$, $\tau_{90}$, and $\tau_{95}$.
}
\label{fig:smc-ablation-mixed-cauchy}
\vspace{-3mm}
\end{figure*}

The results in \Cref{fig:smc-ablation-mixed-cauchy} are qualitatively consistent with those obtained using SVDD in \Cref{fig:mixed-cauchy}. CVaR-SMC produces a uniformly lower survival curve than vanilla SMC over essentially the full range considered. In particular, the paired CVaR-minus-vanilla difference remains negative around $\tau_{80}$ and approaches zero only after both methods have already driven the corresponding tail probabilities to very small values. Since the behavior closely parallels the SVDD results, we omit a repetition of the detailed tail analysis here.

This experiment provides evidence that the improvement from CVaR guidance is not specific to the local candidate-resampling mechanism of SVDD. The same risk-adjusted value function can instead be used as a potential in a global SMC resampling procedure and yields a similar reduction in high-cost outcomes. Therefore, our framework is compatible with different inference-time sampling mechanisms: the CVaR construction determines the risk-sensitive change of measure, whereas SVDD and SMC provide different numerical procedures for approximately realizing this tilted distribution.

\paragraph{Ablation on the temperature $\alpha$.}
We next study the effect of varying the SVDD temperature $\alpha$. Recall the optimal path-wise distribution in \Cref{thm:risk-averse-reduction}.
Specifically, a small value of $\alpha$ produces a sharper and more nearly deterministic preference for candidates with high estimated utility, whereas a large value of $\alpha$ flattens the resampling probabilities and moves the sampler closer to the unguided transition.

\begin{figure*}[t]
\centering
\begin{minipage}{0.47\linewidth}
    \centering
    \includegraphics[
        width=\linewidth
    ]{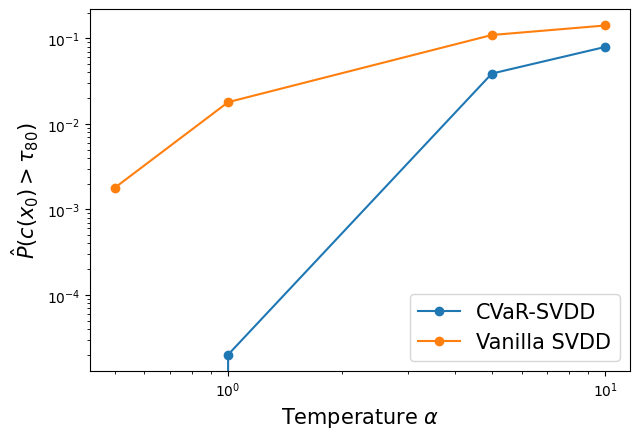}
\end{minipage}
\hfill
\begin{minipage}{0.47\linewidth}
    \centering
    \includegraphics[
        width=\linewidth
    ]{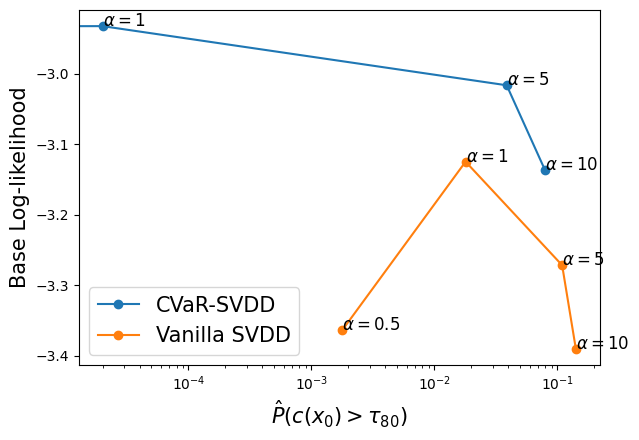}
\end{minipage}
\caption{
Ablation on the temperature $\alpha$ for vanilla SVDD ($\lambda=0$) and CVaR-SVDD ($\lambda=1$).
\emph{Left:} Empirical bad-sample probability $\widehat\mbP\{c(x_0)>\tau_{80}\}$ as a function of $\alpha$. Both axes are logarithmic. Note that the CVaR point at the smallest temperature has zero observed exceedances and therefore cannot be displayed on the logarithmic vertical axis.
\emph{Right:} Base log-likelihood as a function of tail probability. Each point is annotated by its value of $\alpha$. Higher values on the vertical axis indicate that the generated samples lie, on average, in higher-density regions of the pretrained base distribution.
}
\label{fig:alpha-ablation-mixed-cauchy}
\vspace{-3mm}
\end{figure*}

In \Cref{fig:alpha-ablation-mixed-cauchy}, we compare vanilla SVDD, corresponding to $\lambda=0$, with pure CVaR-SVDD ($\lambda=1$) over $\alpha \in \{0.5,1,5,10\}$. For both methods, we fix the CVaR confidence level at $\beta=0.8$ and keep the same sampling parameters otherwise. 

The left panel of \Cref{fig:alpha-ablation-mixed-cauchy} shows that the bad-sample probability decreases with smaller $\alpha$ for both methods, with CVaR-SVDD achieving a better tail probability at the same temperature. 
In the right panel of
\Cref{fig:alpha-ablation-mixed-cauchy}, we investigate sample fidelity at different temperature values. In particular, we plot the sample-averaged log-likelihood $\widehat{\mathcal L}_{q_0} := \frac{1}{N} \sum_{i=1}^{N} \log q_0(x_0^i)$ as a function of the tail probability.
Note that a larger value of $\widehat{\mathcal L}_{q_0}$ means that the generated samples lie, on average, in regions assigned higher density by the pretrained base model.
Compared to vanilla SVDD, CVaR-SVDD exhibits a favorable empirical frontier.

\paragraph{Ablation on the risk weight $\lambda$.}
We investigate how the interpolation weight $\lambda$ affects tail performance. We vary $\lambda\in\{0,0.25,0.5,0.75,1\}$ while fixing the other hyperparameters. For each nonzero $\lambda$, we optimize $\eta$ using the same fixed set of base-model costs. We then generate $10000$ samples for each of five inference seeds and use the same evaluation thresholds across all values of $\lambda$.

\begin{table*}[t]
\centering
\caption{Effect of the risk weight $\lambda$ on synthetic sampling performance. Empirical CVaR is reported as the mean [95\% confidence interval] across five inference seeds. Exceedance probabilities are seed averages. The threshold $\eta$ has no effect when $\lambda=0$ and it is thus omitted.}
\label{tab:lambda-ablation}
\small
\begin{tabular}{@{}cccc@{}}
\toprule
$\lambda$ & $\widehat\eta$ & Empirical CVaR $\downarrow$
& $\widehat{\mathbb P}\{c(x_0)>\tau_{80}\}$ $\downarrow$ \\
\midrule
$0$    & ---        & $1.346\;[1.327,1.365]$ & $0.01796$ \\
$0.25$ & $1.075862$ & $1.192\;[1.181,1.203]$ & $0.00454$ \\
$0.50$ & $1.022672$ & $1.116\;[1.106,1.125]$ & $0.00088$ \\
$0.75$ & $0.982972$ & $1.065\;[1.055,1.076]$ & $0.00022$ \\
$1$    & $0.955166$ & $1.031\;[1.023,1.039]$ & $0.00004$ \\
\bottomrule
\end{tabular}
\end{table*}

As shown in \Cref{tab:lambda-ablation}, increasing $\lambda$ progressively reduces both empirical CVaR and the exceedance probability at $\tau_{80}$. 
The intermediate values of $\lambda$ produce intermediate tail performance, illustrating how the mean-CVaR interpolation provides a tunable degree of tail control.

\paragraph{Ablation on the regularization strength $\kappa$.}
We investigate how quadratic threshold regularization affects mixed-Cauchy CVaR-SVDD. 
We sweep $\kappa\in\{0, 0.0002, 0.002, 0.02, 0.06, 0.2, 2\}$, holding $\lambda=1$, $\beta=0.8$, $\alpha=1$, $T=60$, and $M=10$ fixed. 
Note that we use a separately trained hinge-value network shared across all regularization settings, which differs from that in the main experiment.
We report averages over 5 seeds and $95\%$ Student-$t$ confidence intervals. Empirical CVaR is calculated directly from each run's highest-cost $20\%$, independently of the optimized threshold.

\begin{figure}[t]
    \centering
    \includegraphics[width=0.5\linewidth]{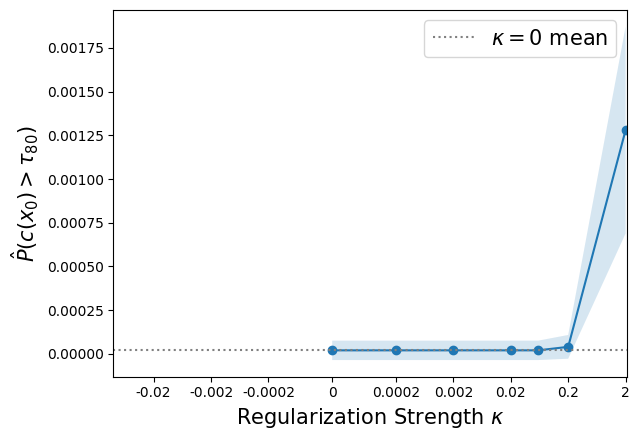}
    \caption{
    Effect of the regularization strength $\kappa$ on the
    empirical exceedance probability $\widehat{\mathbb{P}}\{c(x_0)>\tau_{80}\}$ for CVaR-SVDD.
    Results are averaged over five random seeds, with shaded regions indicating $95\%$ Student-$t$ confidence intervals. The dotted horizontal line marks the mean at $\kappa=0$.
    }
    \label{fig:kappa-ablation}
\end{figure}

\begin{table*}[t]
\centering
\small
\caption{
Effect of the regularization strength $\kappa$ on CVaR-SVDD.
Performance metrics are reported as means [$95\%$ confidence
intervals] across five random seeds.
}
\label{tab:kappa-ablation}
\setlength{\tabcolsep}{6pt}
\begin{tabular}{ccccc}
\toprule
$\kappa$ & $\widehat{\eta}_\kappa$ & Mean cost & Empirical CVaR
\\
\midrule
$0$
& $0.950100$
& $0.6250\;[0.6189,\,0.6311]$
& $1.036\;[1.028,\,1.045]$ \\
$0.0002$
& $0.950100$
& $0.6250\;[0.6189,\,0.6311]$
& $1.036\;[1.028,\,1.045]$ \\
$0.002$
& $0.950492$
& $0.6251\;[0.6190,\,0.6312]$
& $1.037\;[1.028,\,1.045]$ \\
$0.02$
& $0.955166$
& $0.6261\;[0.6200,\,0.6323]$
& $1.038\;[1.030,\,1.046]$ \\
$0.06$
& $0.967847$
& $0.6289\;[0.6227,\,0.6351]$
& $1.042\;[1.034,\,1.050]$ \\
$0.2$
& $1.022609$
& $0.6405\;[0.6345,\,0.6465]$
& $1.061\;[1.054,\,1.069]$ \\
$2$
& $1.425546$
& $0.7198\;[0.7128,\,0.7268]$
& $1.199\;[1.188,\,1.210]$ \\
\bottomrule
\end{tabular}
\end{table*}

In \Cref{fig:kappa-ablation,tab:kappa-ablation}, it is observed that small regularization coefficients preserve the observed tail performance. 
In particular, the default $\kappa=0.02$ increases mean cost and empirical CVaR by only $0.18\%$ and $0.19\%$, respectively, relative to the unregularized case. Larger penalties (e.g., $\kappa=2$), however, produce more pronounced changes. This is expected, as stronger regularization pulls the optimized threshold toward the larger reference value $\eta_c$, raising the threshold at which the hinge penalty becomes active and consequently weakening tail control.

\paragraph{Ablation on value estimation.}
We compare four value estimators obtained by combining
conditional-mean/hinge or normalized-exponential regression targets
with neural networks or clipped linear regressors. We retain the same setup. Results are averaged over five inference seeds, with $10000$ samples per method per seed.

\begin{figure}[t]
    \centering
    \includegraphics[width=0.5\linewidth]{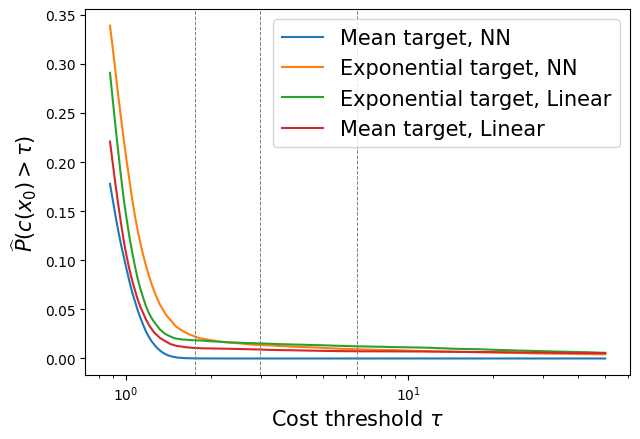}
    \caption{Value-estimator ablation on the synthetic example. We compare conditional-mean/hinge and normalized-exponential targets using neural networks (NN) and clipped linear regressors with fixed bounded features (Linear). Curves show empirical cost survival probabilities averaged over five inference seeds. Dashed vertical lines mark the base-model cost thresholds $\tau_{80}$, $\tau_{90}$, and $\tau_{95}$. Lower curves indicate fewer high-cost outcomes.}
    \label{fig:value-estimator-mixed-cauchy}
\end{figure}

As shown in \Cref{fig:value-estimator-mixed-cauchy}, the conditional-mean/hinge neural estimator provides the strongest empirical tail control among the compared methods. 
This result supports the practical surrogate used in our experiments.

\paragraph{Ablation on the optimization of $\eta$.}
We also investigate the numerical behavior of the one-dimensional outer optimization over the CVaR threshold $\eta$. The objective used in this experiment is the sample-averaged version of the objective in \eqref{eq:regularized-obj-eta}.
Specifically, we use the same fixed set of $N_{\eta}=20000$ base-model costs in every objective evaluation, with $\kappa=0.02$, $\lambda=1$, $\beta=0.8$, and $\alpha=1$.

\begin{figure*}[t]
\vspace{-3mm}
\centering
\begin{minipage}{0.47\linewidth}
    \centering
    \includegraphics[
        width=\linewidth
    ]{Figures/optimizer-eta-mixed-cauchy.png}
\end{minipage}
\hfill
\begin{minipage}{0.47\linewidth}
    \centering
    \includegraphics[
        width=\linewidth
    ]{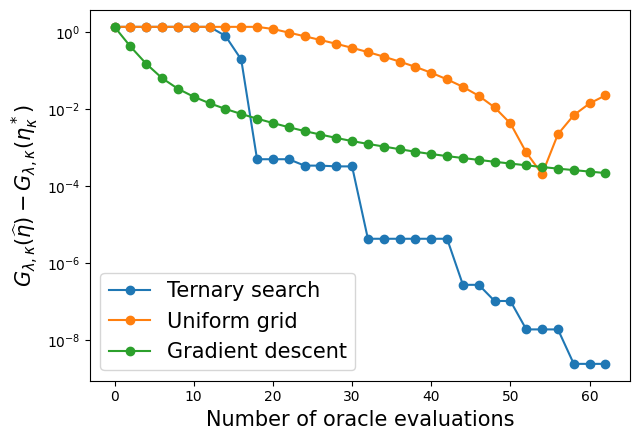}
\end{minipage}
\caption{
Full ablation on the numerical optimization of the regularized CVaR threshold $\eta$. The empirical global minimizer $\eta_\kappa^\star$ is used as reference. 
\emph{Left:} absolute eta error $|\widehat\eta-\eta_\kappa^\star|$ versus the number of oracle evaluations. 
\emph{Right:} empirical objective gap
$\widehat G_{\lambda,\kappa}(\widehat\eta) - \widehat G_{\lambda,\kappa}(\eta_\kappa^\star)$ versus the same oracle-evaluation budget. 
The uniform-grid baseline uses the same number of grid evaluations as the corresponding ternary-search budget of \Cref{alg:onedopt}. For gradient descent, one joint evaluation of the objective and its analytic derivative is counted as one oracle evaluation, and the reported curve is averaged over five randomly sampled learning rates.
}
\label{fig:eta-optimizer-ablation}
\vspace{-3mm}
\end{figure*}

In \Cref{fig:eta-optimizer-ablation}, we compare the CVaR-Optimizer in \Cref{alg:onedopt} with two baselines: uniform-grid search and projected gradient descent. All three methods optimize the same fixed-sample empirical objective $\widehat G_{\lambda,\kappa}(\eta)$ and use the same search interval $[\ell,u] = [ \min_i c(x_0^i), \max_i c(x_0^i) ]$.
We use a global minimizer $\eta_\kappa^\star$ of
$\widehat G_{\lambda,\kappa}$ over $[\ell,u]$
as the reference solution.

Different from the ternary search approach in \Cref{alg:onedopt}, which adaptively contracts the search interval according to objective evaluations at two interior points, the uniform-grid baseline evaluates $\widehat G_{\lambda,\kappa}(\eta)$ at equally spaced points over the entire interval $[\ell,u]$ and returns the point attaining the smallest objective value. For each point in \Cref{fig:eta-optimizer-ablation}, the number of grid points is chosen to match the number of oracle evaluations used by ternary search, so that the two methods have the same evaluation budget.
For gradient descent, we use the analytic derivative of the empirical objective. In particular, for $\lambda=1$, away from empirical cost atoms,
\begin{equation*}
    \widehat G_{1,\kappa}'(\eta)
    =
    1
    -
    \frac{1}{1-\beta}
    \sum_{i=1}^{N_\eta}
    w_i(\eta)
    \mathbf{1}\{c(x_0^i)>\eta\} +\kappa(\eta-\eta_c),
\end{equation*}
where
\begin{equation*}
    w_i(\eta)
    =
    \frac{
        \exp\left(
            -\frac{(c(x_0^i)-\eta)^+}
            {\alpha(1-\beta)}
        \right)
    }{
        \sum_{j=1}^{N_\eta}
        \exp\left(
            -\frac{(c(x_0^j)-\eta)_+}
            {\alpha(1-\beta)}
        \right)
    }.
\end{equation*}
We apply projected gradient descent so that every iterate remains inside $[\ell,u]$. The initialization is set to be the base $\beta$-quantile. To reduce sensitivity to the learning-rate choice, we randomly sample five learning rates log-uniformly from $[0.05,0.5]$ and report the average performance across these five runs. We use a diminishing step size proportional to $1/\sqrt{k}$ at iteration $k$. Since the objective and its derivative can be computed from the same tilted weights, one joint objective-gradient computation is counted as one oracle evaluation. Thus, all methods in \Cref{fig:eta-optimizer-ablation} are compared as a function of the same oracle evaluation budget.

As shown in the left panel of \Cref{fig:eta-optimizer-ablation}, the minimizer by \Cref{alg:onedopt} rapidly approaches the exact empirical solution. Its error initially remains nearly constant while the search interval is broad and the incumbent is located far from $\eta_\kappa^\star$. Once the retained interval enters the neighborhood of the empirical minimizer, the error decreases in a sequence of sharp steps. The plateaus in the curve are expected because we report the best incumbent found so far: a ternary-search iteration can substantially shrink the retained interval without changing the incumbent whenever neither newly evaluated point improves the current best objective value.
\underline{\textit{(a) Comparison with gradient descent:}} By contrast, gradient descent improves much more smoothly. It quickly reduces the eta error initially and then continues to improve gradually as the oracle budget increases. However, its convergence is substantially slower than that of \Cref{alg:onedopt} near the optimum. At the largest displayed budget, the averaged gradient-descent error remains on the order of $10^{-2}$, whereas that of \Cref{alg:onedopt} reaches an error on the order of $10^{-5}$. This difference is particularly notable because gradient descent has access to first-order derivative information, while \Cref{alg:onedopt} uses only objective evaluations.
\underline{\textit{(b) Comparison with uniform-grid:}} Also, the uniform-grid baseline is substantially less efficient over most of the evaluation range. Since it allocates evaluations uniformly over the entire initial interval, many evaluations are spent far from the minimizer. Its error also varies non-monotonically with the number of grid points because successive uniform grids are not nested: increasing the grid size changes the locations of the interior points, and hence a larger grid does not necessarily contain the best point from the previous grid. Although the uniform grid eventually reaches an eta error comparable to gradient descent at one favorable budget, its error increases again when the grid locations shift.

The right panel of \Cref{fig:eta-optimizer-ablation} exhibits the same behavior more clearly in terms of objective suboptimality. \Cref{alg:onedopt} reduces $\widehat G_\lambda(\widehat\eta)-\widehat G_\lambda(\eta_\kappa^\star)$ by many orders of magnitude and eventually reaches the numerical precision floor. Gradient descent decreases the objective gap steadily, but remains several orders of magnitude above ternary search at the largest evaluation budget. Uniform-grid search again converges more slowly and exhibits the same non-monotone behavior caused by the changing grid locations.

Overall, the experiment shows that \Cref{alg:onedopt} provides substantially more accurate optimization of $\eta$ under a fixed evaluation budget than both uniform-grid search and gradient descent on this mixed-Cauchy example. In particular, \Cref{alg:onedopt} achieves much smaller error in both the recovered threshold $\eta$ and the empirical regularized objective value, despite being derivative-free. These results empirically support the convergence behavior of the $\eta$-optimization procedure.

\subsection{Additional details for image experiments}
\label{app:image}

We now provide the implementation details and additional results for the image experiments in \Cref{sec:image-main}. 

\subsubsection{Experimental Setup}
\label{app:image-setup}

We first provide the setup for the image experiment, as follows. The value networks are trained on an NVIDIA H100 GPU, while inference-time sampling is performed on an NVIDIA RTX PRO 6000 Blackwell Server Edition GPU (G4) through Google Colab.

\paragraph{Base diffusion model}
We use the Stable Diffusion v1.5 \cite{rombach2022stable-diffusion}, with checkpoint from Huggingface (\path{runwayml/stable-diffusion-v1-5}), to generate RGB images of resolution $512\times512$ from latents of shape $4\times64\times64$. Sampling uses $T=50$ reverse steps, classifier-free guidance scale $7.5$, and DDIM stochasticity $\eta_{\mathrm{DDIM}}=1$ (which is essentially DDPM \cite{ho2020ddpm}). The diffusion U-Net, VAE decoder, and text encoder are frozen and run in evaluation mode. 

\paragraph{Reward functions}
Both reward functions are employed in \cite{li2025svdd,black2024ddpo,fan2023dpok}. Specifically, 
For compressibility, the reward is defined as the negative file size after JPEG compression, and the cost is defined as the negative reward. 
For aesthetics, we use LAION Aesthetic Predictor V2 \cite{schuhmann2022aesthetic} as the reward (and the negative cost) function, which predicts human aesthetic ratings of generated images, with higher scores indicating greater aesthetic appeal.

\paragraph{Value estimates}
Let $x_t$ denote the current state at time $t$ and $p$ the chosen text prompt. Vanilla SVDD-MC uses a learned terminal-reward estimate, written in cost form as $\widehat m(x_t,t,p)$, and our learned CVaR estimator approximates the corresponding positive part:
\begin{equation}
\label{eq:image-values}
    \widehat m(x_t,t,p) \approx \mathbb E_\mu[c(x_0)\mid x_t,t,p],\quad \widehat h(x_t,t,p,\eta) \approx \mathbb E_\mu [(c(x_0)-\eta)^+\mid x_t,t,p]. 
\end{equation}
As in the synthetic experiment, we follow the conditional-mean approximation of SVDD-MC~\cite{li2025svdd}.

\paragraph{Training prompts and labels}
We use the animal prompts from \cite{li2025svdd,black2024ddpo}:
\[
    \{\texttt{cat}, \texttt{dog}, \texttt{horse}, \texttt{monkey}, \texttt{rabbit}, \texttt{butterfly}, \texttt{panda}\}.
\]
Prompts are sampled uniformly with replacement within training minibatches. Frozen base-model trajectories supply the terminal labels. At a recorded state, the mean-reward model is trained against $r(x_0)$, while the CVaR model is trained against $(c(x_0)-\eta)^+$. One $\eta$ is sampled per trajectory, uniformly from a prompt-specific interval, and is shared by the recorded states on that trajectory. Squared-error losses are averaged over samples and recorded states.

\paragraph{Choosing the inference threshold}
For each prompt and reward, we optimize for $\eta$ on a separate set of $N_\eta=256$ base-model samples, which are generated to approximate $\widehat G_1(\eta)$ (with $\lambda=1$ and $\kappa=0$):
\begin{equation}
    \widehat G_{1,N_\eta}(\eta)
    =\eta-\alpha\log\left(
       \frac{1}{N_\eta}\sum_{i=1}^{N_\eta}
       \exp\left\{-\frac{(c(x_0^i)-\eta)^+}{\alpha(1-\beta)}\right\}
    \right).
\label{eq:image-eta-calibration}
\end{equation}
Note that the empirical quantile cutoff discussed below is also derived from these generated samples. 
To obtain these base-model samples, we follow \cite{li2025svdd} and use deterministic DDIM sampling ($\eta_{\mathrm{DDIM}}=0$), motivated by its reported sample-quality advantage over DDPM at reduced sampling budgets \cite{song2021ddim}.
We also verified that transferring $\eta_{\mathrm{DDIM}}=0$ to $\eta_{\mathrm{DDIM}}=1$ incurs small objective gaps of only $1.558$, $1.318$, and $8.110$ for \texttt{cat}, \texttt{dog}, and \texttt{rabbit}, respectively.

After obtaining \eqref{eq:image-eta-calibration}, we compute an empirical global minimizer by evaluating all observed cost knots together with a uniform grid of $2001$ points on the empirical cost range. For this unregularized objective, we can indeed show that a global minimum is attained at an empirical cost knot, since the objective is concave between consecutive knots. With $N_\eta=256$, this scalar search is inexpensive.
We also verified that the ternary-search procedure produces thresholds very close to the empirical global minimizers.
Specifically, ternary search achieves an absolute difference in the empirical threshold of below $1.6\times10^{-6}$ after 40 iterations for both \texttt{dog} and \texttt{rabbit} and below $1.2\times10^{-5}$ after 80 iterations for \texttt{cat}.
Thus, the procedures produce nearly identical thresholds in all the scenarios considered.

\paragraph{Sampling Hyperparameters}
We compare three inference-time guided methods: the vanilla SVDD-MC sampler \cite{li2025svdd}, the CVaR-SVDD(-MC) sampler. We further introduce a \textit{plug-in CVaR} alternative that directly applies the hinge-type transformation to the vanilla value estimate. They share the same pretrained sampling step, but they differ in how the resampling weights are chosen. Specifically, let these weights be proportional to $\exp(\ell(x_t))$, where
\begin{equation}
\label{eq:image-logweights}
    \ell_{\mathrm{vanilla}}(x_t)=-\widehat m(x_t,t,p)/\alpha, \quad \ell_{\mathrm{CVaR}}(x_t)=-\frac{\widehat h(x_t,t,p,\eta)}{\alpha(1-\beta)},\quad \ell_{\mathrm{plugin}}(x_t) =-\frac{(\widehat m(x_t,t,p)-\eta)^+}{\alpha(1-\beta)}.
\end{equation}
Note that the two CVaR methods use the pure tail-risk objective, corresponding to $\lambda=1$, and share the calculated $\eta$. 
At each of the reverse sampling step before the last one, SVDD generates $M=20$ candidate transitions per current state and selects one with probability proportional to the exponential of its log weight in \eqref{eq:image-logweights}.
All compressibility experiments use $\alpha=5$, and all aesthetic experiments use $\alpha=0.5$. For the two CVaR-based methods, we further set $\beta=0.8$.

We also conduct ablation studies with the SMC sampler \cite{singhal2025fksteering} to assess the effectiveness of our framework beyond SVDD.
Each seed generates $16$ independently initialized populations of $P=32$ particles. 
Following \Cref{alg:smc}, we initialize the particle weights using the value potential at the initial noise state and update them by consecutive potential differences during reverse sampling. At the terminal step, we evaluate the exact transformed cost $c_{\lambda,\eta}(x_0)$ and include its potential $-c_{\lambda,\eta}(x_0)/\alpha$ before the final multinomial resampling. See \Cref{alg:smc} for details.

Each run generates $512$ images per method per inference seed in batches of $32$. The compressibility \texttt{cat}, aesthetic \texttt{cat}, and compressibility \texttt{cat} SMC results use five seeds per method. The prompt ablations on \texttt{dog} and \texttt{rabbit} use three seeds instead.
Initial latents are shared across methods within each seed.
Value networks and the threshold-optimization set are fixed across the inference seeds.

\subsubsection{Complete results and ablations}
\label{app:image-results}

\Cref{tab:image-full} reports the mean and tail-average rewards for all evaluated scenarios, averaged across inference seeds with standard errors rather than the $95\%$ confidence intervals used in the preceding tables.

\begin{table*}[tp]
\centering
\caption{Complete image results in \textbf{reward} (i.e., negative cost). Each entry is the mean $\pm$ standard error of seed-level metrics. All values are rewards and larger is better. The ``lower-tail mean reward'' evaluates the average rewards at or below that run's empirical $20$th-percentile cutoff (including ties). The best result for each metric within each block is boldfaced.}
\label{tab:image-full}
\small
\setlength{\tabcolsep}{8pt}
\begin{tabular}{@{}lllrr@{}}
\toprule
Reward & Prompt & Method & Mean reward $\uparrow$ & Lower-tail mean reward $\uparrow$ \\
\midrule
Compressibility & \texttt{cat} & Base & $-95.986\pm0.224$ & $-123.377\pm1.042$ \\
& & Vanilla SVDD & $-73.034\pm0.121$ & $-89.671\pm0.292$ \\
& & CVaR-SVDD & $\boldsymbol{-67.136\pm0.131}$ & $\boldsymbol{-78.405\pm0.135}$ \\
& & Plug-in CVaR-SVDD & $-71.401\pm0.131$ & $-83.434\pm0.279$ \\
\midrule
Compressibility & \texttt{dog} & Base & $-100.721\pm0.722$ & $-132.286\pm1.366$ \\
& & Vanilla SVDD & $-71.308\pm0.113$ & $-90.612\pm0.209$ \\
& & CVaR-SVDD & $\boldsymbol{-59.784\pm0.163}$ & $\boldsymbol{-71.202\pm0.382}$ \\
& & Plug-in CVaR-SVDD & $-64.753\pm0.044$ & $-73.852\pm0.215$ \\
\midrule
Compressibility & \texttt{rabbit} & Base & $-112.498\pm0.552$ & $-144.164\pm0.436$ \\
& & Vanilla SVDD & $-82.145\pm0.271$ & $-102.609\pm0.097$ \\
& & CVaR-SVDD & $\boldsymbol{-68.133\pm0.285}$ & $\boldsymbol{-82.879\pm0.271}$ \\
& & Plug-in CVaR-SVDD & $-72.417\pm0.511$ & $-83.951\pm0.430$ \\
\midrule
Compressibility & \texttt{cat} & Base
& $-95.986\pm0.214$ & $-123.367\pm1.025$ \\
& & Vanilla SMC
& $\mathbf{-60.004\pm0.737}$ & $-74.573\pm0.905$ \\
& & CVaR-SMC
& $-64.533\pm0.603$ & $\mathbf{-73.591\pm0.184}$ \\
& & Plug-in CVaR-SMC
& $-64.655\pm0.483$ & $-73.608\pm0.183$ \\
\midrule
Aesthetics & \texttt{cat} & Base & $5.8845\pm0.0076$ & $5.4600\pm0.0105$ \\
& & Vanilla SVDD & $5.9165\pm0.0036$ & $5.5154\pm0.0037$ \\
& & CVaR-SVDD & $5.9285\pm0.0041$ & $5.5341\pm0.0043$ \\
& & Plug-in CVaR-SVDD & $\boldsymbol{5.9293\pm0.0061}$ & $\boldsymbol{5.5345\pm0.0073}$ \\
\midrule
Aesthetics & \texttt{dog} & Base & $5.5611\pm0.0121$ & $5.1055\pm0.0063$ \\
& & Vanilla SVDD & $5.5709\pm0.0055$ & $5.0978\pm0.0106$ \\
& & CVaR-SVDD & $5.5847\pm0.0013$ & $\boldsymbol{5.1311\pm0.0078}$ \\
& & Plug-in CVaR-SVDD & $\boldsymbol{5.5913\pm0.0038}$ & $5.1289\pm0.0072$ \\
\midrule
Aesthetics & \texttt{rabbit} & Base & $5.7959\pm0.0031$ & $5.3661\pm0.0033$ \\
& & Vanilla SVDD & $5.8177\pm0.0023$ & $5.4241\pm0.0046$ \\
& & CVaR-SVDD & $\boldsymbol{5.8412\pm0.0080}$ & $\boldsymbol{5.4353\pm0.0097}$ \\
& & Plug-in CVaR-SVDD & $5.8263\pm0.0113$ & $5.4153\pm0.0061$ \\
\bottomrule
\end{tabular}
\end{table*}

\begin{figure*}[tp]
\centering
\begin{minipage}[t]{0.32\textwidth}
    \centering
    \includegraphics[width=\linewidth]
        {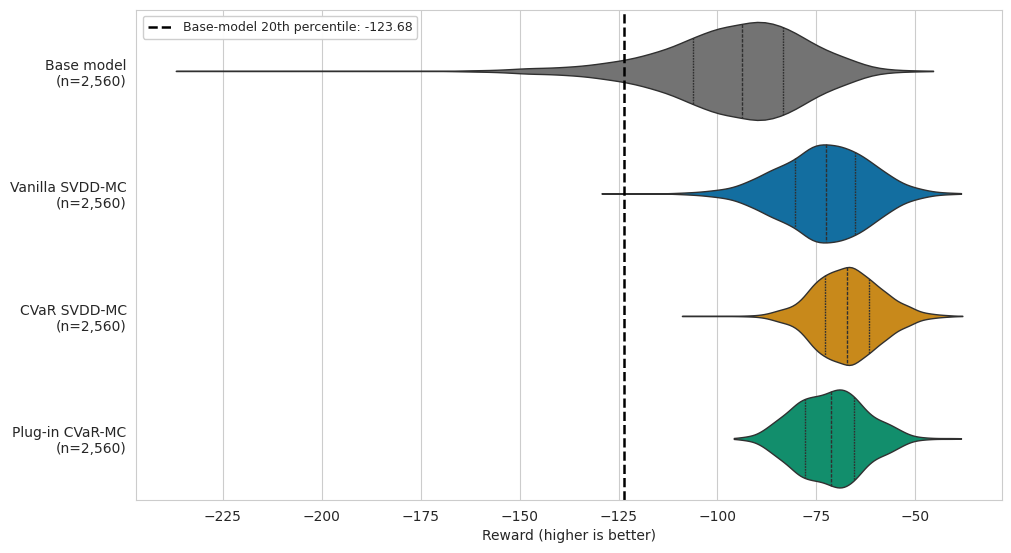}
    \par\small (a) Reward distributions.
\end{minipage}\hfill
\begin{minipage}[t]{0.32\textwidth}
    \centering
    \includegraphics[width=\linewidth]
        {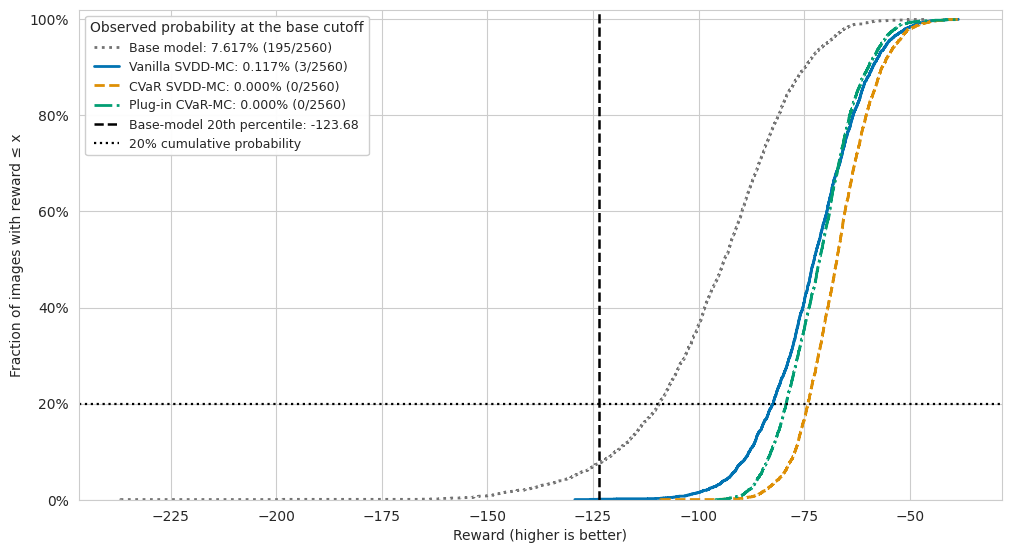}
    \par\small (b) Full reward CDF.
\end{minipage}\hfill
\begin{minipage}[t]{0.32\textwidth}
    \centering
    \includegraphics[width=\linewidth]
        {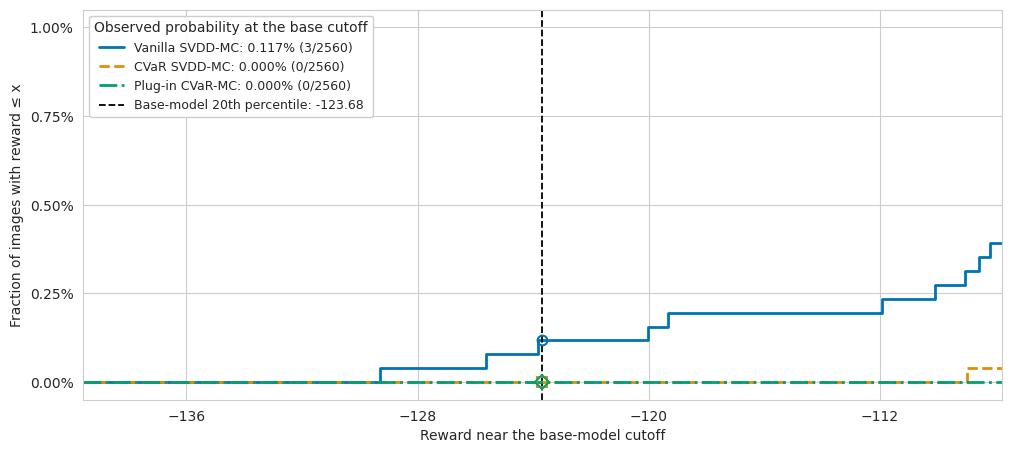}
    \par\small (c) CDF near the reference cutoff.
\end{minipage}
\caption{
Compressibility reward distributions for the \texttt{cat}
prompt.
Vertical lines in the CDF panels mark the $20$th-percentile reward of samples for threshold optimization, held fixed across methods.
Panel (c) contains a magnified view of (b) near the reference cutoff.
}
\label{fig:image-cat-compressibility}
\end{figure*}

\paragraph{Ablation on the plug-in CVaR sampler.}
Plug-in CVaR reuses the vanilla value predictor without additional training (see \eqref{eq:image-logweights}). Across the evaluated compressibility settings using SVDD, plug-in CVaR improves both mean and tail-average rewards over the vanilla sampler, though extra-learning the conditional positive-part cost yields further gains. \Cref{fig:image-cat-compressibility} further illustrates the shift toward higher compressibility rewards. For aesthetics, the differences between learned and plug-in CVaR are smaller and depend on the prompt and metric. Overall, while its tail improvements are less consistent, the plug-in CVaR offers a useful baseline, which demonstrates that the risk-adjusted resampling rule alone can improve tail performance in various settings.

\begin{figure*}[tp]
\centering
\begin{minipage}[t]{0.32\textwidth}
    \centering
    \includegraphics[width=\linewidth]
        {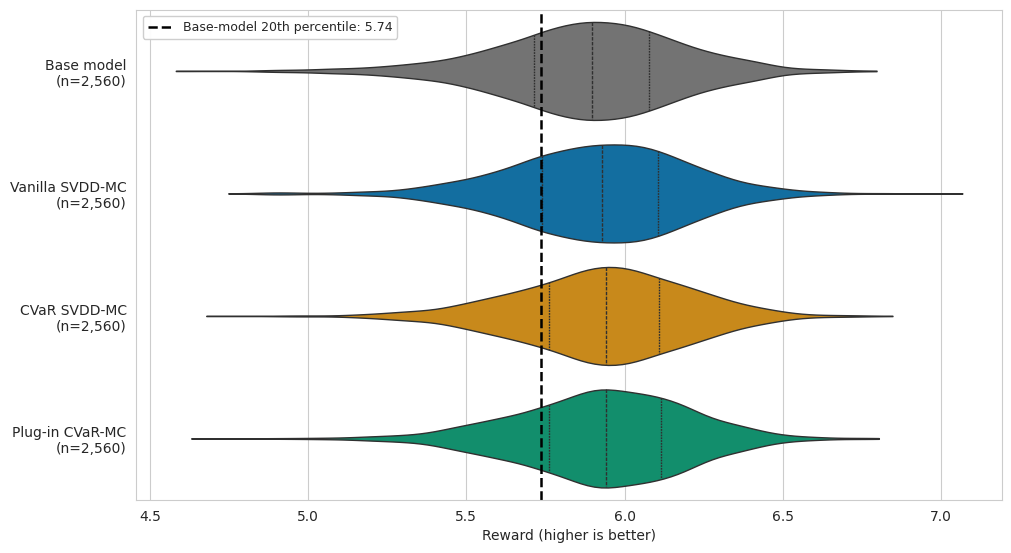}
    \par\small (a) Reward distributions.
\end{minipage}\hfill
\begin{minipage}[t]{0.32\textwidth}
    \centering
    \includegraphics[width=\linewidth]
        {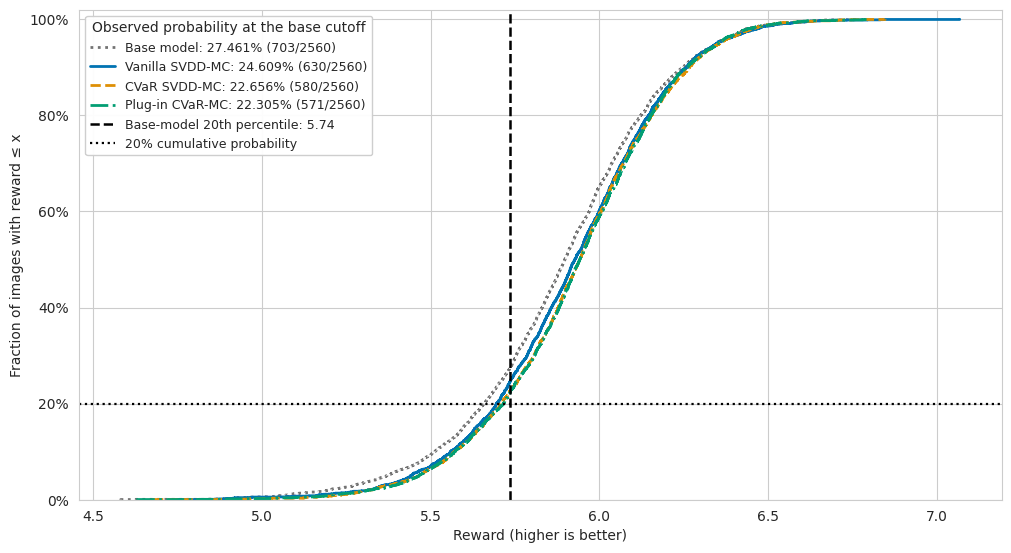}
    \par\small (b) Full reward CDF.
\end{minipage}\hfill
\begin{minipage}[t]{0.32\textwidth}
    \centering
    \includegraphics[width=\linewidth]
        {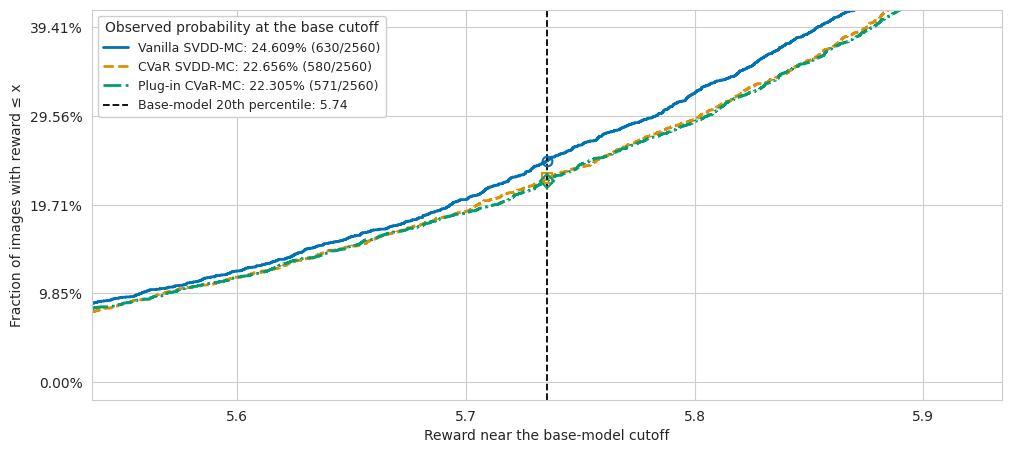}
    \par\small (c) CDF near the reference cutoff.
\end{minipage}
\caption{
Aesthetic reward distributions for the \texttt{cat} prompt.
Vertical lines in the CDF panels mark the $20$th-percentile reward of samples for threshold optimization, held fixed across methods.
Panel (c) contains a magnified view of (b) near the reference cutoff.
}
\label{fig:image-cat-aesthetic}
\end{figure*}

\paragraph{Ablation on the reward: aesthetics.}
CVaR-SVDD improves the observed mean and lower-tail aesthetic rewards relative to vanilla SVDD across the evaluated prompts. For \texttt{cat}, learned and plug-in CVaR yield nearly identical results, with differences small relative to the reported standard errors. The CDF and violin plot in \Cref{fig:image-cat-aesthetic} show that the aesthetic distributions remain close.

\begin{figure*}[tp]
\centering
\begin{minipage}[t]{0.49\textwidth}\centering
\includegraphics[width=\linewidth]{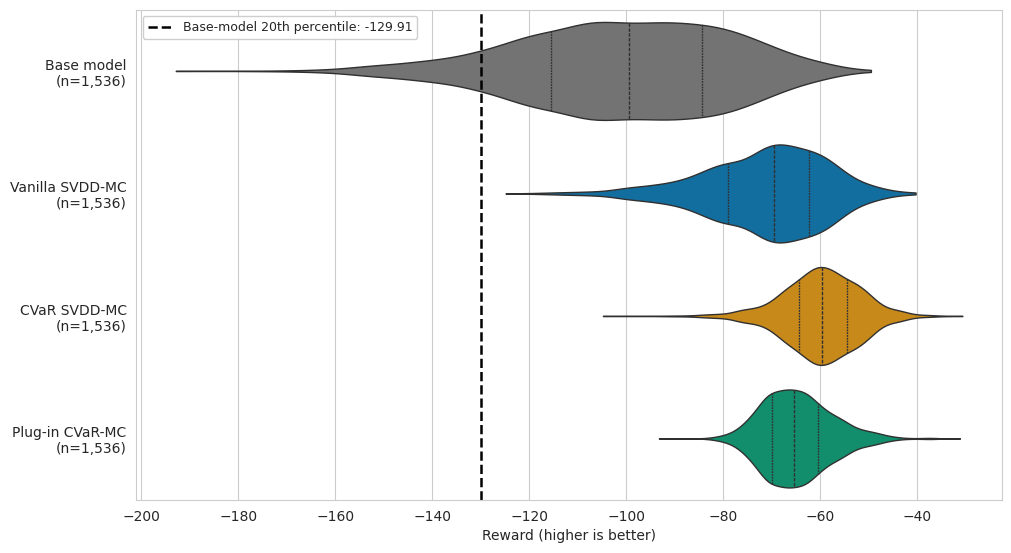}
\small (a) \texttt{dog}: reward distributions.
\end{minipage}\hfill
\begin{minipage}[t]{0.49\textwidth}\centering
\includegraphics[width=\linewidth]{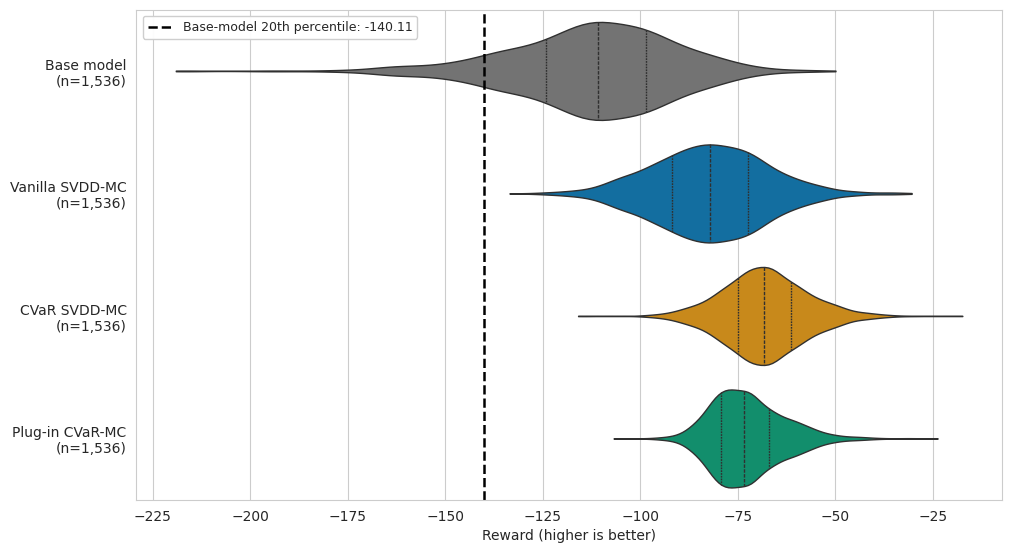}
\small (b) \texttt{rabbit}: reward distributions.
\end{minipage}
\par\medskip
\begin{minipage}[t]{0.49\textwidth}\centering
\includegraphics[width=\linewidth]{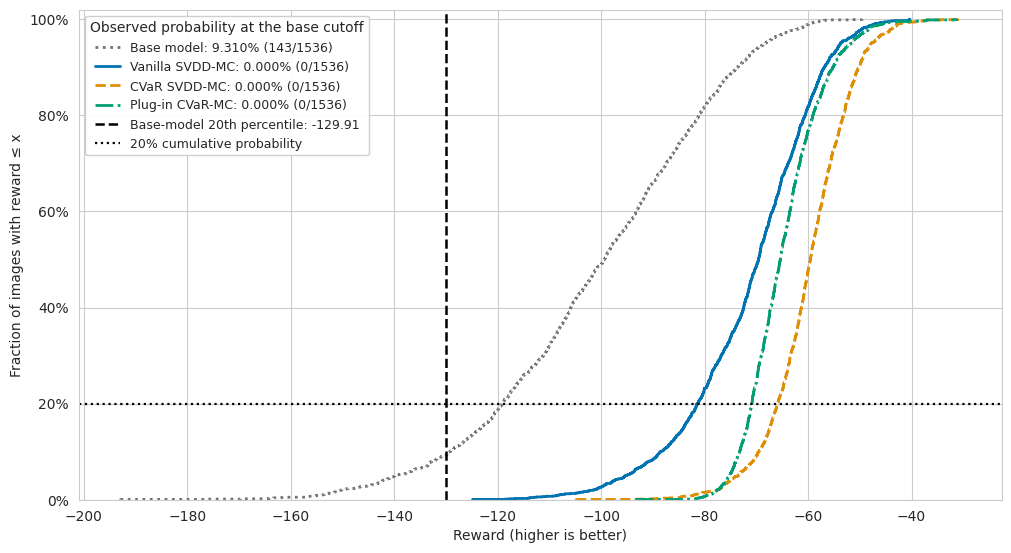}
\small (c) \texttt{dog}: full reward CDF.
\end{minipage}\hfill
\begin{minipage}[t]{0.49\textwidth}\centering
\includegraphics[width=\linewidth]{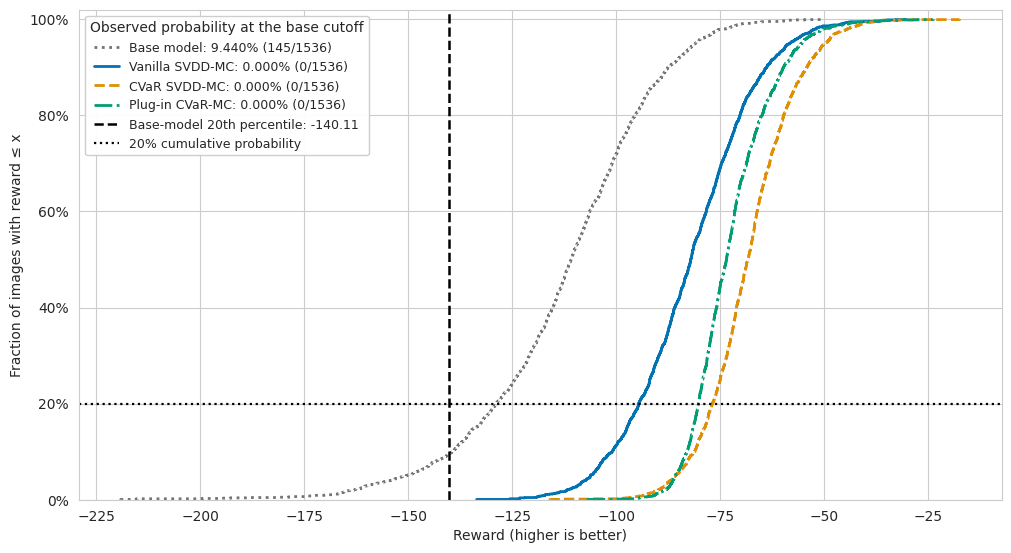}
\small (d) \texttt{rabbit}: full reward CDF.
\end{minipage}
\caption{Prompt ablation for compressibility. Vertical lines in the CDF panels mark the $20$th-percentile reward of samples for threshold optimization, held fixed across methods.}
\label{fig:image-prompt-ablation}
\end{figure*}

\begin{figure*}[tp]
\centering
\begin{minipage}[t]{0.49\textwidth}
\centering
\includegraphics[width=\linewidth]{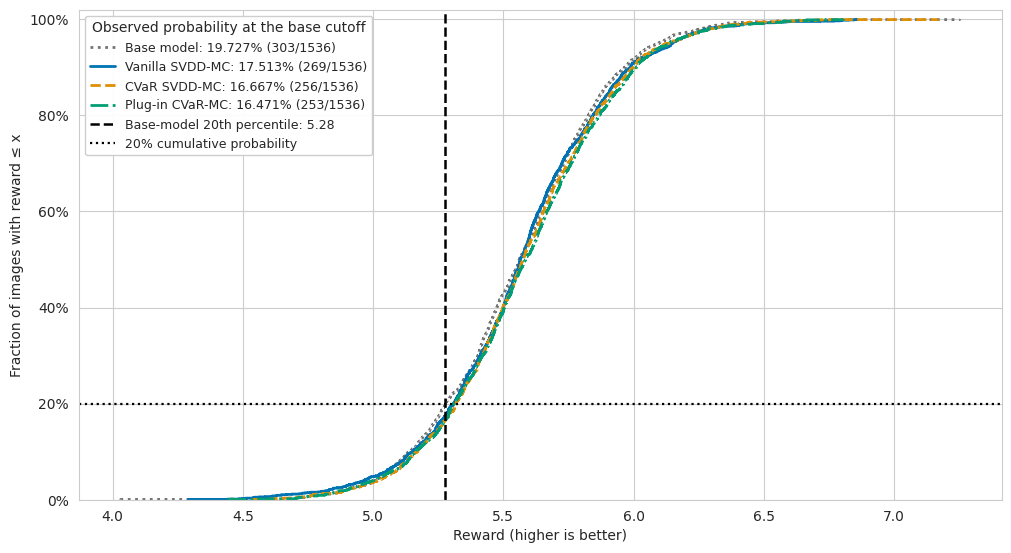}
\small (a) \texttt{dog}: full reward CDF.
\end{minipage}\hfill
\begin{minipage}[t]{0.49\textwidth}
\centering
\includegraphics[width=\linewidth]{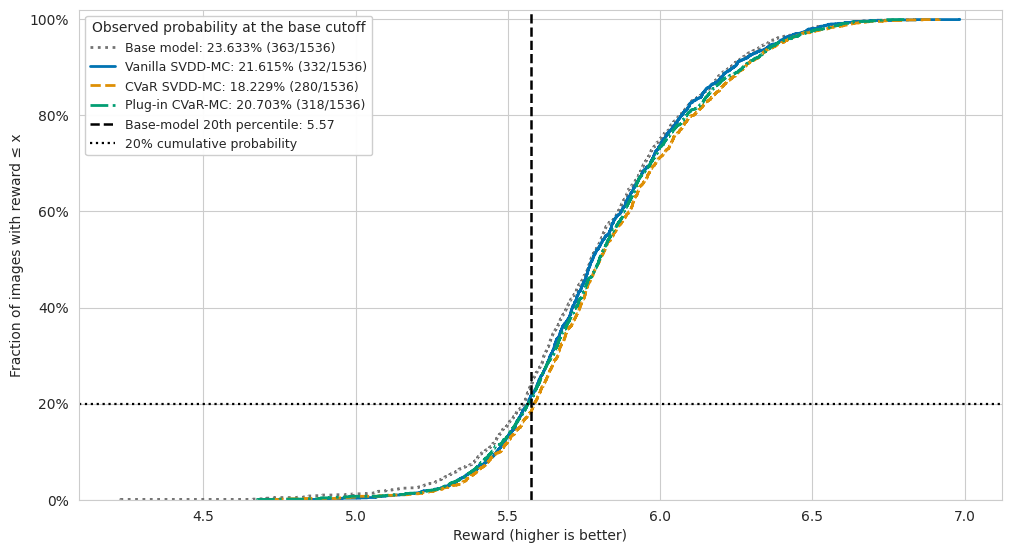}
\small (b) \texttt{rabbit}: full reward CDF.
\end{minipage}

\par\medskip

\begin{minipage}[t]{0.49\textwidth}
\centering
\includegraphics[width=\linewidth]{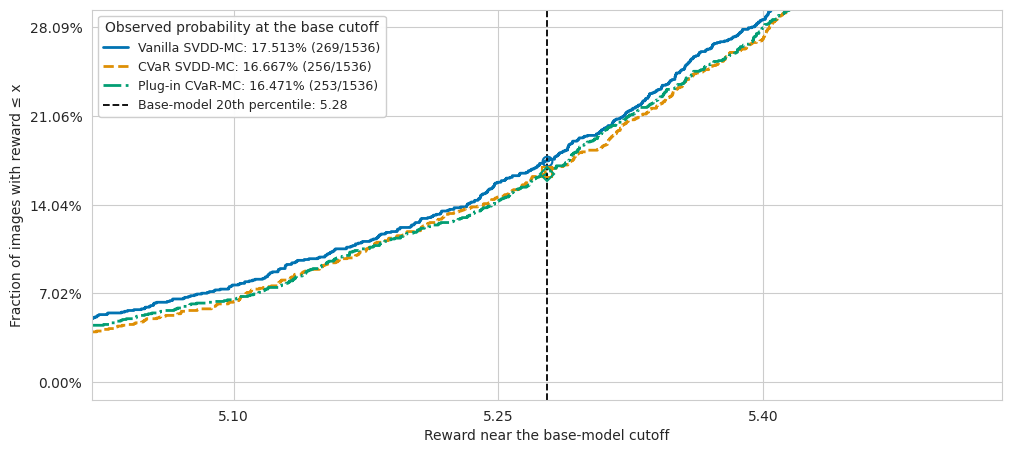}
\small (c) \texttt{dog}: reference-cutoff enlargement.
\end{minipage}\hfill
\begin{minipage}[t]{0.49\textwidth}
\centering
\includegraphics[width=\linewidth]{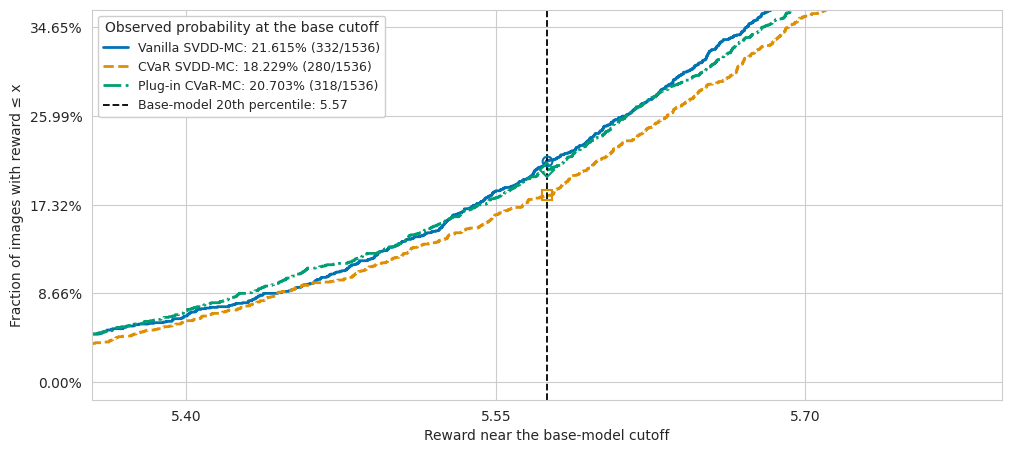}
\small (d) \texttt{rabbit}: reference-cutoff enlargement.
\end{minipage}

\caption{Prompt ablation for aesthetics. Vertical lines in the CDF panels mark the $20$th-percentile reward of samples for threshold optimization, held fixed across methods. Full CDFs include the base sampler; enlarged panels show the
three guided methods.}
\label{fig:image-aesthetic-prompts}
\end{figure*}

\paragraph{Ablation on the prompt.}
The improvements over vanilla SVDD persist for \texttt{dog} and \texttt{rabbit} under both rewards. Learned CVaR also outperforms plug-in CVaR on both compressibility metrics, whereas their aesthetic rankings depend on the prompt and metric. For aesthetic \texttt{rabbit}, plug-in CVaR reduces the cutoff exceedance fraction but worsens the tail-average reward, illustrating that these metrics capture different aspects of the lower tail. We also show the corresponding reward distributions in \Cref{fig:image-prompt-ablation,fig:image-aesthetic-prompts}.
For compressibility, both learned and plug-in CVaR shift the reward distribution toward larger values relative to vanilla SVDD, with learned CVaR providing greater improvements. For aesthetics, the two methods perform similarly for \texttt{cat} and \texttt{dog}, while learned CVaR achieves larger gains for \texttt{rabbit}, where plug-in CVaR fails to improve the tail-average reward over vanilla SVDD.

\begin{figure*}[tp]
\centering
\begin{minipage}[t]{0.49\textwidth}\centering
\includegraphics[width=\linewidth]{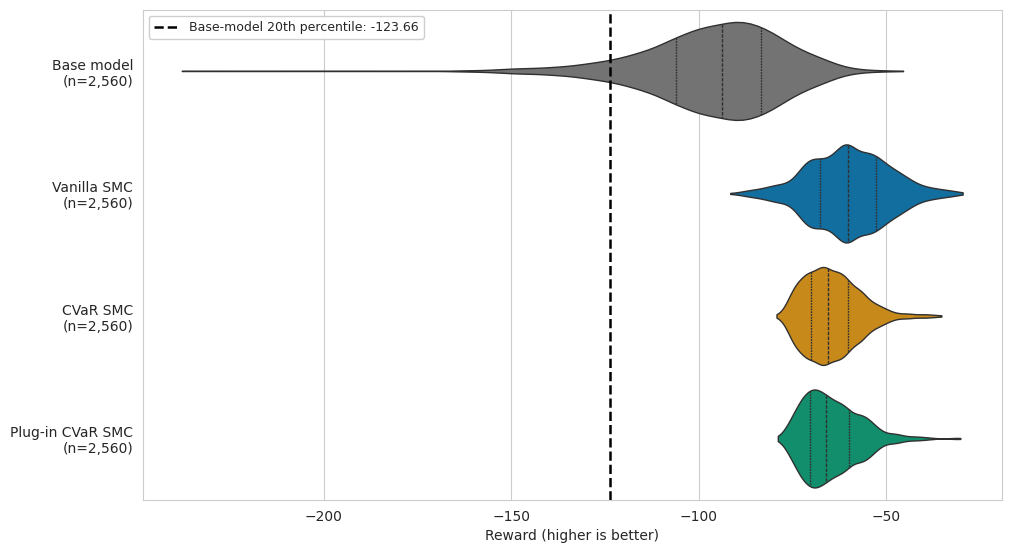}
\small (a) Reward distributions.
\end{minipage}\hfill
\begin{minipage}[t]{0.49\textwidth}\centering
\includegraphics[width=\linewidth]{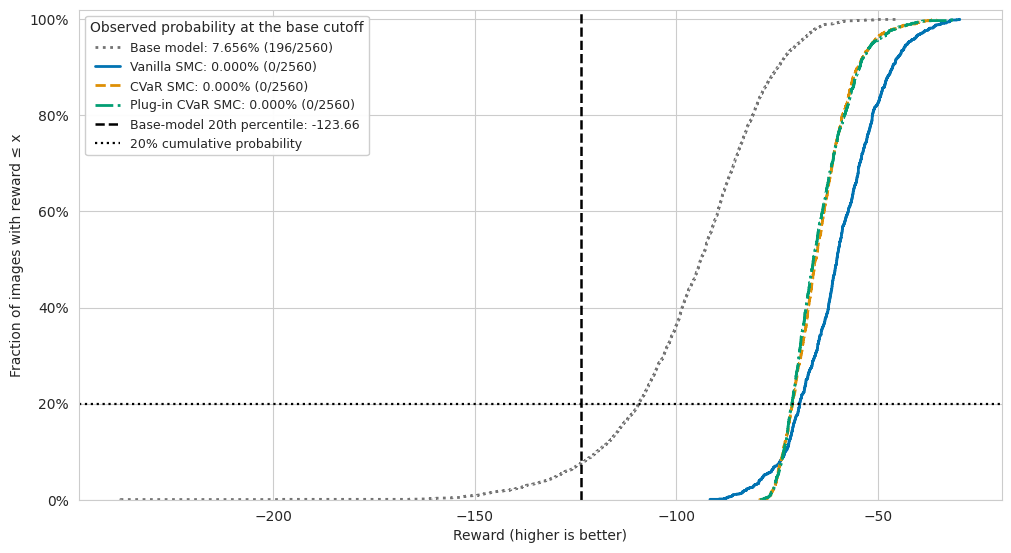}
\small (b) Full reward CDF.
\end{minipage}
\caption{SMC ablation for compressibility with \texttt{cat}. Vertical lines in the CDF panels mark the $20$th-percentile reward of samples for threshold optimization, held fixed across methods.}
\label{fig:image-smc-ablation}
\end{figure*}

\paragraph{Ablation on the sampler: SMC.}
For compressibility with \texttt{cat}, 
CVaR-SMC exhibits an empirical mean-tail trade-off relative to vanilla SMC: its observed lower-tail mean reward increases, while its overall mean reward decreases. Here learned and plug-in CVaR-SMC yield similar results. Such a trade-off is compatible with the pure CVaR objective ($\lambda=1$), which does not reward further cost reductions below the threshold $\eta$. See \Cref{fig:image-smc-ablation} for the reward distributions.

During the ablation study, we also considered an SMC variant that omits the terminal-cost correction before final output, as well as the initial importance weighting. With this variant, vanilla SMC achieves mean and lower-tail rewards of $-75.477$ and $-89.215$, respectively, while the corresponding rewards for CVaR-SMC are $-67.399$ and $-76.288$. Both methods attain lower rewards than the corrected implementation reported in \Cref{tab:image-full}, with a larger difference for vanilla SMC. This comparison highlights the practical importance of the SMC weight corrections.
Also, intuitively, such a larger gain for vanilla SMC is consistent with the different terminal objectives: the vanilla tilting factor $\exp(-c(x_0)/\alpha)$ favors \textit{every} reduction in cost, whereas the pure-CVaR tilting factor $\exp(-(c(x_0)-\eta)^+/(\alpha(1-\beta)))$ becomes \textit{constant} for all $c(x_0) \leq \eta$. In other words, the vanilla objective continues to favor further cost reductions below the threshold, while the pure-CVaR objective does not distinguish among these outcomes.

\paragraph{Remark on the wall-clock sampling time.}
Within each evaluated scenario, vanilla SVDD-MC, learned CVaR-SVDD, and plug-in CVaR-SVDD have similar wall-clock sampling times. For example, for aesthetics with \texttt{rabbit}, generating $512$ images in batches of $32$ takes approximately $767$, $770$, and $769$ seconds, respectively. These times cover guided image generation and exclude value-network training and the separate threshold-optimization stage. Notably, the CVaR transformation introduces little additional sampling overhead relative to vanilla SVDD-MC in the evaluated settings.

\subsubsection{Qualitative tail examples}
\label{app:image-tail-examples}

\begin{figure*}[tp]
\centering
\includegraphics[width=0.93\textwidth]{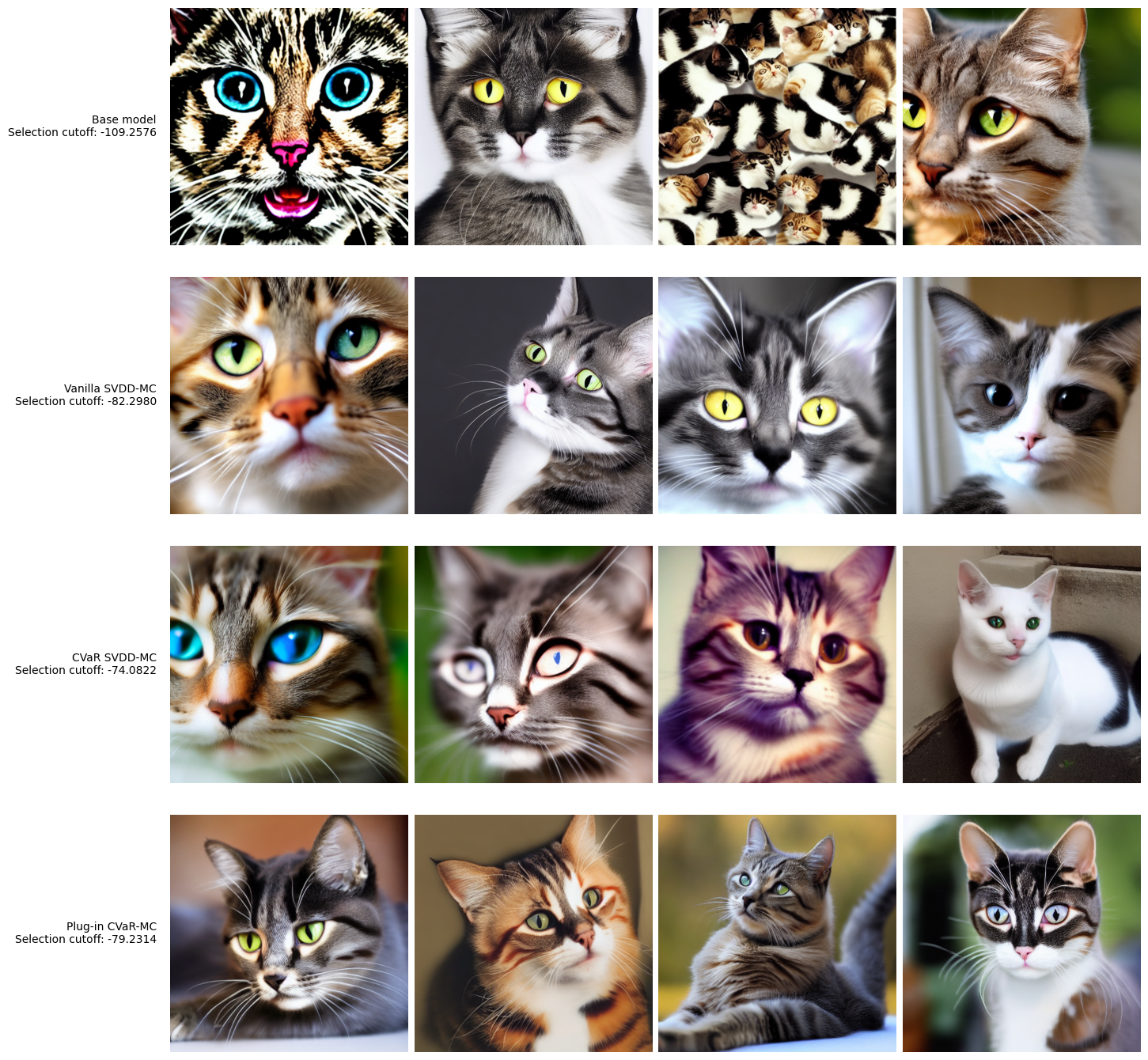}
\caption{Random examples from each method's own worst-$20\%$ compressibility-reward tail for \texttt{cat}. Rows are base, vanilla SVDD, learned CVaR-SVDD, and plug-in CVaR-SVDD. Each row uses its own pooled reward quantile, shown at left. Columns are not paired comparisons.}
\label{fig:image-tail-comp-cat}
\end{figure*}

\begin{figure*}[tp]
\centering
\includegraphics[width=0.93\textwidth]{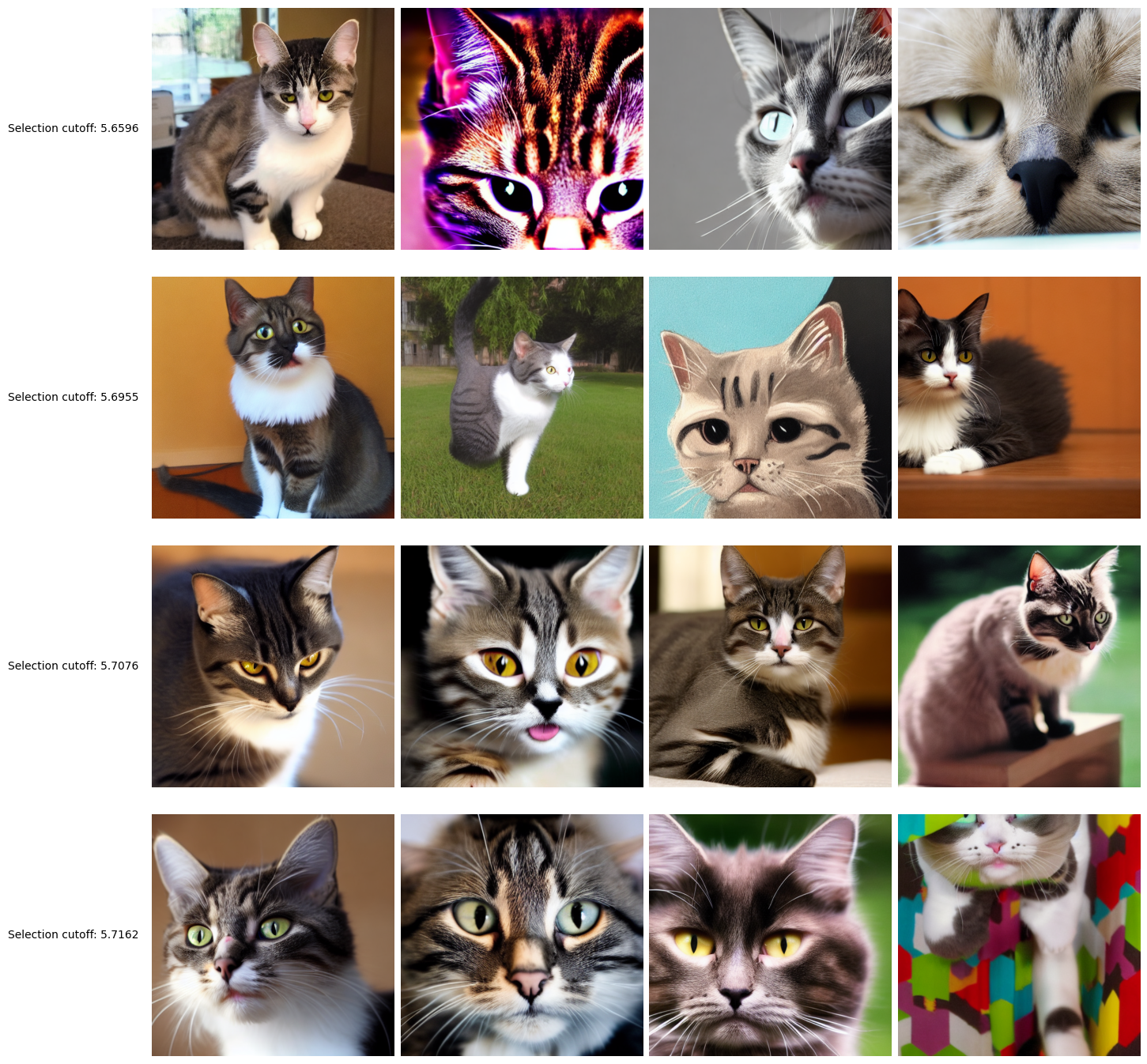}
\caption{Random examples from each method's own worst-$20\%$ aesthetic-reward tail for \texttt{cat}. Rows are base, vanilla SVDD, learned CVaR-SVDD, and plug-in CVaR-SVDD. Each row uses its own pooled reward quantile, shown at left. Columns are not paired comparisons.}
\label{fig:image-tail-aesthetic}
\end{figure*}

\begin{figure*}[tp]
\centering
\includegraphics[width=0.93\textwidth]{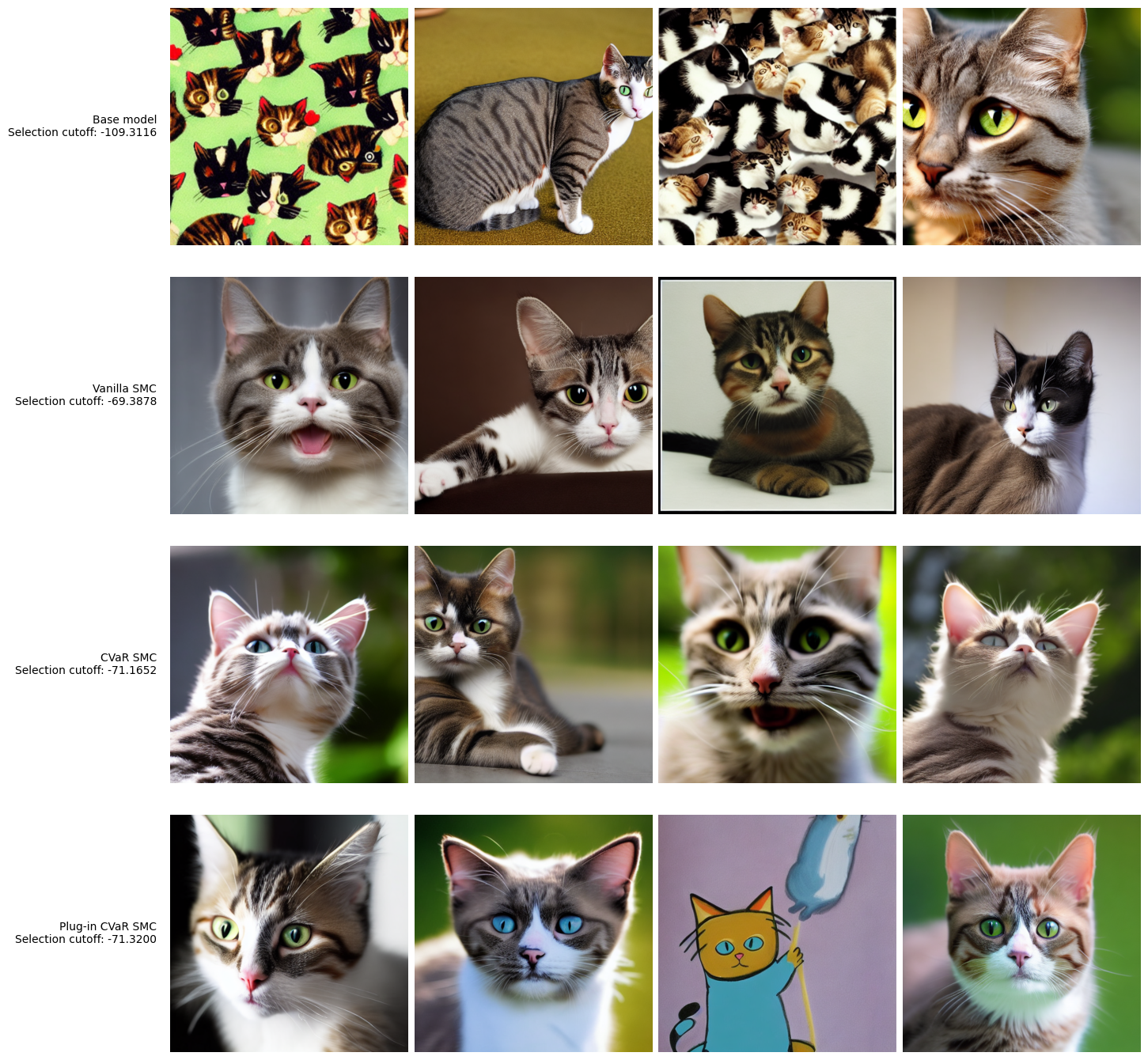}
\caption{Compressibility tail examples for the SMC ablation with \texttt{cat}. Rows are base, vanilla SMC, learned CVaR-SMC, and plug-in CVaR-SMC. Each row uses its own pooled reward quantile, shown at left. Columns are not paired comparisons.}
\label{fig:image-tail-smc}
\end{figure*}

\begin{figure*}[tp]
\centering
\begin{minipage}[t]{0.49\textwidth}
\centering
\includegraphics[width=\linewidth]{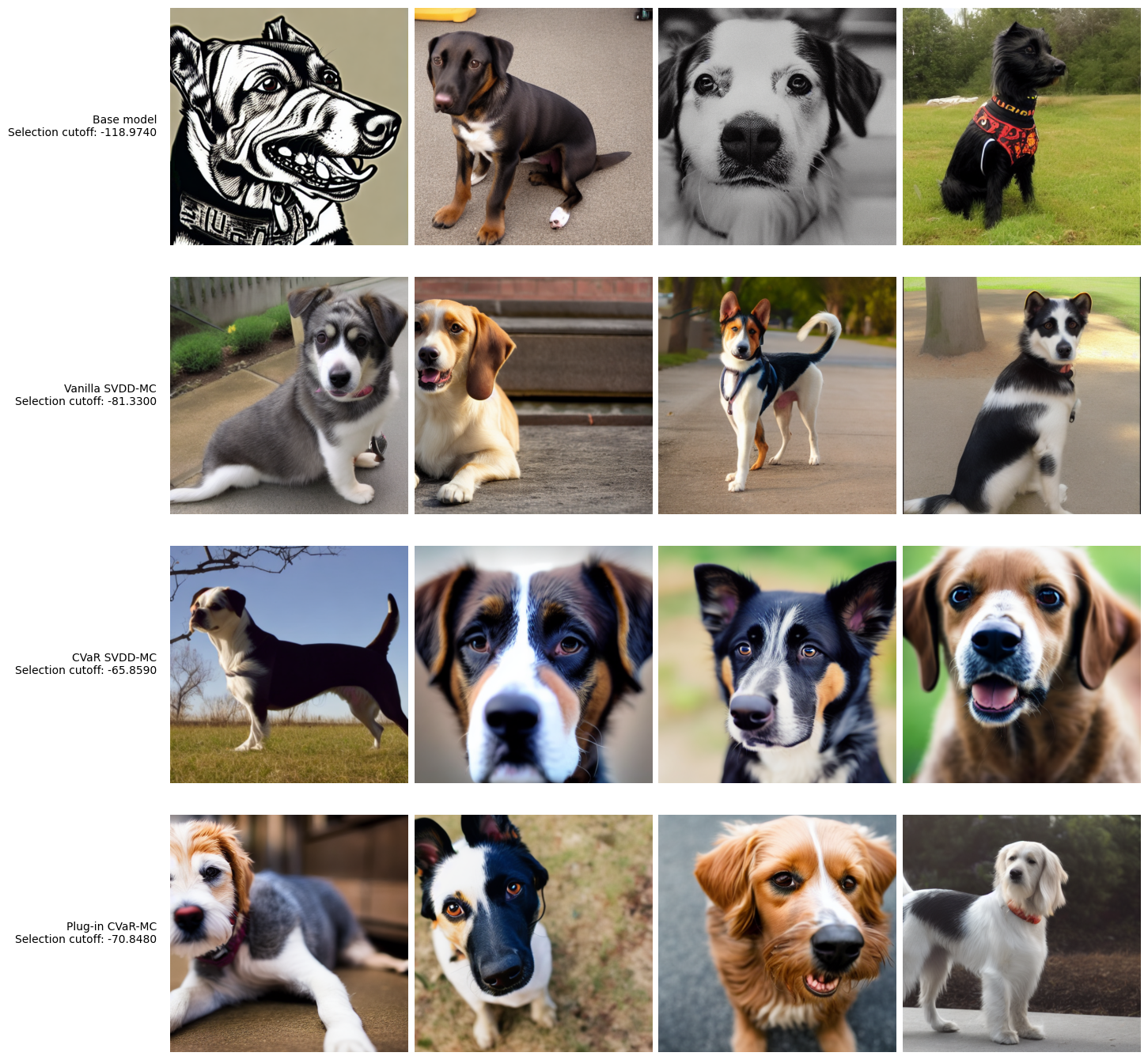}
\par\small (a) Compressibility: \texttt{dog}.
\end{minipage}\hfill
\begin{minipage}[t]{0.49\textwidth}
\centering
\includegraphics[width=\linewidth]{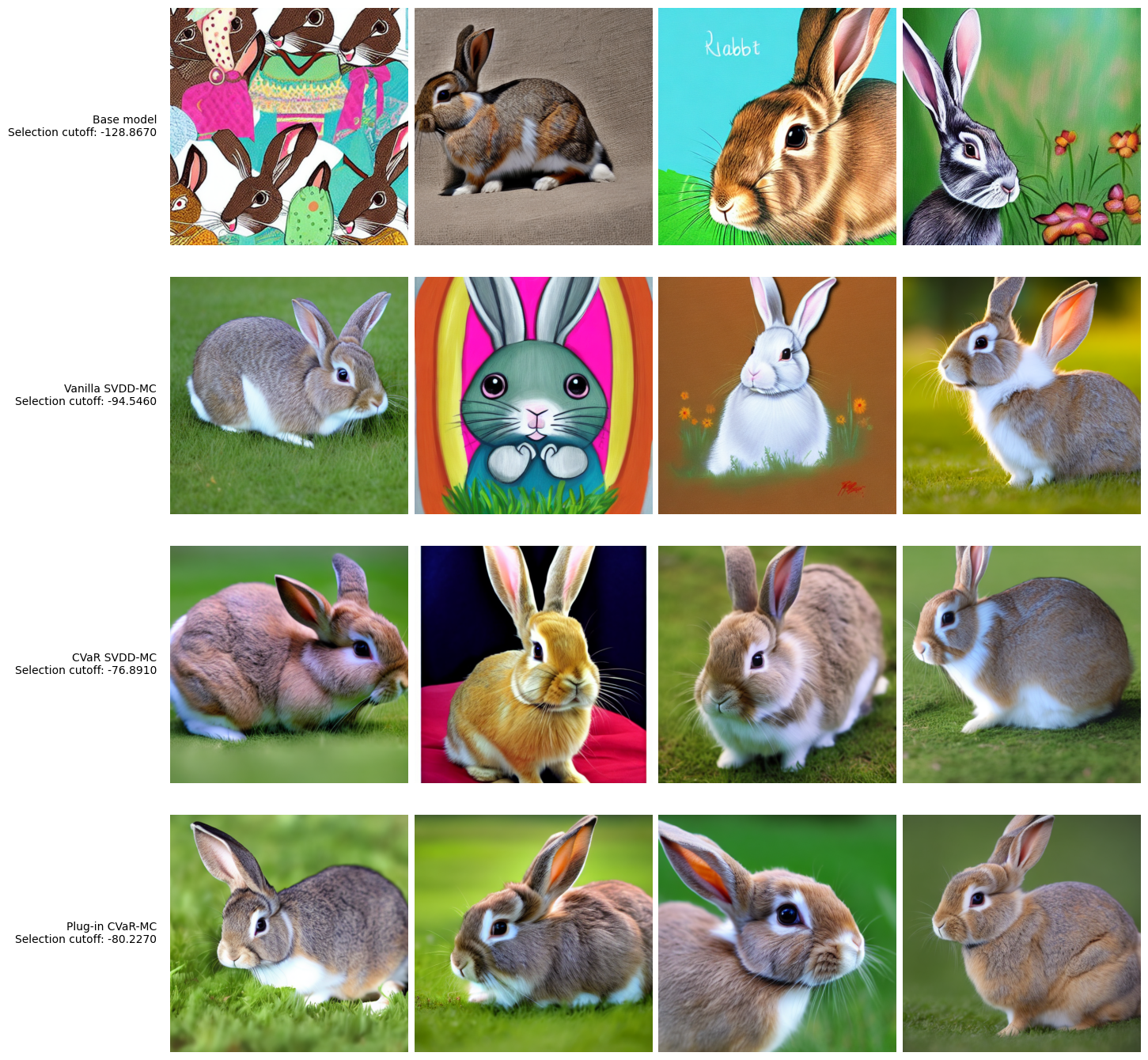}
\par\small (b) Compressibility: \texttt{rabbit}.
\end{minipage}

\par\medskip
\begin{minipage}[t]{0.49\textwidth}
\centering
\includegraphics[width=\linewidth]{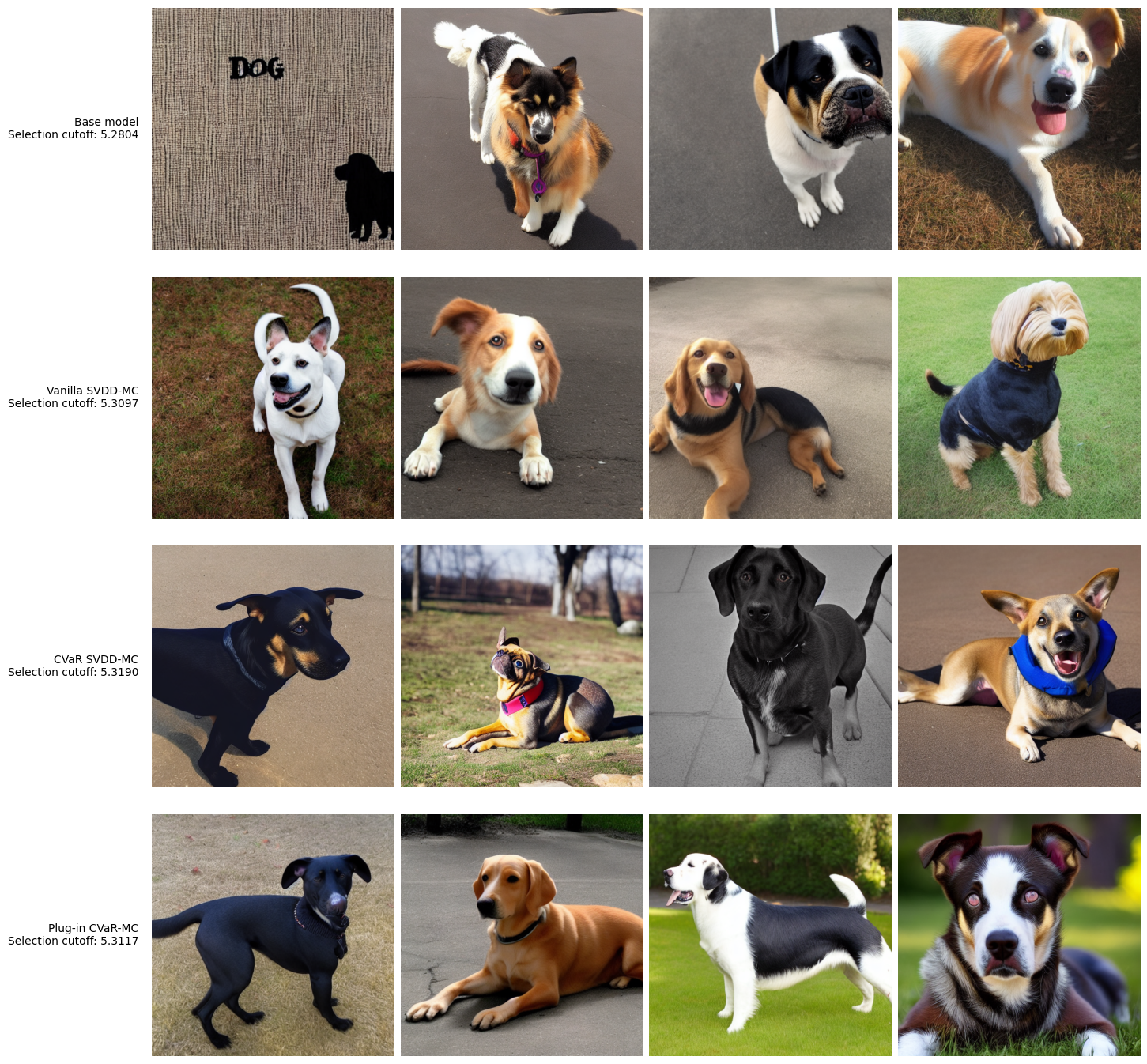}
\par\small (c) Aesthetics: \texttt{dog}.
\end{minipage}\hfill
\begin{minipage}[t]{0.49\textwidth}
\centering
\includegraphics[width=\linewidth]{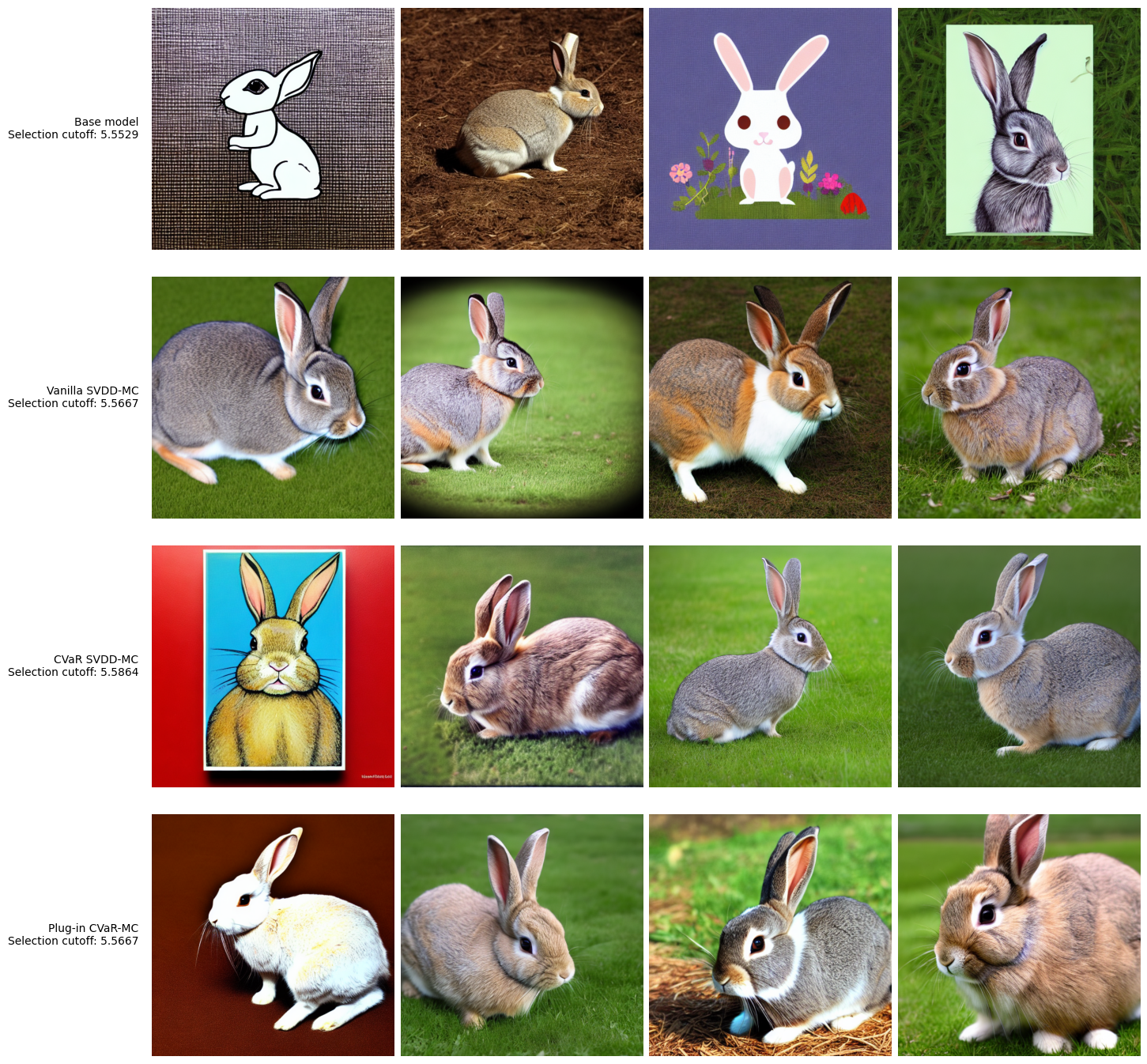}
\par\small (d) Aesthetics: \texttt{rabbit}.
\end{minipage}

\caption{Tail examples for \texttt{dog} and \texttt{rabbit} under compressibility and aesthetic rewards. In each panel, rows correspond to base, vanilla SVDD, learned CVaR-SVDD, and plug-in CVaR-SVDD, with four random images from each method's own pooled lower tail. Columns are not paired comparisons.}
\label{fig:image-tail-prompts}
\end{figure*}

For each method, we pool its evaluation rewards across the displayed seeds, compute its own $20$th-percentile reward, and select four images uniformly without replacement from observations at or below that threshold, including ties. 
\Cref{fig:image-tail-comp-cat,fig:image-tail-aesthetic,fig:image-tail-smc,fig:image-tail-prompts} display the saved examples. For compressibility, reduced image detail and smoother regions can contribute to smaller JPEG sizes. Such changes explain the reward but should not be equated with an improvement in perceptual quality or image diversity. The aesthetic examples illustrate the contents of the low-scoring tail, while the quantitative oracle measurements provide the comparison. 

%% file: Xian_bib.bib
@article{rockafellar2002conditional,
	author = {Rockafellar, R Tyrrell and Uryasev, Stanislav},
	journal = {Journal of Banking \& Finance},
	number = {7},
	pages = {1443--1471},
	publisher = {Elsevier},
	title = {Conditional value-at-risk for general loss distributions},
	volume = {26},
	year = {2002}}

@article{rockafellar2000optimization,
	author = {Rockafellar, R Tyrrell and Uryasev, Stanislav and others},
	journal = {Journal of Risk},
	number = {3},
	pages = {21--42},
	title = {Optimization of conditional value-at-risk},
	volume = {2},
	year = {2000}}


%% file: diffusion.bib
@inproceedings{wang2026cvar,
	author = {Zifan Wang and Riccardo De Santi and Xiaoyu Mo and Michael M. Zavlanos and Andreas Krause and Karl Henrik Johansson},
	booktitle = {Forty-third International Conference on Machine Learning},
	title = {Efficient Tail-Aware Generative Optimization via Flow Model Fine-Tuning},
	year = {2026}}

@misc{schuhmann2022aesthetic,
	author = {Schuhmann, Christoph},
	howpublished = {\url{https://github.com/christophschuhmann/improved-aesthetic-predictor}},
	note = {GitHub repository},
	title = {{LAION-Aesthetics Predictor V2}},
	year = {2022}}

@article{fan2023dpok,
	author = {Ying Fan and Olivia Watkins and Yuqing Du and Hao Liu and Moonkyung Ryu and Craig Boutilier and Pieter Abbeel and Mohammad Ghavamzadeh and Kangwook Lee and Kimin Lee},
	journal = {arXiv preprint arXiv:2305.16381},
	title = {DPOK: Reinforcement Learning for Fine-tuning Text-to-Image Diffusion Models},
	year = {2023}}

@inproceedings{zhang2025daps,
	author = {Zhang, Bingliang and Chu, Wenda and Berner, Julius and Meng, Chenlin and Anandkumar, Anima and Song, Yang},
	booktitle = {2025 IEEE/CVF Conference on Computer Vision and Pattern Recognition (CVPR)},
	title = {Improving Diffusion Inverse Problem Solving with Decoupled Noise Annealing},
	year = {2025}}

@inproceedings{wu2023tds,
	author = {Luhuan Wu and Brian L. Trippe and Christian A Naesseth and John Patrick Cunningham and David Blei},
	booktitle = {Thirty-seventh Conference on Neural Information Processing Systems},
	title = {Practical and Asymptotically Exact Conditional Sampling in Diffusion Models},
	year = {2023}}

@inproceedings{narasimhan2025constrained,
	author = {Sai Shankar Narasimhan and Shubhankar Agarwal and Litu Rout and Sanjay Shakkottai and Sandeep P. Chinchali},
	booktitle = {The Thirty-ninth Annual Conference on Neural Information Processing Systems},
	title = {Constrained Posterior Sampling: Time Series Generation with Hard Constraints},
	year = {2025}}

@inproceedings{christopher2024constrained,
	author = {Jacob K Christopher and Stephen Baek and Ferdinando Fioretto},
	booktitle = {The Thirty-eighth Annual Conference on Neural Information Processing Systems},
	title = {Constrained Synthesis with Projected Diffusion Models},
	year = {2024}}

@article{clark2024draft,
	author = {Kevin Clark and Paul Vicol and Kevin Swersky and David J Fleet},
	journal = {arXiv preprint arXiv:2309.17400},
	title = {Directly Fine-Tuning Diffusion Models on Differentiable Rewards},
	year = {2024}}

@article{black2024ddpo,
	author = {Kevin Black and Michael Janner and Yilun Du and Ilya Kostrikov and Sergey Levine},
	journal = {arXiv preprint arXiv:2305.13301},
	title = {Training Diffusion Models with Reinforcement Learning},
	year = {2024}}

@inproceedings{zhu2026doit,
	author = {Qijie Zhu and Zeqi Ye and Han Liu and Zhaoran Wang and Minshuo Chen},
	booktitle = {Forty-third International Conference on Machine Learning},
	title = {Training-Free Adaptation of Diffusion Models via Doob's \$h\$-Transform},
	year = {2026}}

@article{gong2026rsdiffuser,
	author = {Shiqiang Gong},
	journal = {arXiv preprint arXiv:2606.27766},
	title = {RS-Diffuser: Risk-Sensitive Diffusion Planning with Distributional Value Guidance},
	year = {2026}}

@inproceedings{desanti2025fdc,
	author = {Riccardo De Santi and Marin Vlastelica and Ya-Ping Hsieh and Zebang Shen and Niao He and Andreas Krause},
	booktitle = {The Thirty-ninth Annual Conference on Neural Information Processing Systems},
	title = {Flow Density Control: Generative Optimization Beyond Entropy-Regularized Fine-Tuning},
	year = {2025}}

@inproceedings{ma2025robotics,
	author = {Hao Ma and Sabrina Bodmer and Andrea Carron and Melanie Zeilinger and Michael Muehlebach},
	booktitle = {9th Annual Conference on Robot Learning},
	title = {Constraint-Aware Diffusion Guidance for Robotics: Real-Time Obstacle Avoidance for Autonomous Racing},
	year = {2025}}

@inproceedings{xiao2025safediffuser,
	author = {Wei Xiao and Tsun-Hsuan Wang and Chuang Gan and Ramin Hasani and Mathias Lechner and Daniela Rus},
	booktitle = {The Thirteenth International Conference on Learning Representations},
	title = {SafeDiffuser: Safe Planning with Diffusion Probabilistic Models},
	year = {2025}}

@article{melnik2024video-diffusion-survey,
	author = {Andrew Melnik and Michal Ljubljanac and Cong Lu and Qi Yan and Weiming Ren and Helge Ritter},
	journal = {arXiv preprint arXiv:2405.03150},
	title = {Video Diffusion Models: A Survey},
	year = {2024}}

@article{delmoral2006sequential,
	author = {{Del Moral}, Pierre and Doucet, Arnaud and Jasra, Ajay},
	journal = {Journal of the Royal Statistical Society: Series B (Statistical Methodology)},
	title = {Sequential {Monte Carlo} samplers},
	year = {2006}}

@inproceedings{madhow2026lcb,
	author = {Sunil Madhow and Yuchen Liang and Ness Shroff and Yingbin Liang and Yu-Xiang Wang},
	booktitle = {The Fortieth Annual Conference on Neural Information Processing Systems},
	title = {Learnable Chernoff Baselines for Provable Inference-Time Alignment},
	year = {2026}}

@inproceedings{uehara2024finetuning,
	author = {Masatoshi Uehara and Yulai Zhao and Ehsan Hajiramezanali and Gabriele Scalia and G{\"o}kcen Eraslan and Avantika Lal and Sergey Levine and Tommaso Biancalani},
	booktitle = {The Thirty-eighth Annual Conference on Neural Information Processing Systems},
	title = {Bridging Model-Based Optimization and Generative Modeling via Conservative Fine-Tuning of Diffusion Models},
	year = {2024}}

@inproceedings{singhal2025fksteering,
	author = {Raghav Singhal and Zachary Horvitz and Ryan Teehan and Mengye Ren and Zhou Yu and Kathleen McKeown and Rajesh Ranganath},
	booktitle = {Forty-second International Conference on Machine Learning},
	title = {A General Framework for Inference-time Scaling and Steering of Diffusion Models},
	year = {2025}}

@article{zhu2026smc,
	author = {Youheng Zhu and Yiping Lu},
	journal = {arXiv preprint arXiv:2602.01381},
	title = {On the Power of (Approximate) Reward Models for Inference-Time Scaling},
	year = {2026}}

@inproceedings{li2025svdd,
	author = {Xiner Li and Yulai Zhao and Chenyu Wang and Gabriele Scalia and G{\"o}kcen Eraslan and Surag Nair and Tommaso Biancalani and Shuiwang Ji and Aviv Regev and Sergey Levine and Masatoshi Uehara},
	booktitle = {The Thirty-ninth Annual Conference on Neural Information Processing Systems},
	title = {Derivative-Free Guidance in Continuous and Discrete Diffusion Models with Soft Value-based Decoding},
	year = {2025}}

@article{uehara2025survey-alignment,
	author = {Masatoshi Uehara and Yulai Zhao and Chenyu Wang and Xiner Li and Aviv Regev and Sergey Levine and Tommaso Biancalani},
	journal = {arXiv preprint arXiv:2501.09685},
	title = {Inference-Time Alignment in Diffusion Models with Reward-Guided Generation: Tutorial and Review},
	year = {2025}}

@article{uehara2024survey-finetuning,
	author = {Masatoshi Uehara and Yulai Zhao and Tommaso Biancalani and Sergey Levine},
	journal = {arXiv preprint arXiv:2407.13734},
	title = {Understanding Reinforcement Learning-Based Fine-Tuning of Diffusion Models: A Tutorial and Review},
	year = {2024}}

@inproceedings{wang2025finetuning,
	author = {Chenyu Wang and Masatoshi Uehara and Yichun He and Amy Wang and Avantika Lal and Tommi Jaakkola and Sergey Levine and Aviv Regev and Hanchen and Tommaso Biancalani},
	booktitle = {The Thirteenth International Conference on Learning Representations},
	title = {Fine-Tuning Discrete Diffusion Models via Reward Optimization with Applications to {DNA} and Protein Design},
	year = {2025}}

@inproceedings{liang2025conditional,
	author = {Yuchen Liang and Peizhong Ju and Yingbin Liang and Ness Shroff},
	booktitle = {The Thirteenth International Conference on Learning Representations},
	title = {Theory on Score-Mismatched Diffusion Models and Zero-Shot Conditional Samplers},
	year = {2025}}

@inproceedings{song2021ddim,
	author = {Jiaming Song and Chenlin Meng and Stefano Ermon},
	booktitle = {International Conference on Learning Representations},
	title = {Denoising Diffusion Implicit Models},
	year = {2021}}

@inproceedings{chung2023dps,
	author = {Hyungjin Chung and Jeongsol Kim and Michael Thompson Mccann and Marc Louis Klasky and Jong Chul Ye},
	booktitle = {The Eleventh International Conference on Learning Representations},
	title = {Diffusion Posterior Sampling for General Noisy Inverse Problems},
	year = {2023}}

@inproceedings{song2023pgdm,
	author = {Jiaming Song and Arash Vahdat and Morteza Mardani and Jan Kautz},
	booktitle = {International Conference on Learning Representations},
	title = {Pseudoinverse-Guided Diffusion Models for Inverse Problems},
	year = {2023}}

@article{dalle2,
	author = {Aditya Ramesh and Prafulla Dhariwal and Alex Nichol and Casey Chu and Mark Chen},
	journal = {arXiv preprint arXiv:2204.06125},
	title = {Hierarchical Text-Conditional Image Generation with CLIP Latents},
	year = {2022}}

@inproceedings{rombach2022stable-diffusion,
	author = {Rombach, Robin and Blattmann, Andreas and Lorenz, Dominik and Esser, Patrick and Ommer, Bj\"orn},
	booktitle = {Proceedings of the IEEE/CVF Conference on Computer Vision and Pattern Recognition (CVPR)},
	month = {June},
	pages = {10684-10695},
	title = {High-Resolution Image Synthesis With Latent Diffusion Models},
	year = {2022}}

@inproceedings{song2020sde,
	author = {Yang Song and Jascha Sohl{-}Dickstein and Diederik P. Kingma and Abhishek Kumar and Stefano Ermon and Ben Poole},
	booktitle = {The Ninth International Conference on Learning Representations},
	title = {Score-Based Generative Modeling through Stochastic Differential Equations},
	year = {2021}}

@article{tweedie2011efron,
	author = {Bradley Efron},
	journal = {Journal of the American Statistical Association},
	number = {496},
	pages = {1602-1614},
	title = {Tweedie's Formula and Selection Bias},
	volume = {106},
	year = {2011}}

@inproceedings{ho2020ddpm,
	author = {Ho, Jonathan and Jain, Ajay and Abbeel, Pieter},
	booktitle = {Advances in Neural Information Processing Systems},
	pages = {6840-6851},
	title = {Denoising Diffusion Probabilistic Models},
	volume = {33},
	year = {2020}}


%% file: refs.bib
@article{prekopa1973logarithmic,
  title={On logarithmic concave measures and functions},
  author={Pr{\'e}kopa, Andr{\'a}s},
  journal={Acta Sci. Math.},
  volume={34},
  pages={335},
  year={1973}
}

@inproceedings{cohen2016geometric,
  title={Geometric median in nearly linear time},
  author={Cohen, Michael B and Lee, Yin Tat and Miller, Gary and Pachocki, Jakub and Sidford, Aaron},
  booktitle={Proceedings of the forty-eighth annual ACM symposium on Theory of Computing},
  pages={9--21},
  year={2016}
}
